\pdfoutput=1
\documentclass[11pt]{article}

\usepackage[final]{acl}

\usepackage{times}
\usepackage{latexsym}
\usepackage{multirow}
\usepackage{multicol}

\usepackage[framemethod=TikZ]{mdframed}
\usepackage{listings}
\newmdenv[%
    backgroundcolor=gray!10,
    linecolor=black,
    outerlinewidth=0.5pt,
    roundcorner=1mm,
    skipabove=\topsep,
    skipbelow=\topsep,
    font=\ttfamily\footnotesize,
]{promptbox}

\usepackage{cellspace, tabularx}
\addparagraphcolumntypes{X}

\usepackage[T1]{fontenc}
\usepackage[utf8]{inputenc}

\usepackage{microtype}

\usepackage{graphicx}
\usepackage{amsmath, amssymb}
\usepackage{lscape} 
\usepackage{booktabs}
\usepackage{subcaption}
\newcommand{\malp}{\textsc{MultiAlpaca}}
\newcommand{\llm}{\textsc{LLaMA-2-7B}}
\newcommand{\mis}{\textsc{Mistral-7B}}

\usepackage{nameref}
\usepackage{varioref}
\usepackage{hyperref}
\usepackage{cleveref}

\newcommand*\samethanks[1][\value{footnote}]{\footnotemark[#1]}

\title{MAPLE: Multilingual Evaluation of Parameter Efficient Finetuning of Large Language Models}

\author{
Divyanshu Aggarwal$^\dagger$\thanks{Joint first authors}, Ashutosh Sathe$^{\dagger\oplus}$\samethanks, Ishaan Watts$^\dagger$, Sunayana Sitaram$^\dagger$ \\
  $^\dagger$Microsoft Research India\;\;\;$^\oplus$Indian Institute of Technology, Bombay \\
  Contact: \texttt{t-daggarwal@microsoft.com}%aclcheck fails for full list of emails
}
\begin{document}
\maketitle
\begin{abstract}

Parameter Efficient Finetuning (PEFT) has emerged as a viable solution for improving the performance of Large Language Models (LLMs) without requiring massive resources and compute. Prior work on multilingual evaluation has shown that there is a large gap between the performance of LLMs on English and other languages. Further, there is also a large gap between the performance of smaller open-source models and larger LLMs. Finetuning can be an effective way to bridge this gap and make language models more equitable. In this work, we finetune the \llm\ and \mis\ models on two synthetic multilingual instruction tuning datasets to determine its effect on model performance on six downstream tasks covering forty languages in all. Additionally, we experiment with various parameters, such as rank for low-rank adaptation and values of quantisation to determine their effects on downstream performance and find that higher rank and higher quantisation values benefit low-resource languages. We find that PEFT of smaller open-source models sometimes bridges the gap between the performance of these models and the larger ones, however, English performance can take a hit. We also find that finetuning sometimes improves performance on low-resource languages, while degrading performance on high-resource languages. 
\end{abstract}

\section{Introduction}

\begin{figure}[t]
    \centering
    \includegraphics[width=0.5\textwidth]{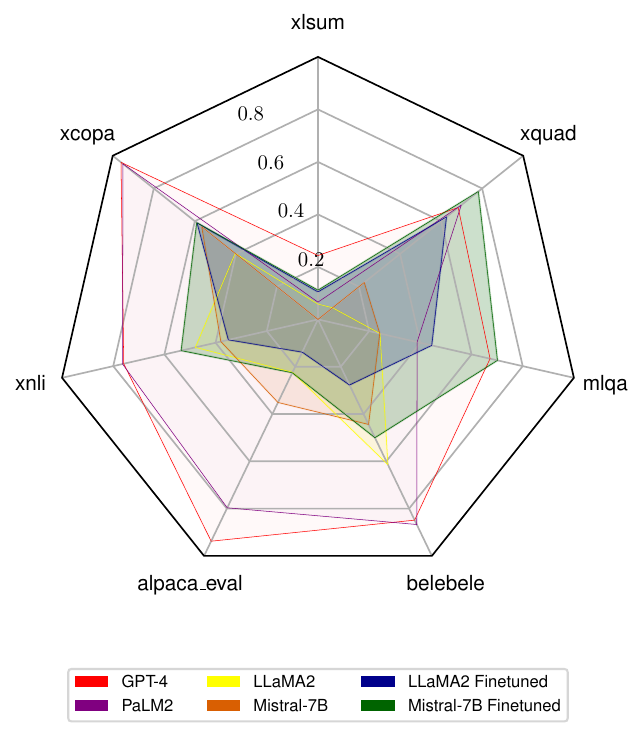}
    \caption{\textbf{Comparison of best parameter efficient instruction finetuned models with other off the shelf LLMs.} %We find modest improvements in performance on average. 
    Notably, the best Mistral instruction finetuned model is able to outperform even GPT-4 and PaLM2 on ``MLQA'' and ``XQUAD'' tasks.}
    \label{fig:all_tasks_all_models}
\end{figure}

Large Language Models (LLMs) show impressive performance on several tasks, sometimes even surpassing human performance. This has been attributed to the vast amounts of training data used during the pretraining phase, as well as various techniques used to align the models during the finetuning phase. Several variants of finetuning exist, including supervised finetuning (SFT) and instruction tuning \cite{gpt4techreport}, where the model is finetuned with task specific data, or instructions on how to perform tasks. However, finetuning all the parameters of the model can be expensive and time consuming, due to the increasing size of language models in recent years. Parameter Efficient Finetuning (PEFT) has emerged as a viable alternative to full finetuning \cite{chen2023parameterefficient}. 

Most studies on LLMs focus on training, finetuning and evaluating models in performing tasks in English. Recent work on comprehensive evaluation of LLM capabilities in non-English settings \cite{ahuja-etal-2023-mega} have shown that LLMs perform far worse on languages other than English. Studies that compare multilingual performance across different models \cite{ahuja2023megaverse}, show that there is a large performance gap between large, proprietary and closed models such as GPT-4 and PaLM2 and smaller open-source models like \llm\ and \mis. 

PEFT techniques like LoRA \cite{hu2022lora} have been shown to strengthen the multilingual capabilities of these open-source LLMs \cite{zhao2024llama}. Moreover, Adapters \cite{pfeiffer-etal-2020-adapterhub} have been proposed to boost the capabilities of language models to newer languages. Since full finetuning of models is not always feasible due to resource and compute constraints, exploring how far PEFT techniques can take us in boosting performance on non-English languages is a promising direction. Adoption of model quantisation techniques \cite{dettmers20228bit, liu2023llmfp4} has also made PEFT LLMs more accessible. 

Not much work has been done on analyzing the impact of different choices, configurations and settings for PEFT on multilingual downstream tasks. In this work we aim to analyse how LoRA rank and quantisation affects the performance of finetuned models across 6 downstream tasks, covering 40 languages in all. We are interested in knowing whether multilingual PEFT can lead to reasonable gains in performance, or whether full finetuning of models is required. Furthermore, we check if it is better to generate multilingual instructions from larger base models or translate existing english instruction to more languages. Lastly, we also study the effect of multilingual finetuning on English performance.

% We also study the variation in the percentage of trainable parameters on high resource and low resource language, can multilingual finetuning affect the performance on english and if PEFT can be an alternative to full finetuning in a multilingual setting. 

Prior works have demonstrated that LoRA 
finetuning is the most effective PEFT out of existing techniques so far \cite{zhuo2024astraios}. Hence, we keep our study limited to analysing different LoRA and QLoRA \cite{dettmers2023qlora} configurations to answer our research questions. Our contributions are as follows:
\vspace{-2mm}
\begin{itemize}
    %\item We benchmark the Llama and Mistral models finetuned on various ranks and quantisation using the  MultiAlpaca dataset and compare the results to models without any finetuning of comparable size as well as much larger size.
    \item We benchmark effects of various ranks and quantisation with \llm\ and \mis\ models finetuned on \textsc{MultiAlpaca} and \textsc{Bactrian-X-22} dataset. We analyse the effects of \% of trainable parameters and quantisation on 6 various tasks and 40 languages.
    \vspace{-2mm}
    \item We study efficacy of finetuning by comparing results with non-finetuned models of similar or larger sizes.
    \vspace{-2mm}
    \item We analyse the effects of multilingual PEFT on English performance to check for degradations due to forgetting.
    \vspace{-2mm}
    \item We experiment with the choice of instruction finetuning dataset to study any variations in model performance on our downstream tasks.
    \vspace{-2mm}
    \item We present results and an analysis of trends across these models and instruction finetuning datasets with directions for future research.
\end{itemize}

\section{Related Work}

\paragraph{Parameter Efficient Finetuning:}
Recently, Parameter Efficient Finetuning has gained significant attention in the NLP research community since full finetuning of LLMs is prohibitively expensive for most organizations. Following early works on adapters \cite{houlsby2019parameter, pfeiffer-etal-2020-adapterhub}, several finetuning techniques like LoRA \cite{hu2022lora}, (IA)$^3$ \cite{liu2022fewshot}, P-Tuning \cite{liu-etal-2022-p} and Prefix Tuning \cite{li2021prefixtuning} have been proposed. These techniques make the compute costs manageable by significantly reduce the number of trainable parameters during finetuning. Several works have used these techniques for efficient cross lingual transfer \cite{ansell-etal-2022-composable}, to tackle catastrophic forgetting \cite{vu-etal-2022-overcoming} or compose multiple adapters \cite{pfeiffer-etal-2021-adapterfusion} for multi-task performance.

%to significantly reduce the amount of compute and cost needed to finetune these models while preserving performance comparable to full finetuning. 

\paragraph{Multilingual Instruction Finetuning:}
Recently, there has been a lot of interest in creating multilingual instruction finetuning datasets to enhance the reasoning capabilities of LLMs on languages other than English \cite{li2023bactrianx, wei2023polylm}. Such datasets, are being used to create LLMs that can serve to larger demographic and can also be more efficient during inference time \cite{jiang2023mistral}. \citet{li2023bactrianx} gained significant performance on multilingual tasks by LoRA finetuning LLaMA and BLOOM on the \textsc{Bactrian-X} dataset. These instruction datasets are generally derived by generation or translation. \citet{wei2023polylm} translated seed tasks of \textsc{Alpaca} dataset \cite{alpaca} to 11 languages and then prompted GPT-3.5-Turbo to generated more instructions in those languages, while \citet{li2023bactrianx} translated \textsc{Alpaca} instructions to 50 other languages using google translate API and generated responses using GPT-3.5-Turbo. Moreover, efforts like \citep{singh2024aya} attempts to create crowdsourced multilinugal instruction datasets to capture better linguistic and cultural nuances.

\paragraph{Quantisation for Model Compression:}
Model quantisation is another way of reducing the overall memory footprint of the LLM. While many popular LLMs (notably \llm\ \cite{touvron2023llama} and \mis\ \cite{jiang2023mistral}) are pre-trained with weights represented in 16 bit floating point numbers \cite{Wu2020IntegerQF}, it is shown that finetuning with lower quantisation yields similar performance. The most popular quantisation techniques -- LLM:Int8() \cite{dettmers20228bit} and 4 bit \cite{liu2023llmfp4} are usually combined with LoRA \cite{dettmers2023qlora} to further reduce the memory footprint of LLM finetuning. 

%\paragraph{Multilingual LLM Evaluation:}
\paragraph{LLM Evaluation:}
Principled LLM evaluation has gained significant interest with demonstrations of increasingly complex abilities of LLMs \cite{brown2020language, cobbe2021training, wei2022chain, shi2023language} on various tasks. However, many evaluations are monolingual or English-only and multilingual evaluation of LLMs \cite{ahuja-etal-2023-mega, asai2023buffet, ahuja2023megaverse} remains a challenging problem. Past work by \citet{ramesh-etal-2023-comparative} has evaluated the effects of model compression techniques such as quantisation, distillation and pruning on LLMs performance on downstream tasks in multilingual setting.

\section{Experiments}

\subsection{Setup}

\paragraph{Finetuning Models:} We finetune open-source, multilingual LLMs on multilingual instruction finetuning datasets. We pick models that are pretrained on multilingual data as it would be unfair to compare English-only LLMs when finetuning on multilingual data. Specifically, we explore PEFT on \llm\ \cite{touvron2023llama} and \mis\ \cite{jiang2023mistral} models.

\paragraph{Finetuning Dataset:} We finetuned our models on \textsc{MultiAlpaca} \cite{wei2023polylm} and \textsc{Bactrian-X} \cite{li2023bactrianx} datasets for all our experiments. 

\paragraph{\textsc{MultiAlpaca}} is a self instruct dataset which follows the same approach as (English-only) \textsc{Alpaca} dataset \cite{alpaca}
by translating seed tasks to 11 languages and then using GPT-3.5-Turbo for response collection. The languages included in the dataset are Arabic, German, Spanish, French, Indonesian, Japanese, Korean, Portuguese, Russian, Thai and Vietnamese. 

\paragraph{\textsc{Bactrian-X}} is a machine translated dataset of the original alpaca-52k and dolly-15k \cite{DatabricksBlog2023DollyV2} datasets. In this dataset, the instructions were translated using google translate API  to 52 diverse languages and responses were generated using GPT-3.5-Turbo. In our experiments we finetune our models on a subset of 11 and 22 languages respectively. We name the 2 datasets as \textsc{Bactrian-X-11} and \textsc{Bactrian-X-22} respectively. For \textsc{Bactrian-X-11} we keep the languages same as in \textsc{MultiAlpaca}. For \textsc{Bactrian-X-11} we pick a subset of 22 languages namely, Afrikaans, Arabic, Bengali, Chinese-Simplified, Dutch, French, German, Gujarati, Hindi, Indonesian, Japanese, Korean,  Marathi, Portuguese, Russian,  Spanish, Swahili, Tamil, Telugu, Thai, Urdu and Vietnamese.  We rename this dataset to \textsc{Bactrian-X-22}. 
Each language in \textsc{Bactrian-X-22} and \textsc{Bactrian-X-11} consists of 67k instructions parallel and responses whereas \textsc{MultiAlpaca} consists of nearly 100k instructions. For \textsc{Bactrian-X-11} to get a dataset of comparable size, we take 20\% instructions (13400) from each language giving us close to 150k instructions in 11 languages. For \textsc{Bactrian-X-22} to get a dataset of comparable size, we take 10\% instructions (6700) from each language giving us close to 150k instructions in 22 languages. 
\\ \\
To sample these subsets of \textsc{Bactrian-X} we first shuffle and partition indices 0-67k and divide them into 10 partitions. Each partition now consists of random indices from 0-67k. Then we iterate over all 11 and 22 languages assigning language $i$ to partition $i \operatorname{mod} 10$. This gives us a partition of 13400 and 6700 indices from each languages which we use to form the instruction tuning dataset from each language. This means that every instruction at index from 0-67k is included in at least two languages. We study whether this enhanced sampling ensuring at least two languages per instruction helps in cross-lingual transfer.

\paragraph{Finetuning Techniques:} We follow the LoRA \cite{dettmers2023qlora, hu2022lora} finetuning recipe for each finetuning run. We finetune models on various ranks and quantisations, specifically LoRA Ranks 8, 16, 32, 64 and 128 and 4bit, 8bit and 16bit quantisation.

\begin{figure*}[ht!]
    \centering
    \includegraphics[width=\textwidth]{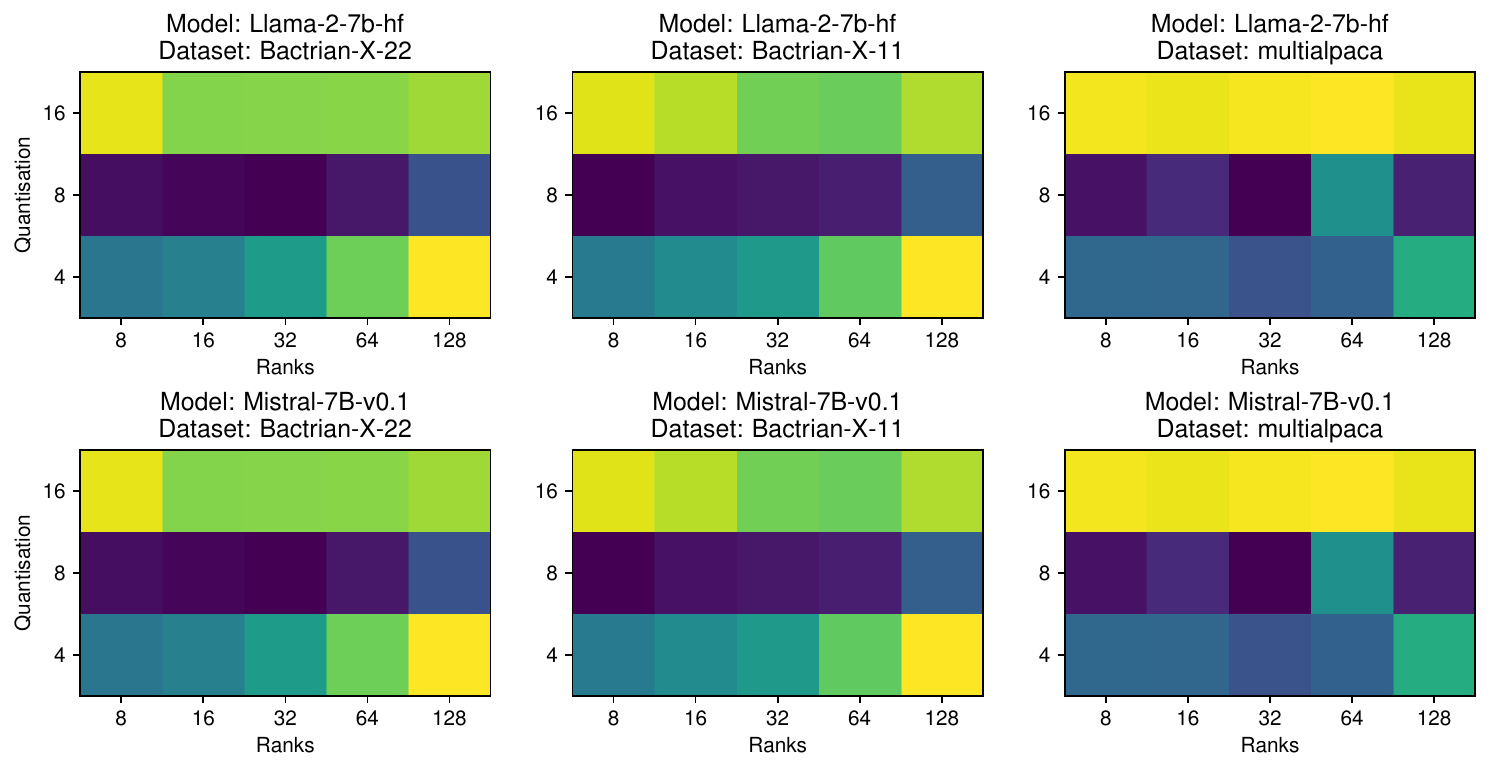}
    \caption{Average model performance of \llm\ and \mis\ finetuned on \textsc{Bactrian-X-22}, \textsc{Bactrian-X-11} and \malp\ across tasks on all rank-quantisation configurations.}
    \label{fig: avg rank quant}
\end{figure*}

\begin{figure*}[ht!]
    \centering
    \includegraphics[width=\textwidth]{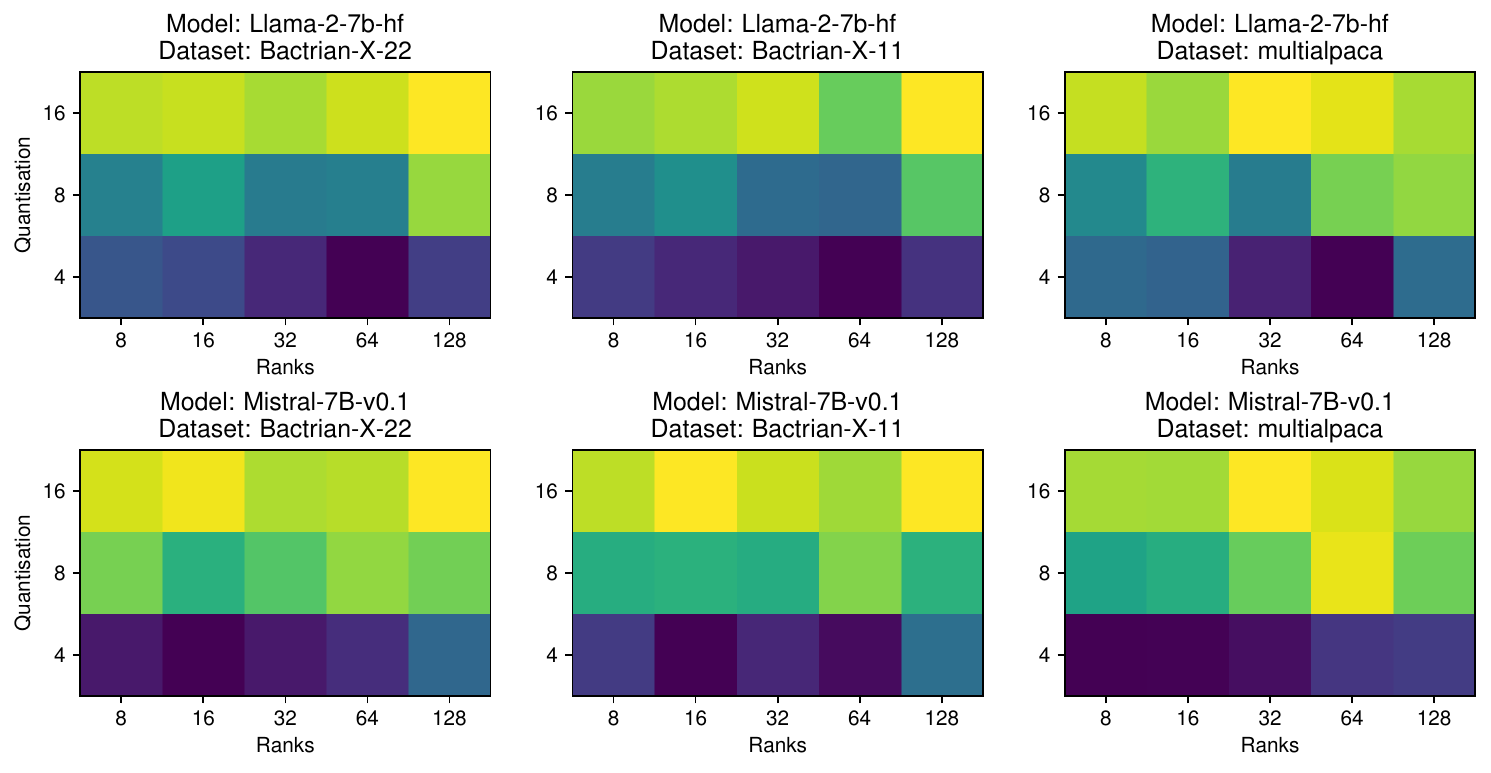}
    \caption{Belebele evaluation results of \llm\ and \mis\ finetuned on \textsc{Bactrian-X-22}, \textsc{Bactrian-X-11} and \malp\ across tasks on all rank-quantisation configurations.}
    \label{fig: belebele rank quant}
\end{figure*}

\subsection{Evaluation} 
\label{sec: evaluation}

We evaluate multilingual capabilities of our finetuned models on three Classification tasks, two Question-Answering tasks and one Summarisation task. We use prompts that are similar to those used in the MEGA benchmarking study \cite{ahuja-etal-2023-mega} but adapted to the Alpaca-style \cite{alpaca} instruction format. We study the impact of multilingual finetuning on English capabilities using Alpaca Eval \cite{li2023alpacaeval}. We use lm-eval-harness \cite{gao2021framework} for the evaluations. LM-Evaluation-Harness is a unified framework for few shot evaluation of language models. This framework standardises the inference and few shot example selection pipeline across tasks and models. We created the task configurations from MEGAVERSE \cite{ahuja2023megaverse} with the \textsc{Alpaca}-style prompt template. We mention these prompts in detail in the Appendix Section \ref{sec:eval prompts}.
%LoRA finetuned models. %We use custom prompts from \citet{ahuja-etal-2023-mega}. 

\paragraph{Classification Datasets:} As part of our evaluation process, we benchmark our finetuned models on several datasets. The \textbf{Belebele} dataset \cite{bandarkar2023belebele}, which is parallel across 122 languages, is evaluated on a subset of 33 languages. We report our results on 30\% of the test split in the zero-shot setting due to resource constraints. We report the results in Table \ref{tab:results_belebele_llama_alpaca},
\ref{tab:results_belebele_mistral_alpaca},
\ref{tab:results_belebele_llama_multialpaca},
\ref{tab:results_belebele_mistral_multialpaca}, 
\ref{tab:results_belebele_llama_bactrian_x_11},
\ref{tab:results_belebele_mistral_bactrian_x_11},
\ref{tab:results_belebele_llama_bactrian} and  \ref{tab:results_belebele_mistral_bactrian}. The \textbf{XNLI} dataset \cite{Conneau2018xnli} consists of 122k training, 2490 validation, and 5010 test examples in 15 languages. We have evaluated our models on 1000 examples from test split with 4 in-context examples sampled from the validation split and report our results in Table \ref{tab:results_xnli_llama_alpaca},
\ref{tab:results_xnli_llama_multialpaca}, 
\ref{tab:results_xnli_llama_bactrian_x_11},
\ref{tab:results_xnli_llama_bactrian},
\ref{tab:results_xnli_mistral_alpaca},
\ref{tab:results_xnli_mistral_multialpaca},
\ref{tab:results_xnli_mistral_bactrian_x_11} and
 \ref{tab:results_xnli_mistral_bactrian}. The \textbf{XCOPA} dataset \cite{ponti2020xcopa} covers 11 languages, and we evaluate our models on Estonian, Indonesian, Italian, Quechua, Thai and   Vietnamese in the 4-shot setting similar to XNLI. We report our results in Table \ref{tab:results_xcopa_llama_alpaca}, 
 \ref{tab:results_xcopa_llama_multialpaca},
\ref{tab:results_xcopa_llama_bactrian_x_11},
\ref{tab:results_xcopa_llama_bactrian}, 
 \ref{tab:results_xcopa_mistral_alpaca}, 
\ref{tab:results_xcopa_mistral_multialpaca},
\ref{tab:results_xcopa_mistral_bactrian_x_11} and
\ref{tab:results_xcopa_mistral_bactrian}. 

\paragraph{Question Answering Datasets:} The \textbf{MLQA} dataset \cite{lewis2020mlqa} contains 5K extractive question-answering instances in 7 languages. For the interest of time, we evaluate our models for 1000 examples of the test split in a 4-shot setting and report our results in Table \ref{tab:results_mlqa_llama_alpaca}, 
\ref{tab:results_mlqa_llama_multialpaca},
\ref{tab:results_mlqa_llama_bactrian_x_11},
\ref{tab:results_mlqa_llama_bactrian},
 \ref{tab:results_mlqa_mistral_alpaca}, 
\ref{tab:results_mlqa_mistral_multialpaca},
\ref{tab:results_mlqa_mistral_bactrian_x_11} and 
\ref{tab:results_mlqa_mistral_bactrian}. The \textbf{XQuAD} dataset \cite{artetxe2020cross} consists of a subset of 240 paragraphs and 1190 question-answer pairs across 11 languages. We use a 4-shot setting similar to MLQA and evaluate 1000 examples of the test split. We report our results in Table \ref{tab:results_xquad_llama_alpaca}, 
\ref{tab:results_xquad_llama_multialpaca},
\ref{tab:results_xquad_llama_bactrian_x_11},
\ref{tab:results_xquad_llama_bactrian},
 \ref{tab:results_xquad_mistral_alpaca},
\ref{tab:results_xquad_mistral_multialpaca},
\ref{tab:results_xquad_mistral_bactrian_x_11}
and  \ref{tab:results_xquad_mistral_bactrian}.

\paragraph{Summarisation Dataset:} The \textbf{XLSUM} dataset \cite{hasan-etal-2021-xl} spans 45 languages, and we evaluate our models in Arabic, Chinese-Simplified, English, French, Hindi, Japanese and Spanish. We evaluate our models on 100 text-summarization pairs from the test split in a zero-shot setting and report our results in Table \ref{tab:results_xlsum_llama_alpaca}, 
\ref{tab:results_xlsum_llama_multialpaca},
\ref{tab:results_xlsum_llama_bactrian_x_11},\ref{tab:results_xlsum_llama_bactrian}, 
 \ref{tab:results_xlsum_mistral_alpaca}, \ref{tab:results_xlsum_mistral_multialpaca},
\ref{tab:results_xlsum_mistral_bactrian_x_11},
and \ref{tab:results_xlsum_mistral_bactrian}. 

\paragraph{English Instruction Following Dataset:} We also use \textbf{AlpacaEval} \cite{li2023alpacaeval} to benchmark English proficiency. We evaluate our models against text-davinci-003 responses on 800 instructions and use GPT4 (gpt-4-32k) as the evaluator. We report our results in Tables \ref{tab:results_alpaca_eval_llama_alpaca},
\ref{tab:results_alpaca_eval_llama_multialpaca}, 
\ref{tab:results_alpaca_eval_llama_bactrian_x_11},
\ref{tab:results_alpaca_eval_llama_bactrian}, 
\ref{tab:results_alpaca_eval_mistral_alpaca},
\ref{tab:results_alpaca_eval_mistral_multialpaca},
\ref{tab:results_alpaca_eval_mistral_bactrian_x_11}, and
\ref{tab:results_alpaca_eval_mistral_bactrian}.

We discuss more about these benchmark datasets in detail in Appendix Section \ref{sec:evaluation dataset details}.
\begin{figure*}[htb!]
    \centering
    \begin{minipage}[c]{0.65\textwidth}
    \includegraphics[width=\textwidth]{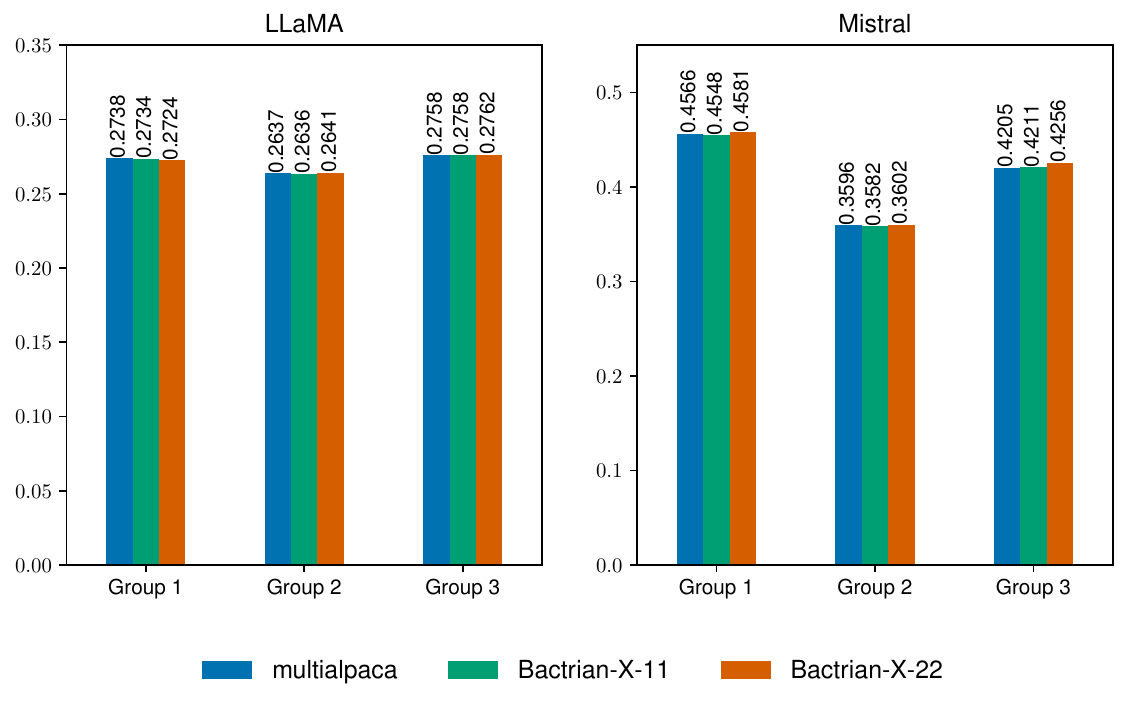}
    \end{minipage}\hfill
    \begin{minipage}[c]{0.3\textwidth}
    \caption{\textbf{Effect of diversity of languages in fine-tuning on downstream task (belebele).} Here Group 1 is the set of 11 languages from \textsc{MultiAlpaca}, Group 2 is the set of 11 languages in \textsc{Bactrian-X-22} but not in \textsc{MultiAlpaca} and Group 3 contains 13 languages present in neither. We find that both models trained on either datasets perform very similar to each other across all 3 groups. Additional details in Tables \ref{tab:results_belebele_llama_alpaca} to \ref{tab:results_belebele_mistral_bactrian}.}
    \label{fig:belebele_analysis}
    \end{minipage}
\end{figure*}

\begin{figure*}[ht!]
\centering
\includegraphics[width=\textwidth]{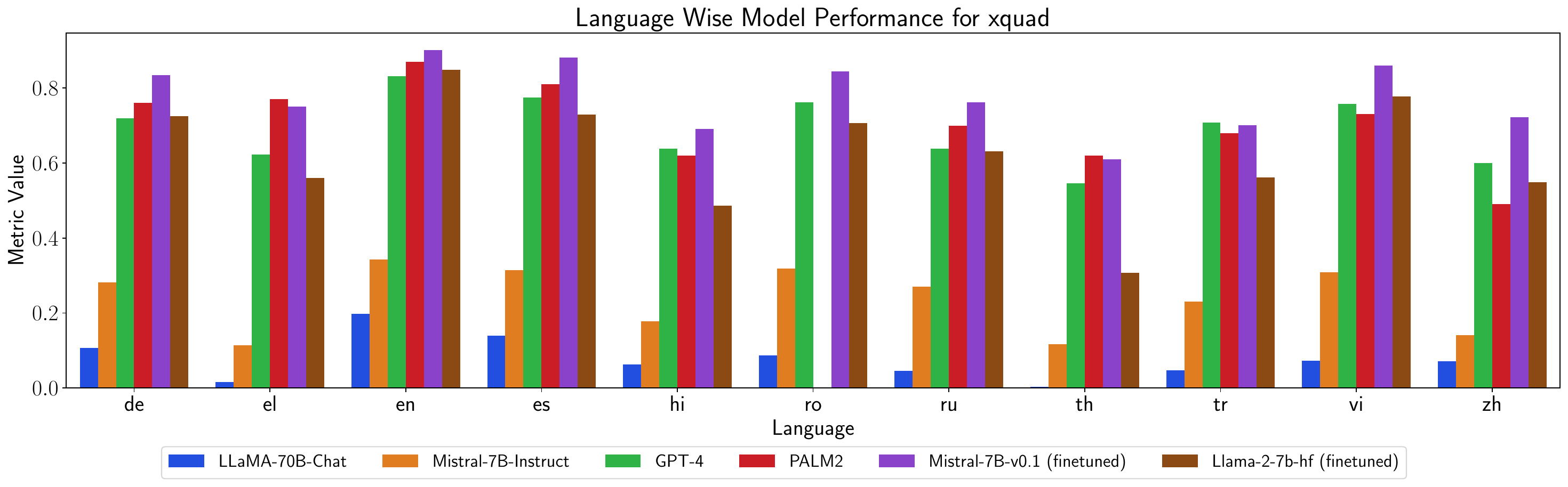}
    \caption{Detailed language-wise comparison of our finetuned \mis\ and \llm\ models with other baselines \cite{ahuja2023megaverse} on Arabic, German, Greek, English, Spanish, Hindi, Romanian, Russian, Thai, Turkish and Vietnamese for XQUAD \cite{artetxe2020cross}.}
    \label{fig:langwise xquad}
\end{figure*}

\section{Analysis of Results}

\paragraph{Analysis of Rank and Quantisation}

In this study we aim to analyse the trade-offs between cost of compute and model performance. Both the \llm\ and \mis\ models were finetuned on all rank-quantisation configurations using the \textsc{MultiAlpaca} and \textsc{Bactrian-X-22} datasets resulting into 60 models.

% We do this by LoRA fine-tuning the LLaMA and Mistral models and varying the LoRA Rank and model Quantisation. A smaller rank will decrease the number of trainable parameters while a smaller quantisation will lower the model precision and vice-versa.

We evaluate our finetuned models across the six benchmarking datasets mentioned in Section \ref{sec: evaluation}. We present the averaged results across these datasets in Fig \ref{fig: avg rank quant}. Lighter colours (yellow) indicate higher performance and it decreases with darker shades (blue). For \mis\, we can see a clear trend for both the finetuning datasets. Decreasing the quantisation can lead to a hit in model performance. For \llm\, the trend is not very clear but the highest quantisation gives the best results. Additionally, higher ranks seem to give slightly better performance. According to our studies, using Rank 32 or 64 with 16bit Quantisation works the best on average. This can be inferred very clearly from our results on Belebele in Fig \ref{fig: belebele rank quant}. To delve deeper, we provide a detailed task-wise performance in Fig \ref{fig:all_rank_quant}. 
% For classfication tasks like Belebele, XCOPA and XNLI, performance decreases with quantisation for all variations except Mistral on XNLI where there is no clear trend. 

\paragraph{\textsc{MultiAlpaca} v/s \textsc{Bactrian-X-11} as Instruction Finetuning Dataset} Here, we aim to study the model performance finetuned on multilingual instruction dataset created in 2 different settings i.e. LLM generated (\textsc{multialpaca}) and machine-translated (\textsc{Bactrian-X-11}). In \textsc{MultiAlpaca}, both multilingual instructions and their responses are generated using GPT-3.5-Turbo from translated \textsc{Alpaca} seed instructions. While in \textsc{Bactrian-X-11}, the final set of \textsc{Alpaca} instructions were translated and then responses were collected using GPT-3.5-Turbo.

In our findings, we observe that models trained on both datasets give similar performance on average across tasks and languages, implying that method of instruction finetuning data creation has little to no effect on the model performance. Rather, we observe that multilingual capabilities of base model is a good indicator of the finetuned model performance across tasks. From \citet{ahuja2023megaverse} we know that \mis\ is a better base multilingual model than \llm\,and  we observe that multilingual capabilities of \mis\ also reaps greater benefits of multilingual instruction finetuning.

Hence, in our experiments, we observe that for creating multilingual instruction datasets both approaches are equivalent - generating multilingual data from seed tasks or translating an existing English instruction dataset to more languages. 

% In tables \ref{tab:results_belebele_llama_multialpaca}, \ref{tab:results_belebele_llama_bactrian} , \ref{tab:results_belebele_mistral_bactrian}, \ref{tab:results_belebele_mistral_multialpaca}, \ref{tab:results_mlqa_llama_bactrian}, \ref{tab:results_mlqa_llama_multialpaca}, 

\begin{table*}[htb!]
\small
\centering
\begin{tabular}{llrrrrrrc}
    \toprule
        \textbf{model} & \textbf{finetuning dataset} & \textbf{xnli} & \textbf{xcopa} & \textbf{xquad} & \textbf{belebele} & \textbf{mlqa} & \textbf{xlsum} & \textbf{Model Average}\\ 
        \midrule
        GPT-4 & NA & 0.75 & 0.90 & 0.69 & 0.85 & 0.67 & \textbf{0.25} & 0.69\\ 
        Mistral-7B-Instruct& NA & 0.38 & 0.53 & 0.23 & 0.44 & 0.24 & NA & 0.37\\ 
        Llama-2-70b-chat & NA & 0.48 & 0.39 & 0.07 & 0.61 & 0.24 & 0.08 & 0.31\\ 
        PaLM2& NA & \textbf{0.76} & \textbf{0.96} & 0.70 & \textbf{0.87} & 0.39 & 0.07 & 0.62\\ 
        \midrule
        \multirow{3}{*}{Llama-2-7b} & \malp & 0.35 & 0.58 & 0.64 & 0.28 & 0.41 & 0.10 & 0.39\\ 
         & \textsc{Bactrian-X-22} & 0.35 & 0.58 & 0.63 & 0.28 & 0.44 & 0.08 & 0.39\\ 
         & \textsc{Bactrian-X-11} & 0.35 & 0.59 & 0.63 & 0.28 & 0.44 & 0.07 & 0.39\\ 
         & alpaca & 0.35 & 0.58 & 0.63 & 0.28 & 0.35 & 0.07 & 0.38\\ 
        \midrule
        \multirow{3}{*}{Mistral-7b} & \malp & 0.53 & 0.59 & \textbf{0.79} & 0.43 & \textbf{0.70} & 0.14 & 0.53\\ 
         & \textsc{Bactrian-X-22} & 0.52 & 0.59 & \textbf{0.79} & 0.42 & \textbf{0.70} & 0.14 & 0.53\\ 
         & \textsc{Bactrian-X-11} & 0.53 & 0.60 & \textbf{0.79} & 0.42 & \textbf{0.70} & 0.10 & 0.52\\
         & alpaca & 0.53 & 0.59 & 0.78 & 0.45 & \textbf{0.70} & 0.10 & 0.52\\ 
        \bottomrule
    \end{tabular}

\caption{Detailed Task Wise Performance Comparison between GPT-4, PaLM-2, LLaMA-70B-chat, Mistral-7B-Instruct and finetuned models with best rank quantisation. Baseline numbers are referred from \citet{ahuja2023megaverse}.}
\label{tab:taskwise_average_results}

\end{table*}

\begin{figure*}[htb!]
    \centering
    \includegraphics[width=\textwidth]{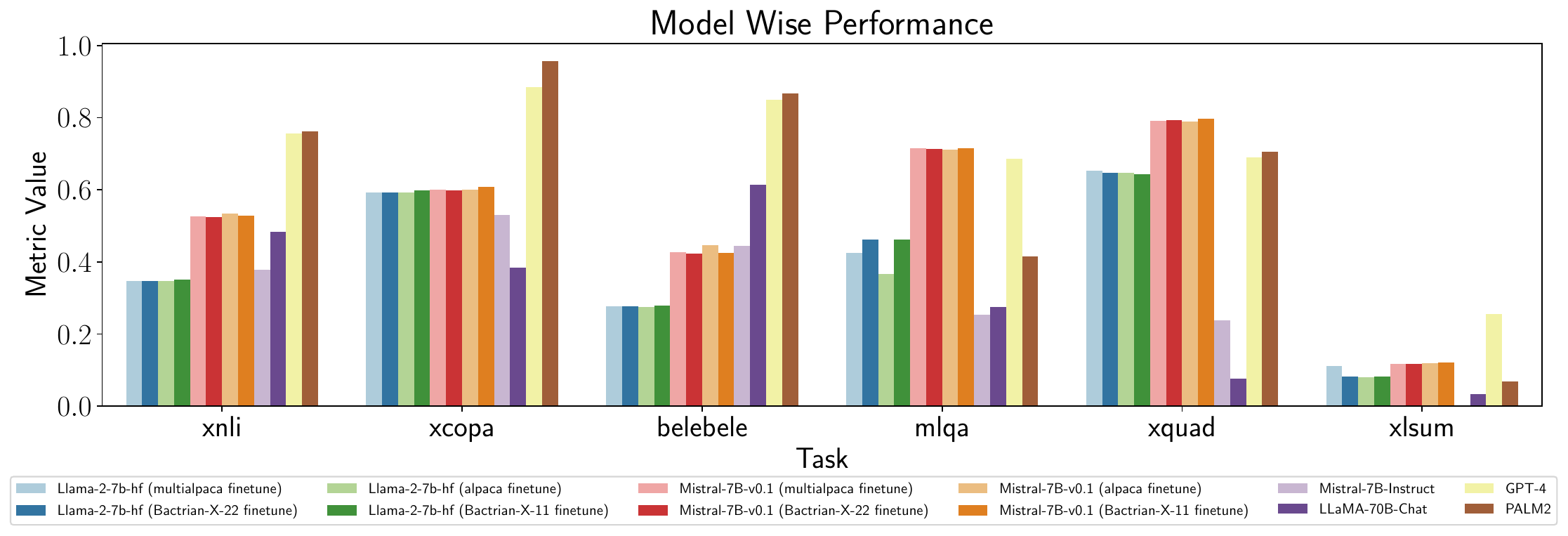}
    \caption{Task Wise Performance Comparison of Llama-2-70B-Chat, GPT-4, PaLM-2, Mistral-7B-Chat and our finetuned models averaged across languages.
    }
    \label{fig:taskwise performance }
\end{figure*}

\paragraph{Effect of Number of Languages in Training Data}
\label{sec:effect_of_langs}

Our finetuning datasets \textsc{Bactrian-X-11} and \textsc{Bactrian-X-22} have 11 and 22 languages respectively. We want to study if the additional 11 languages in \textsc{Bactrian-X-22} help the model perform better at multilingual tasks. % Any other study we performing here?

We do a case study on the ``belebele'' task which consists of 50+ languages. We sample 35 of these and divide them into 3 groups. The first group (Group 1) consists of 11 languages from \textsc{MultiAlpaca}, the next group (Group 2) consists of 11 languages present in \textsc{Bactrian-X-22} but absent from \textsc{Bactrian-X-11}. The final group (Group 3) contains 13 languages that are not present in any of our finetuning datasets. We compute average accuracy of finetuned models across all ranks on the 3 groups and present them in Figure \ref{fig:belebele_analysis}. We find that the larger number of languages in \textsc{Bactrian-X-22} do not necessarily help. This behavior is consistent with the findings of \citet{shaham2024multilingual}. Moreover,we observe that for other tasks as well \textsc{Bactrian-X-22} dataset has no edge over \malp\ as we can see from Table \ref{tab:taskwise_average_results} and Figure \ref{fig:taskwise performance }.

Subsequently we observe that \textsc{Bactrian-X-11} performs more similarly (though marginally) with \malp\ than \textsc{Bactrian-X-22}, showing that number of languages in the training data can play a big role in downstream task performance.

\begin{table}[t!]
    \scriptsize
    \centering
    \begin{tabular}{llrrc}
    \toprule
    %\textbf{model} & \textbf{dataset} & \textbf{rank} &  \textbf{quantisation} & \textbf{winrate} \\
    \multicolumn{3}{c}{\textbf{model}} & & \textbf{winrate} \\
    \midrule
    
      \multicolumn{3}{c}{GPT-4} & & \textbf{93.78} \\
      \multicolumn{3}{c}{PaLM2} & & 79.66 \\
      \multicolumn{3}{c}{Llama-70B-Chat}  & & 22.36  \\
      \multicolumn{3}{c}{Mistral-7B-Instruct} & & 35.12 \\
      \midrule
      \textbf{model} & \textbf{dataset} & \textbf{rank} &  \textbf{quantisation} & \textbf{winrate} \\
     \midrule
      \multirow{3}{*}{Llama-2-7B} & Alpaca & 128 & 16 & 13.28 \\
         & \textsc{Bactrian-X-22} & 64 & 16 & 13.73 \\
         & \textsc{Bactrian-X-11} & 16 & 16 & \textbf{13.83}\\
         & \malp & 128 & 16 & \textbf{13.73} \\
         \midrule
      \multirow{3}{*}{Mistral-7B} & Alpaca & 64 & 8 & \textbf{24.47} \\
       & \textsc{Bactrian-X-22} & 16 & 8 & 22.07 \\
       & \textsc{Bactrian-X-11} & 128 & 16 & 22.57\\
       & \malp & 32 & 8 & 22.45 \\
    
    \bottomrule
    \end{tabular}
    \caption{Best AlpacaEval Scores for each model, dataset, rank and quantisation configuration and GPT-4, PaLM-2, LLaMA-70B-Chat and Mistral-7B-Instruct baselines.}
    \label{tab:alpaca eval best scores}
\end{table}

\paragraph{Effect of Multilingual Finetuning on performance on downstream tasks for High Resource v/s Low Resource Languages}

For XNLI and MLQA, we observe that finetuning improves performance on low-resource languages but worsens performance on high-resource languages. In Belebele, we find that finetuning worsens the performance for all languages for both models. For XCOPA, we get the same or better results for all languages with finetuning. For XQUAD, multilingual finetuning boosts performance for all languages for both \llm\ and \mis\ and even surpasses GPT-4 as seen in Fig \ref{fig:langwise xquad}. Moreover, for XLSUM parameter efficient finetuned models does not perform at par with fully finetuned or larger LLMs due to the generative nature of the task. This shows that overall, PEFT on smaller LLMs using multilingual instruction data can prove to be beneficial and can bridge the gap between smaller open-source models and large proprietary models.

Moreover, we observe that language-wise performance does not get affected by the choice of the training dataset, i.e. \malp\, \textsc{Bactrian-X-11} and \textsc{Bactrian-X-22} have similar performances across languages and tasks. \textsc{Bactrian-X-22} sometimes leads to better performance on some low-resources languages as it includes more languages in the training data. 

\paragraph{Analysis of Performance on English}

In Table \ref{tab:alpaca eval best scores} we compare models on the English AlpacaEval benchmark. Overall, the capability of model to follow English instruction reduces drastically after multilingual finetuning. In general, finetuning on English-only \textsc{Alpaca} dataset or using higher capacity adapters (higher rank, better quantisation) seem to help in preserving the performance on English instruction finetuning. 

More concisely, we observe that \mis\ is able to preserve more English capabilities than \llm\ on AlpacaEval. While, \textsc{Bactrian-X-11}, \textsc{Bactrian-X-22} and \malp have difference in performance in English for the respective finetuned model.

Moreover, \textsc{Bactrian-X-11} is able to preserve more english knowledge than \textsc{Bactrian-X-22}. This could be attributed to higher number of languages in \textsc{Bactrian-X-22} than \textsc{Bactrian-X-11} as the curse of multilinguality kicks in leading to forgetting in english.

Furthermore, in a task-wise analysis we observe that multilingual finetuning leads to deterioration in English performance in Belebele, while in XNLI it deteriorates for \llm\ and improves for \mis\. For question-answering tasks MLQA and XQUAD, multilingual finetuning leads to improvement in performance. In XLSUM the performance improves for \llm\ when finetuned on multilingual data, while for \mis\ the performance decreases or remains the same.

\paragraph{Task Wise Performance Analysis}
\label{sec:taskwise_perf}
We analyse the average performance across languages on each task for GPT-4, PaLM-2, LLaMA-2-70B-Chat, Mistral-7B-Instruct and \llm\ and \mis\ models finetuned using \malp, \textsc{Alpaca}, \textsc{Bactrian-X-11} and \textsc{Bactrian-X-22}. In Figure \ref{fig:taskwise performance } we can see that finetuning is usually better or at par with LLaMA-70B-Chat and Mistral-7B-Instruct which are not finetuned on multilingual data. We also compare the performance of finetuned models with LLMs like GPT-4 and PaLM2 and English instruction tuned versions of these models provided in \cite{ahuja2023megaverse}. Table \ref{tab:taskwise_average_results} shows the best model score averaged across languages per task. We observe that \mis\ beats GPT-4 and fairs as the best model on XQUAD and MLQA. While in XNLI, XCOPA and Belebele, it bridges the gap between GPT-4 and PaLM-2 by 20\% on an average.  

While we see some gap being bridged (20\% on average) on classification tasks (XNLI, XCOPA and Belebele) using PEFT, it tends to beat larger models on question answering tasks (MLQA and XQUAD) like GPT-4. While on XLSUM there is no significant difference in performance after PEFT. 

Interestingly, we observe that finetuning \llm\ and \mis\ on \textsc{Alpaca} leads to comparable results with finetuning on \malp\, \textsc{Bactrian-X-11} and \textsc{Bactrian-X-22}. This can be due to the parameter efficient nature of the finetuning which prevents catastrophic forgetting and helps the model learn the instruction following ability from the English instruction data. Second  reason can be the difference in the token fertility of \llm\ and \mis\ as shown by \citet{ahuja2023megaverse}. We can deduce that \mis\ having higher token fertility and being a better base multilingual model can benefit greatly from English instruction dataset and show excellent cross-lingual transfer. While \llm\ having a lower token fertility does not benefit greatly from even multilingual instruction finetuning, as most multilingual LLaMAs resort to vocabulary expansion during pre-finetuning phase \cite{zhao2024llama}. We illustrate the language-wise analysis of our model results in Figure \ref{fig:langwise xquad}, \ref{fig:langwise belebele 0},% \ref{fig:langwise belebele 7}, \ref{fig:langwise belebele 14}, \ref{fig:langwise belebele 21}, 
\ref{fig:langwise mlqa}, \ref{fig:langwise xcopa}, \ref{fig:langwise xnli}, \ref{fig:langwise xnli} and \ref{fig:langwise xlsum}. 

While we compare our multilingual finetuned models with models finetuned on English, we should note that we do not have complete information about instruction datasets used for Llama-70B-chat and Mistral-7B-Instruct. Hence, there may be chances of data contamination for some datasets in these models \cite{ahuja-etal-2023-mega} or the presence of multilingual instruction data in them.

% \paragraph{RQ1: Which Rank and Quantisation Combination leads to most optimal performance for our Multilingual Tasks?}

% \paragraph{RQ2: Out of Generated and translated instruction data which one leads to a better instruction tuned model?}

% \paragraph{RQ3: Does more Languages in training data leads to better performance?}

% \paragraph{RQ4: does multilingual finetuning leads to catastrophic forgetting on English?}

% \paragraph{RQ5: How much gap is being closed by multilingual parameter efficient finetuning?}

% \section{Key insights}

% \section{Future Work}

\section{Conclusion}

% In this work we perform an extensive analysis of effects of rank, quantisation and model when finetuned using LoRA. We study their effects on various multilingual tasks and Alpaca Eval to study the effect on English performance. Our findings shows that there is no one size fits all approach when finetuning LLMs using PEFT. However, Quantisation 8 and 16 with rank 64 are usually the safest bet to achieve decent performance in most tasks while 4 bit quantisation tends to be  more unstable. We also show that finetuned model performance is very much dependent on the base model as well, where a better base model leads to a better finetuned model in multilingual setting. We find that PEFT with multilingual data can bridge the gap between the performance of smaller open source models and larger closed models on some downstream tasks, however, full fine-tuning, better representation of non-English languages during fine-tuning or other techniques applied during modeling may be required for improving multilingual performance.

In this paper we perform an extensive analysis of how rank, quantisation, finetuning dataset and base LLM effects the performance of the finetuned models on 6 multilingual tasks and AlpacaEval when finetuned in a paramater efficient manner. 

\begin{itemize}
    \item Crosslingual transfer DOES happen even in parameter efficient finetuning.
    \item \textsc{Alpaca} (English-only instruction finetuning dataset) is comparable to \malp and Bactrian-X-22 in multilingual downstream task performance. We hypothesize that this is due to Mistral being a superior model due to its better tokenizer, and that PEFT prevents catastrophic forgetting compared to full finetuning.
    \item Having more languages in the finetuning datasets does not necessarily mean significantly better multilingual performance (Section \ref{sec:effect_of_langs}) if the dataset sizes are comparable.
    \item There are no significant differences on the downstream tasks when the models are finetuned on translated or LLM generated training datasets. 
    \item Quality and abilities of the base model far outweigh the dataset or training method for parameter efficient multilingual instruction finetuning.
    \item Higher capacity adapters (i.e. higher ranks or better quantisations) are better at maintaining English performance along with multilingual downstream task performance.
\end{itemize}

%We find that higher ranks leads to gains in performance while choice quantisation during finetuning has little effect downstream task performance. We also observe that choice of finetuning dataset amongst \malp, Bactrian-X-22 and Alpaca. We observe that even alpaca finetuning on english data can lead to signification performance gains on multilingual downstream tasks. Moreover, our study also observes that higher number of languages in instruction finetuning dataset does not necessarily leads to better performance on multilingual downstream tasks. Most Importantly, a stronger mulitlingual base LLM can lead to better gains post finetuning and is usually the most important factor to consider during multilingual finetuning. We also observe a drop in english instruction following capabilities by benchmarking Alpaca Eval on these models.  

We also beat GPT-4 on question-answering tasks (MLQA and XQUAD) using just multilingual PEFT on \mis\, showing that multilingual finetuning of 7B parameter LLMs is a promising direction for the future to bridge the gap of performance on multilingual downstream tasks.

\section{Future Work}
\paragraph{More PEFT Techniques}
This study explores the effects of PEFT using LoRA, while newer techniques by \citet{ansell2024scaling} can also be promising to study the parameter efficient techniques for multilingual instruction tuning for these LLMs. We can also work towards building better PEFT techniques for specifically multilingual settings or crosslingual transfer for LLMs like MAD-X for encoder-like models \cite{pfeiffer-etal-2020-mad}.

\paragraph{Mitigating Curse of Multilinguality}
As more multilingual LLMs release with time, it is also important to understand how many languages a X billion parameter can handle as the curse of multilinguality kicks in as we add more languages in the training data of LLMs \cite{chang2023multilinguality}. This makes it important to study the effect of vocabulary expansion \cite{zhao2024llama} and scaling laws \cite{hoffmann2022training} to better understand the capabilities of LLMs for diverse languages in pretraining as well as instruction training phase. Lastly, How to mitigate the curse of multilinguality can be a another possible direction for future work \cite{blevins2024breaking, pfeiffer-etal-2022-lifting}.  

\paragraph{Better Multilingual Instruction Datasets}
While the datasets used in this study are derived from \textsc{Alpaca}, newer instruction datasets using Chain Of Thought prompting \cite{mukherjee2023orca, mitra2023orca} leads to better reasoning capabilities on smaller LLMs (7 billion parameters), which can be explored as future work. More controlled crowd-sourcing efforts like \citet{singh2024aya} can also lead to better multilingual instruction datasets.

\section{Limitations}

Our evaluation is performed using standard benchmarks, which has known limitations. Datasets used to create benchmarks may have been seen by models during pretraining or finetuning, and due to lack of transparency about the datasets used for training we cannot rule out test data contamination. Second, we use synthetic datasets that are created by prompting LLMs to finetune our models, this can lead to bias, which is also a limitation of the work. Finally, we compare the results obtained by our models to results from the MEGAVERSE benchmarking study while comparing the differences between finetuned models and models that are not finetuned for multilingual performance, which may have some differences in prompting and setup.

% \section{Discussion}

\bibliography{anthology,custom}

\begin{thebibliography}{52}
\expandafter\ifx\csname natexlab\endcsname\relax\def\natexlab#1{#1}\fi

\bibitem[{Ahuja et~al.(2023{\natexlab{a}})Ahuja, Diddee, Hada, Ochieng, Ramesh, Jain, Nambi, Ganu, Segal, Ahmed, Bali, and Sitaram}]{ahuja-etal-2023-mega}
Kabir Ahuja, Harshita Diddee, Rishav Hada, Millicent Ochieng, Krithika Ramesh, Prachi Jain, Akshay Nambi, Tanuja Ganu, Sameer Segal, Mohamed Ahmed, Kalika Bali, and Sunayana Sitaram. 2023{\natexlab{a}}.
\newblock \href {https://aclanthology.org/2023.emnlp-main.258} {{MEGA}: Multilingual evaluation of generative {AI}}.
\newblock In \emph{Proceedings of the 2023 Conference on Empirical Methods in Natural Language Processing}, pages 4232--4267, Singapore. Association for Computational Linguistics.

\bibitem[{Ahuja et~al.(2023{\natexlab{b}})Ahuja, Aggarwal, Gumma, Watts, Sathe, Ochieng, Hada, Jain, Axmed, Bali, and Sitaram}]{ahuja2023megaverse}
Sanchit Ahuja, Divyanshu Aggarwal, Varun Gumma, Ishaan Watts, Ashutosh Sathe, Millicent Ochieng, Rishav Hada, Prachi Jain, Maxamed Axmed, Kalika Bali, and Sunayana Sitaram. 2023{\natexlab{b}}.
\newblock \href {http://arxiv.org/abs/2311.07463} {{MEGAVERSE}: Benchmarking large language models across languages, modalities, models and tasks}.

\bibitem[{Ansell et~al.(2022)Ansell, Ponti, Korhonen, and Vuli{\'c}}]{ansell-etal-2022-composable}
Alan Ansell, Edoardo Ponti, Anna Korhonen, and Ivan Vuli{\'c}. 2022.
\newblock \href {https://doi.org/10.18653/v1/2022.acl-long.125} {Composable sparse fine-tuning for cross-lingual transfer}.
\newblock In \emph{Proceedings of the 60th Annual Meeting of the Association for Computational Linguistics (Volume 1: Long Papers)}, pages 1778--1796, Dublin, Ireland. Association for Computational Linguistics.

\bibitem[{Ansell et~al.(2024)Ansell, Vulić, Sterz, Korhonen, and Ponti}]{ansell2024scaling}
Alan Ansell, Ivan Vulić, Hannah Sterz, Anna Korhonen, and Edoardo~M. Ponti. 2024.
\newblock \href {http://arxiv.org/abs/2401.16405} {Scaling sparse fine-tuning to large language models}.

\bibitem[{Artetxe et~al.(2020)Artetxe, Ruder, and Yogatama}]{artetxe2020cross}
Mikel Artetxe, Sebastian Ruder, and Dani Yogatama. 2020.
\newblock On the cross-lingual transferability of monolingual representations.
\newblock In \emph{Proceedings of the 58th Annual Meeting of the Association for Computational Linguistics}, pages 4623--4637.

\bibitem[{Asai et~al.(2023)Asai, Kudugunta, Yu, Blevins, Gonen, Reid, Tsvetkov, Ruder, and Hajishirzi}]{asai2023buffet}
Akari Asai, Sneha Kudugunta, Xinyan~Velocity Yu, Terra Blevins, Hila Gonen, Machel Reid, Yulia Tsvetkov, Sebastian Ruder, and Hannaneh Hajishirzi. 2023.
\newblock Buffet: Benchmarking large language models for few-shot cross-lingual transfer.
\newblock \emph{arXiv cs.CL 2305.14857}.

\bibitem[{Bandarkar et~al.(2023)Bandarkar, Liang, Muller, Artetxe, Shukla, Husa, Goyal, Krishnan, Zettlemoyer, and Khabsa}]{bandarkar2023belebele}
Lucas Bandarkar, Davis Liang, Benjamin Muller, Mikel Artetxe, Satya~Narayan Shukla, Donald Husa, Naman Goyal, Abhinandan Krishnan, Luke Zettlemoyer, and Madian Khabsa. 2023.
\newblock The belebele benchmark: a parallel reading comprehension dataset in 122 language variants.
\newblock \emph{arXiv preprint arXiv:2308.16884}.

\bibitem[{Blevins et~al.(2024)Blevins, Limisiewicz, Gururangan, Li, Gonen, Smith, and Zettlemoyer}]{blevins2024breaking}
Terra Blevins, Tomasz Limisiewicz, Suchin Gururangan, Margaret Li, Hila Gonen, Noah~A. Smith, and Luke Zettlemoyer. 2024.
\newblock \href {http://arxiv.org/abs/2401.10440} {Breaking the curse of multilinguality with cross-lingual expert language models}.

\bibitem[{Brown et~al.(2020)Brown, Mann, Ryder, Subbiah, Kaplan, Dhariwal, Neelakantan, Shyam, Sastry, Askell, Agarwal, Herbert-Voss, Krueger, Henighan, Child, Ramesh, Ziegler, Wu, Winter, Hesse, Chen, Sigler, Litwin, Gray, Chess, Clark, Berner, McCandlish, Radford, Sutskever, and Amodei}]{brown2020language}
Tom Brown, Benjamin Mann, Nick Ryder, Melanie Subbiah, Jared~D Kaplan, Prafulla Dhariwal, Arvind Neelakantan, Pranav Shyam, Girish Sastry, Amanda Askell, Sandhini Agarwal, Ariel Herbert-Voss, Gretchen Krueger, Tom Henighan, Rewon Child, Aditya Ramesh, Daniel Ziegler, Jeffrey Wu, Clemens Winter, Chris Hesse, Mark Chen, Eric Sigler, Mateusz Litwin, Scott Gray, Benjamin Chess, Jack Clark, Christopher Berner, Sam McCandlish, Alec Radford, Ilya Sutskever, and Dario Amodei. 2020.
\newblock \href {https://proceedings.neurips.cc/paper_files/paper/2020/file/1457c0d6bfcb4967418bfb8ac142f64a-Paper.pdf} {Language models are few-shot learners}.
\newblock In \emph{Advances in Neural Information Processing Systems}, volume~33, pages 1877--1901. Curran Associates, Inc.

\bibitem[{Chang et~al.(2023)Chang, Arnett, Tu, and Bergen}]{chang2023multilinguality}
Tyler~A. Chang, Catherine Arnett, Zhuowen Tu, and Benjamin~K. Bergen. 2023.
\newblock \href {http://arxiv.org/abs/2311.09205} {When is multilinguality a curse? language modeling for 250 high- and low-resource languages}.

\bibitem[{Chen et~al.(2023)Chen, Zhang, Shi, Li, Smola, and Yang}]{chen2023parameterefficient}
Jiaao Chen, Aston Zhang, Xingjian Shi, Mu~Li, Alex Smola, and Diyi Yang. 2023.
\newblock \href {http://arxiv.org/abs/2301.01821} {Parameter-efficient fine-tuning design spaces}.

\bibitem[{Cobbe et~al.(2021)Cobbe, Kosaraju, Bavarian, Chen, Jun, Kaiser, Plappert, Tworek, Hilton, Nakano, Hesse, and Schulman}]{cobbe2021training}
Karl Cobbe, Vineet Kosaraju, Mohammad Bavarian, Mark Chen, Heewoo Jun, Lukasz Kaiser, Matthias Plappert, Jerry Tworek, Jacob Hilton, Reiichiro Nakano, Christopher Hesse, and John Schulman. 2021.
\newblock \href {http://arxiv.org/abs/2110.14168} {Training verifiers to solve math word problems}.

\bibitem[{Conneau et~al.(2018)Conneau, Rinott, Lample, Williams, Bowman, Schwenk, and Stoyanov}]{Conneau2018xnli}
Alexis Conneau, Ruty Rinott, Guillaume Lample, Adina Williams, Samuel Bowman, Holger Schwenk, and Veselin Stoyanov. 2018.
\newblock {XNLI}: Evaluating cross-lingual sentence representations.
\newblock In \emph{Proceedings of EMNLP 2018}, pages 2475--2485.

\bibitem[{Conover et~al.(2023)Conover, Hayes, Mathur, Xie, Wan, Shah, Ghodsi, Wendell, Zaharia, and Xin}]{DatabricksBlog2023DollyV2}
Mike Conover, Matt Hayes, Ankit Mathur, Jianwei Xie, Jun Wan, Sam Shah, Ali Ghodsi, Patrick Wendell, Matei Zaharia, and Reynold Xin. 2023.
\newblock \href {https://www.databricks.com/blog/2023/04/12/dolly-first-open-commercially-viable-instruction-tuned-llm} {Free dolly: Introducing the world's first truly open instruction-tuned llm}.

\bibitem[{Dettmers et~al.(2022)Dettmers, Lewis, Shleifer, and Zettlemoyer}]{dettmers20228bit}
Tim Dettmers, Mike Lewis, Sam Shleifer, and Luke Zettlemoyer. 2022.
\newblock \href {http://arxiv.org/abs/2110.02861} {8-bit optimizers via block-wise quantization}.

\bibitem[{Dettmers et~al.(2023)Dettmers, Pagnoni, Holtzman, and Zettlemoyer}]{dettmers2023qlora}
Tim Dettmers, Artidoro Pagnoni, Ari Holtzman, and Luke Zettlemoyer. 2023.
\newblock \href {http://arxiv.org/abs/2305.14314} {{QLoRA}: Efficient finetuning of quantized {LLMs}}.
\newblock In \emph{Neural Information Processing Systems (NeurIPS)}.

\bibitem[{Gao et~al.(2021)Gao, Tow, Biderman, Black, DiPofi, Foster, Golding, Hsu, McDonell, Muennighoff et~al.}]{gao2021framework}
Leo Gao, Jonathan Tow, Stella Biderman, Sid Black, Anthony DiPofi, Charles Foster, Laurence Golding, Jeffrey Hsu, Kyle McDonell, Niklas Muennighoff, et~al. 2021.
\newblock A framework for few-shot language model evaluation.
\newblock \emph{Version v0. 0.1. Sept}.

\bibitem[{Hasan et~al.(2021)Hasan, Bhattacharjee, Islam, Mubasshir, Li, Kang, Rahman, and Shahriyar}]{hasan-etal-2021-xl}
Tahmid Hasan, Abhik Bhattacharjee, Md.~Saiful Islam, Kazi Mubasshir, Yuan-Fang Li, Yong-Bin Kang, M.~Sohel Rahman, and Rifat Shahriyar. 2021.
\newblock \href {https://aclanthology.org/2021.findings-acl.413} {{XL}-sum: Large-scale multilingual abstractive summarization for 44 languages}.
\newblock In \emph{Findings of the Association for Computational Linguistics: ACL-IJCNLP 2021}, pages 4693--4703, Online. Association for Computational Linguistics.

\bibitem[{Hoffmann et~al.(2022)Hoffmann, Borgeaud, Mensch, Buchatskaya, Cai, Rutherford, de~Las~Casas, Hendricks, Welbl, Clark, Hennigan, Noland, Millican, van~den Driessche, Damoc, Guy, Osindero, Simonyan, Elsen, Rae, Vinyals, and Sifre}]{hoffmann2022training}
Jordan Hoffmann, Sebastian Borgeaud, Arthur Mensch, Elena Buchatskaya, Trevor Cai, Eliza Rutherford, Diego de~Las~Casas, Lisa~Anne Hendricks, Johannes Welbl, Aidan Clark, Tom Hennigan, Eric Noland, Katie Millican, George van~den Driessche, Bogdan Damoc, Aurelia Guy, Simon Osindero, Karen Simonyan, Erich Elsen, Jack~W. Rae, Oriol Vinyals, and Laurent Sifre. 2022.
\newblock \href {http://arxiv.org/abs/2203.15556} {Training compute-optimal large language models}.

\bibitem[{Houlsby et~al.(2019)Houlsby, Giurgiu, Jastrzebski, Morrone, De~Laroussilhe, Gesmundo, Attariyan, and Gelly}]{houlsby2019parameter}
Neil Houlsby, Andrei Giurgiu, Stanislaw Jastrzebski, Bruna Morrone, Quentin De~Laroussilhe, Andrea Gesmundo, Mona Attariyan, and Sylvain Gelly. 2019.
\newblock \href {https://proceedings.mlr.press/v97/houlsby19a.html} {Parameter-efficient transfer learning for {NLP}}.
\newblock In \emph{Proceedings of the 36th International Conference on Machine Learning}, volume~97 of \emph{Proceedings of Machine Learning Research}, pages 2790--2799. PMLR.

\bibitem[{Hu et~al.(2022)Hu, yelong shen, Wallis, Allen-Zhu, Li, Wang, Wang, and Chen}]{hu2022lora}
Edward~J Hu, yelong shen, Phillip Wallis, Zeyuan Allen-Zhu, Yuanzhi Li, Shean Wang, Lu~Wang, and Weizhu Chen. 2022.
\newblock \href {https://openreview.net/forum?id=nZeVKeeFYf9} {Lo{RA}: Low-rank adaptation of large language models}.
\newblock In \emph{International Conference on Learning Representations}.

\bibitem[{Jiang et~al.(2023)Jiang, Sablayrolles, Mensch, Bamford, Chaplot, de~las Casas, Bressand, Lengyel, Lample, Saulnier, Lavaud, Lachaux, Stock, Scao, Lavril, Wang, Lacroix, and Sayed}]{jiang2023mistral}
Albert~Q. Jiang, Alexandre Sablayrolles, Arthur Mensch, Chris Bamford, Devendra~Singh Chaplot, Diego de~las Casas, Florian Bressand, Gianna Lengyel, Guillaume Lample, Lucile Saulnier, Lélio~Renard Lavaud, Marie-Anne Lachaux, Pierre Stock, Teven~Le Scao, Thibaut Lavril, Thomas Wang, Timothée Lacroix, and William~El Sayed. 2023.
\newblock \href {http://arxiv.org/abs/2310.06825} {Mistral 7b}.

\bibitem[{Lewis et~al.(2020)Lewis, Oguz, Rinott, Riedel, and Schwenk}]{lewis2020mlqa}
Patrick Lewis, Barlas Oguz, Ruty Rinott, Sebastian Riedel, and Holger Schwenk. 2020.
\newblock Mlqa: Evaluating cross-lingual extractive question answering.
\newblock In \emph{Proceedings of the 58th Annual Meeting of the Association for Computational Linguistics}, pages 7315--7330.

\bibitem[{Li et~al.(2023{\natexlab{a}})Li, Koto, Wu, Aji, and Baldwin}]{li2023bactrianx}
Haonan Li, Fajri Koto, Minghao Wu, Alham~Fikri Aji, and Timothy Baldwin. 2023{\natexlab{a}}.
\newblock \href {http://arxiv.org/abs/2305.15011} {Bactrian-x : A multilingual replicable instruction-following model with low-rank adaptation}.

\bibitem[{Li and Liang(2021)}]{li2021prefixtuning}
Xiang~Lisa Li and Percy Liang. 2021.
\newblock \href {http://arxiv.org/abs/2101.00190} {Prefix-tuning: Optimizing continuous prompts for generation}.

\bibitem[{Li et~al.(2023{\natexlab{b}})Li, Zhang, Dubois, Taori, Gulrajani, Guestrin, Liang, and Hashimoto}]{li2023alpacaeval}
Xuechen Li, Tianyi Zhang, Yann Dubois, Rohan Taori, Ishaan Gulrajani, Carlos Guestrin, Percy Liang, and Tatsunori~B. Hashimoto. 2023{\natexlab{b}}.
\newblock \href {https://github.com/tatsu-lab/alpaca_eval} {{AlpacaEval: An Automatic Evaluator of Instruction-following Models}}.

\bibitem[{Liu et~al.(2022{\natexlab{a}})Liu, Tam, Muqeeth, Mohta, Huang, Bansal, and Raffel}]{liu2022fewshot}
Haokun Liu, Derek Tam, Mohammed Muqeeth, Jay Mohta, Tenghao Huang, Mohit Bansal, and Colin Raffel. 2022{\natexlab{a}}.
\newblock \href {http://arxiv.org/abs/2205.05638} {Few-shot parameter-efficient fine-tuning is better and cheaper than in-context learning}.

\bibitem[{Liu et~al.(2022{\natexlab{b}})Liu, Ji, Fu, Tam, Du, Yang, and Tang}]{liu-etal-2022-p}
Xiao Liu, Kaixuan Ji, Yicheng Fu, Weng Tam, Zhengxiao Du, Zhilin Yang, and Jie Tang. 2022{\natexlab{b}}.
\newblock \href {https://doi.org/10.18653/v1/2022.acl-short.8} {{P}-tuning: Prompt tuning can be comparable to fine-tuning across scales and tasks}.
\newblock In \emph{Proceedings of the 60th Annual Meeting of the Association for Computational Linguistics (Volume 2: Short Papers)}, pages 61--68, Dublin, Ireland. Association for Computational Linguistics.

\bibitem[{Loshchilov and Hutter(2019)}]{loshchilov2018decoupled}
Ilya Loshchilov and Frank Hutter. 2019.
\newblock \href {https://openreview.net/forum?id=Bkg6RiCqY7} {Decoupled weight decay regularization}.
\newblock In \emph{International Conference on Learning Representations}.

\bibitem[{Mitra et~al.(2023)Mitra, Corro, Mahajan, Codas, Simoes, Agarwal, Chen, Razdaibiedina, Jones, Aggarwal, Palangi, Zheng, Rosset, Khanpour, and Awadallah}]{mitra2023orca}
Arindam Mitra, Luciano~Del Corro, Shweti Mahajan, Andres Codas, Clarisse Simoes, Sahaj Agarwal, Xuxi Chen, Anastasia Razdaibiedina, Erik Jones, Kriti Aggarwal, Hamid Palangi, Guoqing Zheng, Corby Rosset, Hamed Khanpour, and Ahmed Awadallah. 2023.
\newblock \href {http://arxiv.org/abs/2311.11045} {Orca 2: Teaching small language models how to reason}.

\bibitem[{Mukherjee et~al.(2023)Mukherjee, Mitra, Jawahar, Agarwal, Palangi, and Awadallah}]{mukherjee2023orca}
Subhabrata Mukherjee, Arindam Mitra, Ganesh Jawahar, Sahaj Agarwal, Hamid Palangi, and Ahmed Awadallah. 2023.
\newblock \href {http://arxiv.org/abs/2306.02707} {Orca: Progressive learning from complex explanation traces of gpt-4}.

\bibitem[{OpenAI(2023)}]{gpt4techreport}
OpenAI. 2023.
\newblock \href {https://arxiv.org/pdf/2303.08774.pdf} {Gpt4 technical report}.

\bibitem[{Pfeiffer et~al.(2022)Pfeiffer, Goyal, Lin, Li, Cross, Riedel, and Artetxe}]{pfeiffer-etal-2022-lifting}
Jonas Pfeiffer, Naman Goyal, Xi~Lin, Xian Li, James Cross, Sebastian Riedel, and Mikel Artetxe. 2022.
\newblock \href {https://doi.org/10.18653/v1/2022.naacl-main.255} {Lifting the curse of multilinguality by pre-training modular transformers}.
\newblock In \emph{Proceedings of the 2022 Conference of the North American Chapter of the Association for Computational Linguistics: Human Language Technologies}, pages 3479--3495, Seattle, United States. Association for Computational Linguistics.

\bibitem[{Pfeiffer et~al.(2021)Pfeiffer, Kamath, R{\"u}ckl{\'e}, Cho, and Gurevych}]{pfeiffer-etal-2021-adapterfusion}
Jonas Pfeiffer, Aishwarya Kamath, Andreas R{\"u}ckl{\'e}, Kyunghyun Cho, and Iryna Gurevych. 2021.
\newblock \href {https://doi.org/10.18653/v1/2021.eacl-main.39} {{A}dapter{F}usion: Non-destructive task composition for transfer learning}.
\newblock In \emph{Proceedings of the 16th Conference of the European Chapter of the Association for Computational Linguistics: Main Volume}, pages 487--503, Online. Association for Computational Linguistics.

\bibitem[{Pfeiffer et~al.(2020{\natexlab{a}})Pfeiffer, R{\"u}ckl{\'e}, Poth, Kamath, Vuli{\'c}, Ruder, Cho, and Gurevych}]{pfeiffer-etal-2020-adapterhub}
Jonas Pfeiffer, Andreas R{\"u}ckl{\'e}, Clifton Poth, Aishwarya Kamath, Ivan Vuli{\'c}, Sebastian Ruder, Kyunghyun Cho, and Iryna Gurevych. 2020{\natexlab{a}}.
\newblock \href {https://doi.org/10.18653/v1/2020.emnlp-demos.7} {{A}dapter{H}ub: A framework for adapting transformers}.
\newblock In \emph{Proceedings of the 2020 Conference on Empirical Methods in Natural Language Processing: System Demonstrations}, pages 46--54, Online. Association for Computational Linguistics.

\bibitem[{Pfeiffer et~al.(2020{\natexlab{b}})Pfeiffer, Vuli{\'c}, Gurevych, and Ruder}]{pfeiffer-etal-2020-mad}
Jonas Pfeiffer, Ivan Vuli{\'c}, Iryna Gurevych, and Sebastian Ruder. 2020{\natexlab{b}}.
\newblock \href {https://doi.org/10.18653/v1/2020.emnlp-main.617} {{MAD-X}: {A}n {A}dapter-{B}ased {F}ramework for {M}ulti-{T}ask {C}ross-{L}ingual {T}ransfer}.
\newblock In \emph{Proceedings of the 2020 Conference on Empirical Methods in Natural Language Processing (EMNLP)}, pages 7654--7673, Online. Association for Computational Linguistics.

\bibitem[{Ponti et~al.(2020)Ponti, Glava{\v{s}}, Majewska, Liu, Vuli{\'c}, and Korhonen}]{ponti2020xcopa}
Edoardo~Maria Ponti, Goran Glava{\v{s}}, Olga Majewska, Qianchu Liu, Ivan Vuli{\'c}, and Anna Korhonen. 2020.
\newblock Xcopa: A multilingual dataset for causal commonsense reasoning.
\newblock In \emph{Proceedings of the 2020 Conference on Empirical Methods in Natural Language Processing (EMNLP)}, pages 2362--2376.

\bibitem[{Ramesh et~al.(2023)Ramesh, Chavan, Pandit, and Sitaram}]{ramesh-etal-2023-comparative}
Krithika Ramesh, Arnav Chavan, Shrey Pandit, and Sunayana Sitaram. 2023.
\newblock \href {https://doi.org/10.18653/v1/2023.acl-long.878} {A comparative study on the impact of model compression techniques on fairness in language models}.
\newblock In \emph{Proceedings of the 61st Annual Meeting of the Association for Computational Linguistics (Volume 1: Long Papers)}, pages 15762--15782, Toronto, Canada. Association for Computational Linguistics.

\bibitem[{Roemmele et~al.(2011)Roemmele, Bejan, and Gordon}]{roemmele2011choice}
Melissa Roemmele, Cosmin~Adrian Bejan, and Andrew~S Gordon. 2011.
\newblock Choice of plausible alternatives: An evaluation of commonsense causal reasoning.
\newblock In \emph{AAAI spring symposium: logical formalizations of commonsense reasoning}, pages 90--95.

\bibitem[{Shaham et~al.(2024)Shaham, Herzig, Aharoni, Szpektor, Tsarfaty, and Eyal}]{shaham2024multilingual}
Uri Shaham, Jonathan Herzig, Roee Aharoni, Idan Szpektor, Reut Tsarfaty, and Matan Eyal. 2024.
\newblock \href {http://arxiv.org/abs/2401.01854} {Multilingual instruction tuning with just a pinch of multilinguality}.

\bibitem[{Shi et~al.(2023)Shi, Suzgun, Freitag, Wang, Srivats, Vosoughi, Chung, Tay, Ruder, Zhou, Das, and Wei}]{shi2023language}
Freda Shi, Mirac Suzgun, Markus Freitag, Xuezhi Wang, Suraj Srivats, Soroush Vosoughi, Hyung~Won Chung, Yi~Tay, Sebastian Ruder, Denny Zhou, Dipanjan Das, and Jason Wei. 2023.
\newblock \href {https://openreview.net/forum?id=fR3wGCk-IXp} {Language models are multilingual chain-of-thought reasoners}.
\newblock In \emph{The Eleventh International Conference on Learning Representations}.

\bibitem[{Singh et~al.(2024)Singh, Vargus, Dsouza, Karlsson, Mahendiran, Ko, Shandilya, Patel, Mataciunas, OMahony, Zhang, Hettiarachchi, Wilson, Machado, Moura, Krzemiński, Fadaei, Ergün, Okoh, Alaagib, Mudannayake, Alyafeai, Chien, Ruder, Guthikonda, Alghamdi, Gehrmann, Muennighoff, Bartolo, Kreutzer, Üstün, Fadaee, and Hooker}]{singh2024aya}
Shivalika Singh, Freddie Vargus, Daniel Dsouza, Börje~F. Karlsson, Abinaya Mahendiran, Wei-Yin Ko, Herumb Shandilya, Jay Patel, Deividas Mataciunas, Laura OMahony, Mike Zhang, Ramith Hettiarachchi, Joseph Wilson, Marina Machado, Luisa~Souza Moura, Dominik Krzemiński, Hakimeh Fadaei, Irem Ergün, Ifeoma Okoh, Aisha Alaagib, Oshan Mudannayake, Zaid Alyafeai, Vu~Minh Chien, Sebastian Ruder, Surya Guthikonda, Emad~A. Alghamdi, Sebastian Gehrmann, Niklas Muennighoff, Max Bartolo, Julia Kreutzer, Ahmet Üstün, Marzieh Fadaee, and Sara Hooker. 2024.
\newblock \href {http://arxiv.org/abs/2402.06619} {Aya dataset: An open-access collection for multilingual instruction tuning}.

\bibitem[{Taori et~al.(2023)Taori, Gulrajani, Zhang, Dubois, Li, Guestrin, Liang, and Hashimoto}]{alpaca}
Rohan Taori, Ishaan Gulrajani, Tianyi Zhang, Yann Dubois, Xuechen Li, Carlos Guestrin, Percy Liang, and Tatsunori~B. Hashimoto. 2023.
\newblock Stanford alpaca: An instruction-following llama model.
\newblock \url{https://github.com/tatsu-lab/stanford_alpaca}.

\bibitem[{Team et~al.(2022)Team, Costa-jussà, Cross, Çelebi, Elbayad, Heafield, Heffernan, Kalbassi, Lam, Licht, Maillard, Sun, Wang, Wenzek, Youngblood, Akula, Barrault, Gonzalez, Hansanti, Hoffman, Jarrett, Sadagopan, Rowe, Spruit, Tran, Andrews, Ayan, Bhosale, Edunov, Fan, Gao, Goswami, Guzmán, Koehn, Mourachko, Ropers, Saleem, Schwenk, and Wang}]{nllbteam2022language}
NLLB Team, Marta~R. Costa-jussà, James Cross, Onur Çelebi, Maha Elbayad, Kenneth Heafield, Kevin Heffernan, Elahe Kalbassi, Janice Lam, Daniel Licht, Jean Maillard, Anna Sun, Skyler Wang, Guillaume Wenzek, Al~Youngblood, Bapi Akula, Loic Barrault, Gabriel~Mejia Gonzalez, Prangthip Hansanti, John Hoffman, Semarley Jarrett, Kaushik~Ram Sadagopan, Dirk Rowe, Shannon Spruit, Chau Tran, Pierre Andrews, Necip~Fazil Ayan, Shruti Bhosale, Sergey Edunov, Angela Fan, Cynthia Gao, Vedanuj Goswami, Francisco Guzmán, Philipp Koehn, Alexandre Mourachko, Christophe Ropers, Safiyyah Saleem, Holger Schwenk, and Jeff Wang. 2022.
\newblock \href {http://arxiv.org/abs/2207.04672} {No language left behind: Scaling human-centered machine translation}.

\bibitem[{Touvron et~al.(2023)Touvron, Martin, Stone, Albert, Almahairi, Babaei, Bashlykov, Batra, Bhargava, Bhosale et~al.}]{touvron2023llama}
Hugo Touvron, Louis Martin, Kevin Stone, Peter Albert, Amjad Almahairi, Yasmine Babaei, Nikolay Bashlykov, Soumya Batra, Prajjwal Bhargava, Shruti Bhosale, et~al. 2023.
\newblock Llama 2: Open foundation and fine-tuned chat models.
\newblock \emph{arXiv preprint arXiv:2307.09288}.

\bibitem[{Vu et~al.(2022)Vu, Barua, Lester, Cer, Iyyer, and Constant}]{vu-etal-2022-overcoming}
Tu~Vu, Aditya Barua, Brian Lester, Daniel Cer, Mohit Iyyer, and Noah Constant. 2022.
\newblock \href {https://doi.org/10.18653/v1/2022.emnlp-main.630} {Overcoming catastrophic forgetting in zero-shot cross-lingual generation}.
\newblock In \emph{Proceedings of the 2022 Conference on Empirical Methods in Natural Language Processing}, pages 9279--9300, Abu Dhabi, United Arab Emirates. Association for Computational Linguistics.

\bibitem[{Wei et~al.(2022)Wei, Wang, Schuurmans, Bosma, brian ichter, Xia, Chi, Le, and Zhou}]{wei2022chain}
Jason Wei, Xuezhi Wang, Dale Schuurmans, Maarten Bosma, brian ichter, Fei Xia, Ed~H. Chi, Quoc~V Le, and Denny Zhou. 2022.
\newblock \href {https://openreview.net/forum?id=_VjQlMeSB_J} {Chain of thought prompting elicits reasoning in large language models}.
\newblock In \emph{Advances in Neural Information Processing Systems}.

\bibitem[{Wei et~al.(2023)Wei, Wei, Lin, Li, Zhang, Ren, Li, Wan, Cao, Xie, Hu, Li, Hui, Yu, Liu, Yang, Huang, and Xie}]{wei2023polylm}
Xiangpeng Wei, Haoran Wei, Huan Lin, Tianhao Li, Pei Zhang, Xingzhang Ren, Mei Li, Yu~Wan, Zhiwei Cao, Binbin Xie, Tianxiang Hu, Shangjie Li, Binyuan Hui, Bowen Yu, Dayiheng Liu, Baosong Yang, Fei Huang, and Jun Xie. 2023.
\newblock \href {http://arxiv.org/abs/2307.06018} {Polylm: An open source polyglot large language model}.

\bibitem[{Wu et~al.(2020)Wu, Judd, Zhang, Isaev, and Micikevicius}]{Wu2020IntegerQF}
Hao Wu, Patrick Judd, Xiaojie Zhang, Mikhail Isaev, and Paulius Micikevicius. 2020.
\newblock \href {https://api.semanticscholar.org/CorpusID:216035831} {Integer quantization for deep learning inference: Principles and empirical evaluation}.
\newblock \emph{ArXiv}, abs/2004.09602.

\bibitem[{yang Liu et~al.(2023)yang Liu, Liu, Huang, Dong, and Cheng}]{liu2023llmfp4}
Shih yang Liu, Zechun Liu, Xijie Huang, Pingcheng Dong, and Kwang-Ting Cheng. 2023.
\newblock \href {http://arxiv.org/abs/2310.16836} {Llm-fp4: 4-bit floating-point quantized transformers}.

\bibitem[{Zhao et~al.(2024)Zhao, Zhang, Zhang, Gui, and Huang}]{zhao2024llama}
Jun Zhao, Zhihao Zhang, Qi~Zhang, Tao Gui, and Xuanjing Huang. 2024.
\newblock \href {http://arxiv.org/abs/2401.01055} {Llama beyond english: An empirical study on language capability transfer}.

\bibitem[{Zhuo et~al.(2024)Zhuo, Zebaze, Suppattarachai, von Werra, de~Vries, Liu, and Muennighoff}]{zhuo2024astraios}
Terry~Yue Zhuo, Armel Zebaze, Nitchakarn Suppattarachai, Leandro von Werra, Harm de~Vries, Qian Liu, and Niklas Muennighoff. 2024.
\newblock \href {http://arxiv.org/abs/2401.00788} {Astraios: Parameter-efficient instruction tuning code large language models}.

\end{thebibliography}
\bibliographystyle{acl_natbib}

\appendix

% \section{Hyper Parameters Setting}

\section{Hyperparameters for Finetuning and Training Setup}

Our code for finetuning is based on the open source \texttt{axolotl}\footnote{\url{https://github.com/OpenAccess-AI-Collective/axolotl}} framework. We plan to release our configuration files for better reproducibility. Each finetuning experiment took  $\sim$16-24 hours to complete on a single NVIDIA A100 GPU with 80 GB RAM. Exact hyperparameters for finetuning are mentioned below:

\begin{table}[h]
\centering 
\small
%\begin{tabular}{@{}ll@{}}
\begin{tabularx}{0.45\textwidth}{lX}
\toprule
\textbf{Hyperparameter}    & \textbf{Value} \\ \midrule
Learning rate              & $1\times 10^{-6}$ \\
Epochs                     & $5$ \\ 
Global batch size          & $16$ \\
Scheduler                  & Cosine \\
Warmup                     & Linear \\
Warmup steps               & 10 \\
Optimizer                  & AdamW \tiny{\citep{loshchilov2018decoupled}}\\
%Warmup                     & $0$              \\
Weight decay               & $0$              \\
% Adapter architecture       & Pfeiffer       \\
% Adapter activation         & SiLU \cite{elfwing2017sigmoidweighted}          \\
% Adapter reduction factor   & $16$             \\
% FP16                       & True           \\
% MLM probability            & $0.15$          \\ 
\bottomrule
\end{tabularx}
\caption{Hyperparameters for finetuning.}
\label{tab:indiv_ft}
\end{table}

\section{Evaluation Dataset Details}
\label{sec:evaluation dataset details}
The detailed description of the datasets that we use for evaluation are as follows:

\paragraph{XNLI:} The XNLI (Cross-lingual Natural Language Inference) dataset  \cite{Conneau2018xnli} is an extension of the Multi-Genre NLI (MultiNLI) corpus to 15 languages. The dataset was created by manually translating the validation and test sets of MultiNLI into each of those 15 languages. The English training set was machine translated for all languages. The dataset is composed of 122k train, 2490 validation, and 5010 test examples. XNLI provides a robust platform for evaluating cross-lingual sentence understanding methods. %We Evaluated our models in 4 shot setting.
We evaluated our models on the test split with 4 in-context examples sampled from the validation split.
We report our results in Table \ref{tab:results_xnli_llama_alpaca}, \ref{tab:results_xnli_llama_bactrian}, \ref{tab:results_xnli_llama_multialpaca}, \ref{tab:results_xnli_mistral_alpaca}, \ref{tab:results_xnli_mistral_bactrian} and \ref{tab:results_xnli_mistral_multialpaca}.

\paragraph{XCOPA:} The XCOPA (Cross-lingual Choice of Plausible Alternatives) dataset \cite{ponti2020xcopa} is a benchmark for evaluating the ability of machine learning models to transfer commonsense reasoning across languages. It is a translation and re-annotation of the English COPA dataset \cite{roemmele2011choice} and covers 11 languages from 11 families and several areas around the globe. The dataset is challenging as it requires both the command of world knowledge and the ability to generalize to new languages.
We evaluated our models on Estonian, Thai, Italian, Indonesian, Vietnamese and Southern Quechua. We evaluated our models in the 4-shot setting similar to XNLI. We report our results in Table \ref{tab:results_xcopa_llama_alpaca}, \ref{tab:results_xcopa_llama_bactrian}, \ref{tab:results_xcopa_llama_multialpaca}, \ref{tab:results_xcopa_mistral_alpaca}, \ref{tab:results_xcopa_mistral_bactrian} and \ref{tab:results_xcopa_mistral_multialpaca}.

\paragraph{Belebele:} Belebele \cite{bandarkar2023belebele} is a multiple choice machine reading comprehension (MRC) dataset parallel across 122 languages. Each question is linked to a short passage from the FLORES-200 dataset \cite{nllbteam2022language}. The human annotation procedure was carefully curated to create questions that discriminate between different levels of language comprehension.
% We evaluated German, English, Spanish, French, Italian, Japanese, Portuguese, Chinese Simplified. %We evaluated our finetuned models for 50\% of the test set for interest of time. 
We evaluated our models in the zero-shot setting and report results in Table Table \ref{tab:results_belebele_llama_alpaca},
\ref{tab:results_belebele_mistral_alpaca}
\ref{tab:results_belebele_llama_multialpaca},
\ref{tab:results_belebele_mistral_multialpaca}, \ref{tab:results_belebele_llama_bactrian} and \ref{tab:results_belebele_mistral_bactrian}.

\paragraph{MLQA:} MLQA \cite{lewis2020mlqa} is a multilingual question answering dataset designed for cross lingual question answering. It contains 5K extractive question answering instances. It consists of 7 languages i.e. English, Arabic, Vietnamese, German, Spanish, Hindi and Simplified Chinese. %We Evaluated our models in 4 shot setting.
The evaluation uses a 4-shot setting similar to that of XNLI. We report our results in Table \ref{tab:results_xquad_llama_alpaca}, \ref{tab:results_xquad_llama_bactrian},
\ref{tab:results_xquad_llama_multialpaca}, \ref{tab:results_xquad_mistral_alpaca}, \ref{tab:results_xquad_mistral_bactrian} and \ref{tab:results_xquad_mistral_multialpaca}.

\paragraph{XQuAD:} The XQuAD (Cross-lingual Question Answering Dataset) \cite{artetxe2020cross} is a benchmark dataset for evaluating cross-lingual question answering performance. It consists of a subset of 240 paragraphs and 1190 question-answer pairs from the development set of SQuAD v1.1, along with their professional translations into ten languages: Spanish, German, Greek, Russian, Turkish, Arabic, Vietnamese, Thai, Chinese, and Hindi. As a result, the dataset is entirely parallel across 11 languages. This dataset provides a robust platform for developing and evaluating models on cross-lingual question answering tasks. %We evaluated our finetuned models for 10\% of the test set for interest of time. We Evaluated our models in 4 shot setting.
For evaluation, we use a 4-shot setting similar to MLQA. We report our results in table \ref{tab:results_xquad_llama_alpaca}, \ref{tab:results_xquad_llama_bactrian},
\ref{tab:results_xquad_llama_multialpaca}, \ref{tab:results_xquad_mistral_alpaca}, \ref{tab:results_xquad_mistral_bactrian} and \ref{tab:results_xquad_mistral_multialpaca}.

\paragraph{XLSUM: } XLSUM \cite{hasan-etal-2021-xl} is a comprehensive and diverse dataset for abstractive summarization comprising 1 million human annotated article-summary pairs from BBC. The dataset covers 44 languages ranging from low to high-resource, for many of which no public dataset is currently available. We evaluate our models on a subset of 7 languages, namely, Arabic, Chinese-Simplified, English, Hindi, French, Japanese and Spanish in a zero-shot setting. We present our results in Table \ref{tab:results_xlsum_llama_alpaca}, \ref{tab:results_xlsum_llama_bactrian}, \ref{tab:results_xlsum_llama_multialpaca}, \ref{tab:results_xlsum_mistral_alpaca},  \ref{tab:results_xlsum_mistral_multialpaca} and \ref{tab:results_xlsum_mistral_bactrian}.

\paragraph{AlpacaEval:} AlpacaEval \cite{li2023alpacaeval} is an LLM based automatic evaluator for instruction following models. It consists of around 800 instructions and corresponding responses obtained from (text-davinci-003) GPT3. The benchmark compares responses from GPT3 (or any other ``oracle'' model) with target (finetuned) model using another LLM (typically GPT4) as an evaluator. The evaluator LLM decides which response is better and overall win rate (higher the better) is computed for the target model. For our evaluation, we use the text-davinci-003 responses from the dataset as our oracle/gold responses and use GPT4 (gpt-4-32k) as our evaluator. We report our results in Table \ref{tab:results_alpaca_eval_llama_alpaca}, \ref{tab:results_alpaca_eval_llama_bactrian}, \ref{tab:results_alpaca_eval_llama_multialpaca}, \ref{tab:results_alpaca_eval_mistral_alpaca}, \ref{tab:results_alpaca_eval_mistral_bactrian} and \ref{tab:results_alpaca_eval_mistral_multialpaca}.

\section{Evaluation Prompts}
\label{sec:eval prompts}

For XNLI, XCOPA, Belebele, MLQA, XQUAD, XLSUM we use the standard \textsc{Alpaca} system prompt \textbf{"Below is an instruction that describes a task, paired with an input that provides further context. Write a response that appropriately completes the request."}.

\subsection{XNLI}

\begin{figure}[!h]
\centering
\begin{promptbox}
% \justify
% Below is an instruction that describes a task, paired with an input that provides further context. Write a response that appropriately completes the request. 
%\\ \\
\#\#\# Instruction:
\\
The task is to solve Natural Language Inference (NLI) problems. NLI is the task of determining the inference relation between two (short, ordered) texts: entailment, contradiction, or neutral. Answer as concisely as possible in the same format as the examples below:
\\ %\\
\{\{premise\}\}
\\ %\\
Question: \{\{hypothesis\}\} True, False, or Neither?
\\ %\\
\#\#\# Response

\end{promptbox}
\caption{XNLI Prompt}
\label{fig:XNLI prompt}
\end{figure}

\subsection{XCOPA}

\begin{figure}[!h]
\centering
\begin{promptbox}
% \justify
% Below is an instruction that describes a task, paired with an input that provides further context. Write a response that appropriately completes the request.
% \\ \\
\#\#\# Instruction:
\\
The task is to perform open-domain commonsense causal reasoning. You will be provided a premise and two alternatives, where the task is to select the alternative that more plausibly has a causal relation with the premise. Answer as concisely as possible in the same format as the examples below:
\\ %\\
Given this premise: 
\\
\{\{premise\}\}
\\ %\\
What's the best option?
\\
-choice1 : \{\{choice1\}\}
\\
-choice2 : \{\{choice2\}\}
\\% \\
We are looking for \{\% if question == "cause" \%\} a cause \{\% else \%\} an effect \{\% endif \%\}
\\ %\\
\#\#\# Response:
\\
\end{promptbox}
\caption{XCOPA Prompt}
\label{fig:XCOPA prompt}
\end{figure}

\subsection{Belebele}

\begin{figure}[!h]
\centering
\begin{promptbox}
% \justify
% Below is an instruction that describes a task, paired with an input that provides further context. Write a response that appropriately completes the request.
% \\ \\
\#\#\# Instruction:
\\
The task is to perform reading comprehension task. Given the following passage, query, and answer choices, output the letter corresponding to the correct answer.
\\ \\
Passage: \{\{flores\_passage\}\}
\\
Query: \{\{question\}\}
\\
Choices:\\
A: \{\{mc\_answer1\}\}
\\
B: \{\{mc\_answer2\}\}
\\
C: \{\{mc\_answer3\}\}
\\
D: \{\{mc\_answer4\}\}
\\ \\
\#\#\# Response:
\\
\end{promptbox}
\caption{Belebele Prompt}
\label{fig:Belebele prompt}
\end{figure}

\subsection{MLQA}

\begin{figure}[!h]
\centering
\begin{promptbox}
% \justify
\#\#\# Instruction:
\\
The task is to solve reading comprehension problems. You will be provided questions on a set of passages and you will need to provide the answer as it appears in the passage. The answer should be in the same language as the question and the passage.
\\ \\
Context:\{\{context\}\}
\\
Question:\{\{question\}\}
\\ \\
Referring to the passage above, the correct answer to the given question is
\\ \\
\#\#\# Response:
\\
\end{promptbox}
\caption{MLQA Prompt}
\label{fig:MLQA prompt}
\end{figure}

\subsection{XQUAD}

\begin{figure}[!h]
\centering
\begin{promptbox}
% \justify
\#\#\# Instruction:
\\
The task is to solve reading comprehension problems. You will be provided questions on a set of passages and you will need to provide the answer as it appears in the passage. The answer should be in the same language as the question and the passage.
\\ \\
Context:\{\{context\}\}
\\
Question:\{\{question\}\}
\\ \\
Referring to the passage above, the correct answer to the given question is
\\ \\
\#\#\# Response:
\\
\end{promptbox}
\caption{XQUAD Prompt}
\label{fig:XQUAD prompt}
\end{figure}

\subsection{XLSUM}

\begin{figure}[!h]
\centering
\begin{promptbox}
% \justify
\#\#\# Instruction:
\\
The task is to summarize any given article. You should summarize all important information concisely in the same language in which you have been provided the document. Following the examples provided below:
\\ \\
\{\{text\}\}
\\ \\
\#\#\# Response:
\\
\end{promptbox}
\caption{XLSUM Prompt}
\label{fig:XLSUM prompt}
\end{figure}

\subsection{AlpacaEval}

\begin{figure}[!h]
\centering
\begin{promptbox}
% \justify
\noindent Below is an instruction that describes a task, paired with an input that provides further context. Write a response that appropriately completes the request.
\\ \\ 
\end{promptbox}
\caption{Alpaca Prompt}
\label{fig:Alpaca Eval prompt}
\end{figure}

\section{Further of Analysis of Results}
\subsection{Analysis of Rank and Quantisation}

\begin{figure*}
\centering
\includegraphics[width=\textwidth]{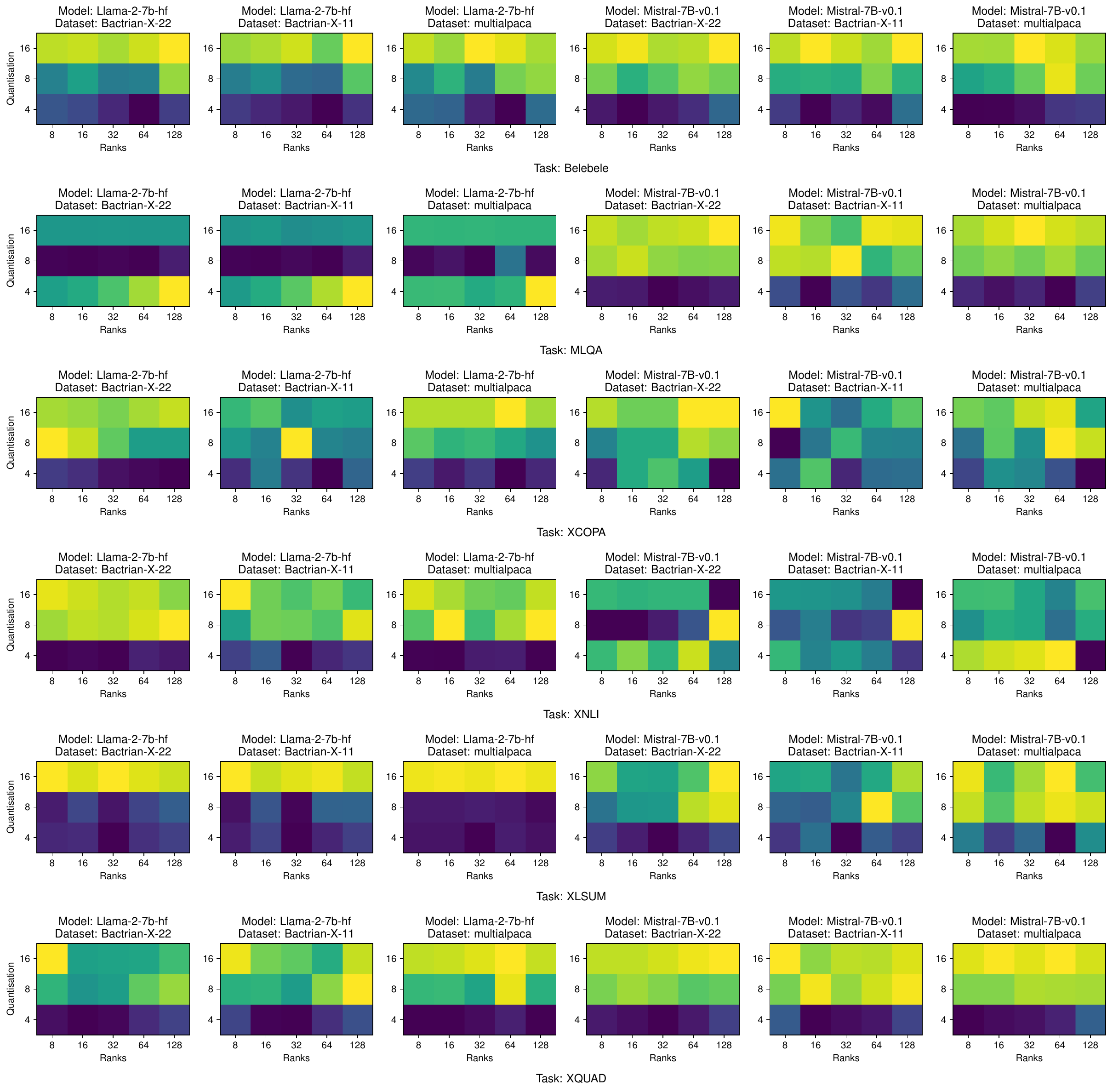}
    \caption{Task-wise performance of \mis\ and \llm\ fine-tuned on \textsc{Bactrian-X-22} and \malp\ averaged across languages on all rank-quantisation configurations.}
    \label{fig:all_rank_quant}
\end{figure*}

\subsection{Tasks Wise and Language Performance Plots}

\begin{figure*}
\centering
\includegraphics[width=\textwidth]{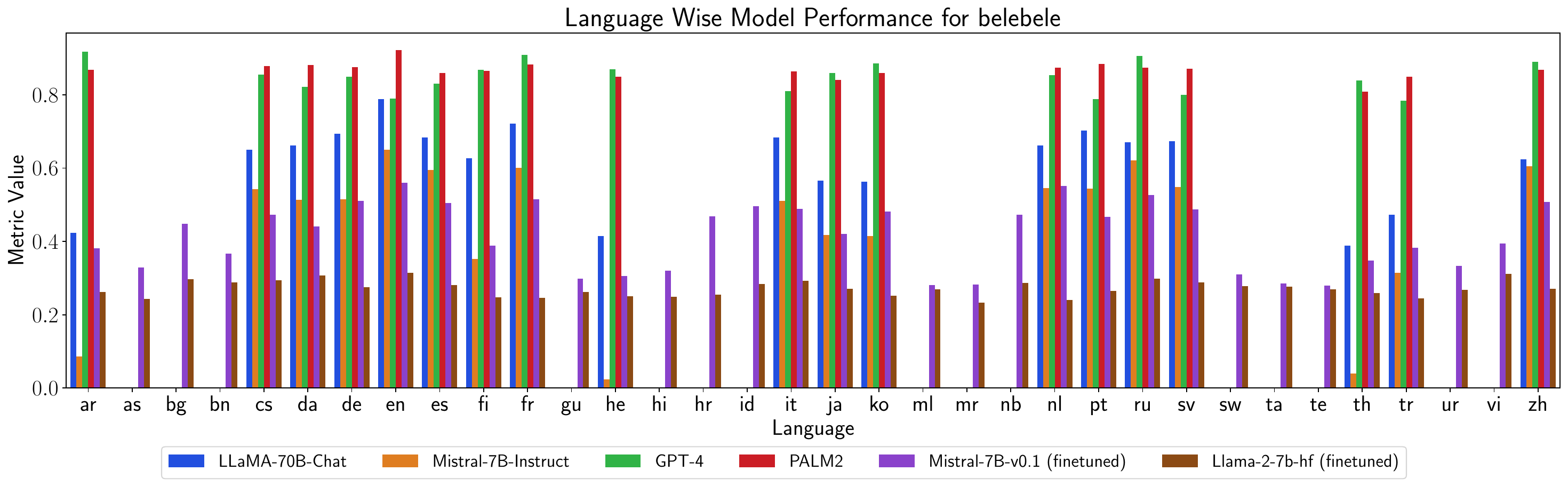}
    \caption{Detailed language-wise comparison of our fine-tuned \mis\ and \llm\ models with other baselines \cite{ahuja2023megaverse} on Afrikaans, Arabic, Assamese, Bulgarian, Bengali, Czech, Danish, German, English, Spanish, Finnish, French, Gujarat, Hebrew, Marathi, Norwegian, Dutch, Portuguese, Russian, Swedish, Swahili, Tamil, Telugu, Thai, Turkish, Urdu, Vietnamese and Chinese-Simplified   for Belebele \cite{bandarkar2023belebele}.}
    %\caption{Detailed language-wise comparison of our fine-tuned \mis\ and \llm\ models with other baselines \cite{ahuja2023megaverse} on Afrikaans, Arabic, Assamese, Bulgarian, Bengali, Czech and Danish for Belebele \cite{bandarkar2023belebele}.}
    \label{fig:langwise belebele 0}
\end{figure*}

\begin{figure*}
\centering
\includegraphics[width=\textwidth]{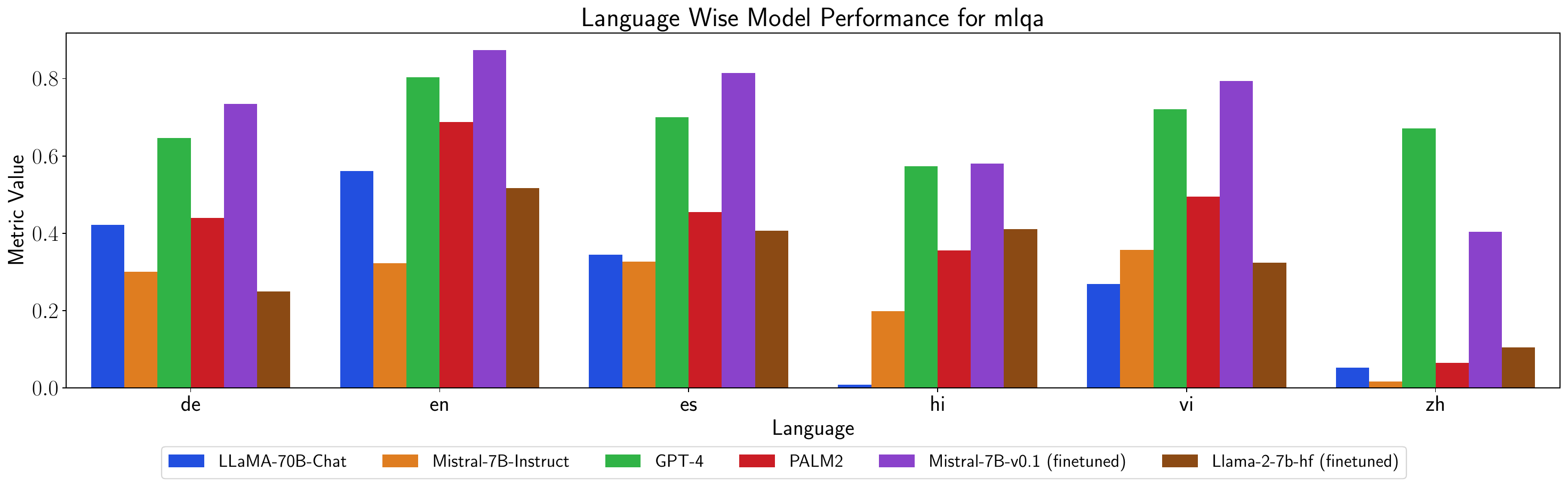}
    \caption{Detailed language-wise comparison of our fine-tuned \mis\ and \llm\ models with other baselines \cite{ahuja2023megaverse} on Arabic, German, English, Spanish, Hindi and Vietnamese for MLQA \cite{lewis2020mlqa}.}
    \label{fig:langwise mlqa}
\end{figure*}

\begin{figure*}
\centering
\includegraphics[width=\textwidth]{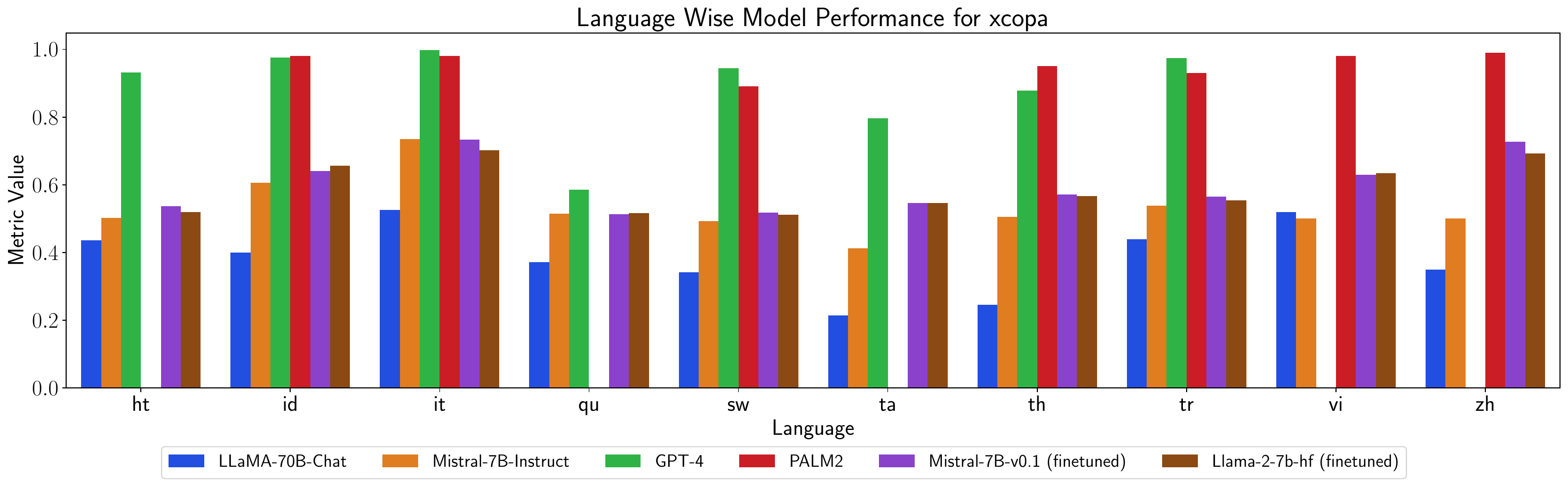}
    \caption{Detailed language-wise comparison of our fine-tuned \mis\ and \llm\ models with other baselines \cite{ahuja2023megaverse} on Estonian, Haitian, Indonesian, Italian, Quechua, Swahili, Tamil, Thai, Turkish and Vietnamese for XCOPA \cite{ponti2020xcopa}.}
    \label{fig:langwise xcopa}
\end{figure*}

\begin{figure*}
\centering
\includegraphics[width=\textwidth]{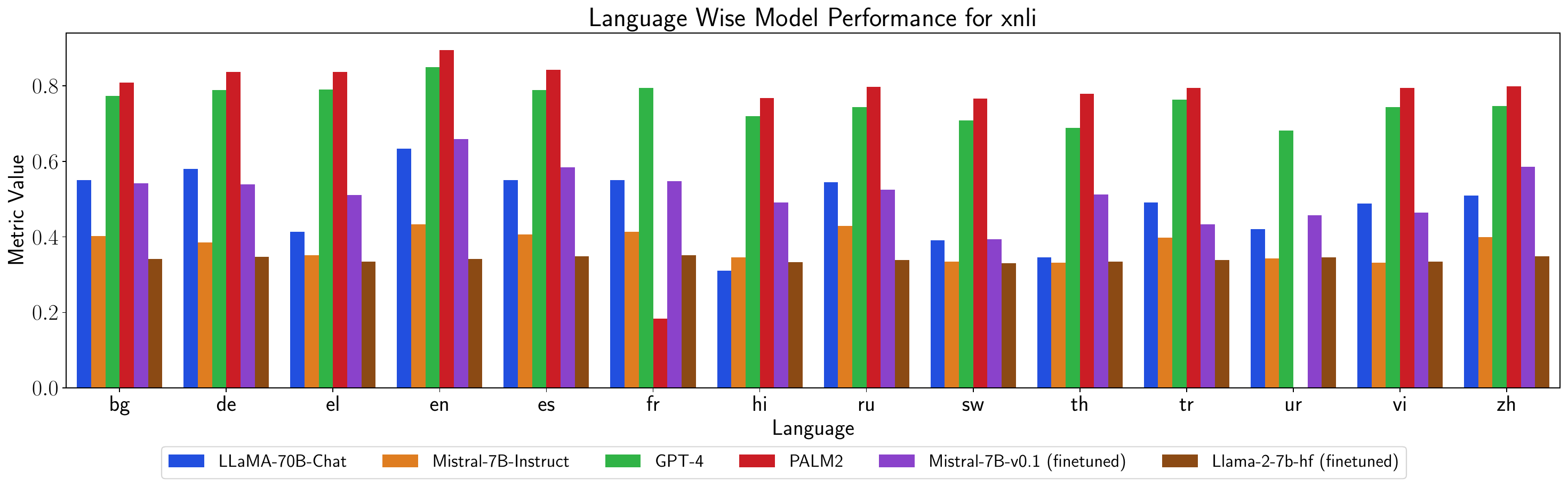}
    \caption{Detailed language-wise comparison of our fine-tuned \mis\ and \llm\ models with other baselines \cite{ahuja2023megaverse} on Arabic, Bulgarian, German, Greek, English, Spanish, French, Hindi, Russian, Swahili, Thai, Turkish, Urdu and Vietnamese for XNLI \cite{Conneau2018xnli}.}
    \label{fig:langwise xnli}
\end{figure*}

\begin{figure*}
\centering
\includegraphics[width=\textwidth]{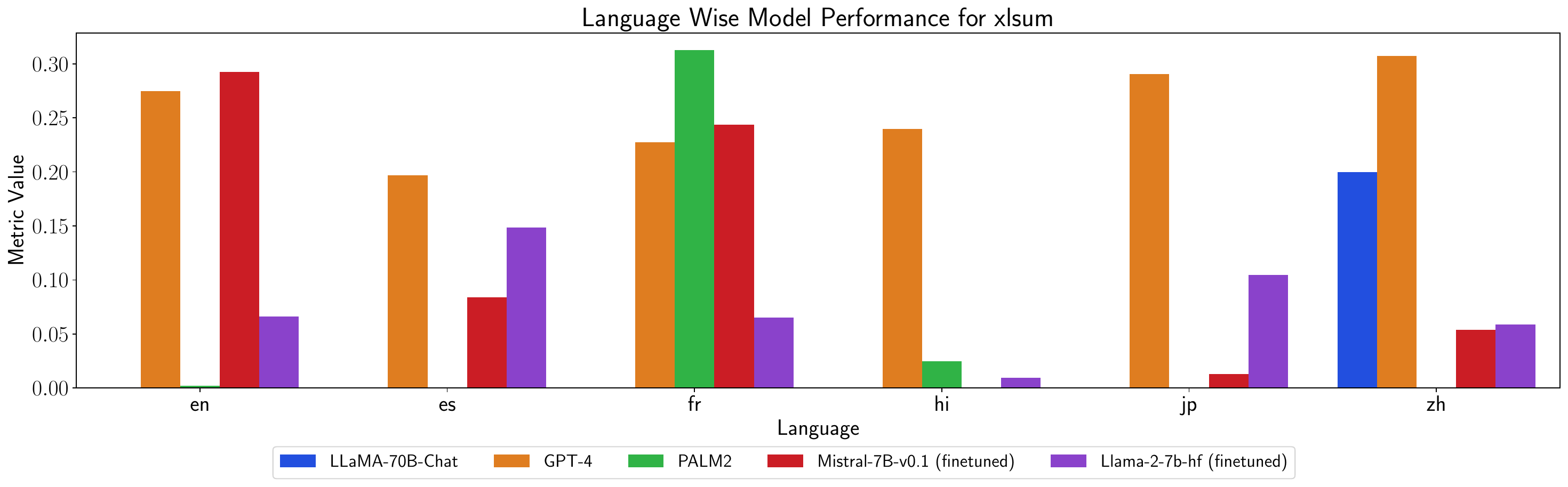}
    \caption{Detailed language-wise comparison of our fine-tuned \mis\ and \llm\ models with other baselines \cite{ahuja2023megaverse} on Arabic, English, Spanish, French, Hindi and Japanese for XLSUM\cite{hasan-etal-2021-xl}.}
    \label{fig:langwise xlsum}
\end{figure*}

\begin{landscape}
    \setlength{\tabcolsep}{3pt}
    \begin{table}
\tiny
\centering
\begin{tabular}{llrrrrrrrrrrrrrrrrrrrrrrrrrrrrrrrrrrrr}
\toprule
\multicolumn{2}{c}{Model} & af & ar & as & bg & bn & cs & da & de & en & es & fi & fr & gu & he & hi & hr & id & it & ja & ko & ml & mr & nb & nl & pt & ru & sv & sw & ta & te & th & tr & ur & vi & zh & avg \\
\midrule
\multicolumn{2}{c}{LLaMA-70B-Chat} & -- & 0.42 & -- & -- & -- & 0.65 & 0.66 & 0.69 & 0.79 & 0.68 & 0.63 & 0.72 & -- & 0.41 & -- & -- & -- & 0.68 & 0.57 & 0.56 & -- & -- & -- & 0.66 & 0.70 & 0.67 & 0.67 & -- & -- & -- & 0.39 & 0.47 & -- & -- & 0.62 & 0.61 \\
\multicolumn{2}{c}{Mistral-7B-Instruct} & -- & 0.09 & -- & -- & -- & 0.54 & 0.51 & 0.52 & 0.65 & 0.59 & 0.35 & 0.60 & -- & 0.02 & -- & -- & -- & 0.51 & 0.42 & 0.41 & -- & -- & -- & 0.55 & 0.54 & 0.62 & 0.55 & -- & -- & -- & 0.04 & 0.31 & -- & -- & 0.60 & 0.44 \\
\multicolumn{2}{c}{GPT-3.5-Turbo} & -- & 0.69 & -- & -- & -- & 0.77 & 0.81 & 0.83 & 0.88 & 0.79 & 0.78 & 0.83 & -- & 0.64 & -- & -- & -- & 0.80 & 0.71 & 0.67 & -- & -- & -- & 0.80 & 0.83 & 0.78 & 0.82 & -- & -- & -- & 0.56 & 0.70 & -- & -- & 0.78 & 0.76 \\
\multicolumn{2}{c}{GPT-4} & -- & \bfseries 0.92 & -- & -- & -- & 0.85 & 0.82 & 0.85 & 0.79 & 0.83 & \bfseries 0.87 & \bfseries 0.91 & -- & \bfseries 0.87 & -- & -- & -- & 0.81 & \bfseries 0.86 & \bfseries 0.89 & -- & -- & -- & 0.85 & 0.79 & \bfseries 0.91 & 0.80 & -- & -- & -- & \bfseries 0.84 & 0.78 & -- & -- & \bfseries 0.89 & 0.85 \\
\multicolumn{2}{c}{PALM2} & -- & 0.87 & -- & -- & -- & \bfseries 0.88 & \bfseries 0.88 & \bfseries 0.88 & \bfseries 0.92 & \bfseries 0.86 & 0.87 & 0.88 & -- & 0.85 & -- & -- & -- & \bfseries 0.86 & 0.84 & 0.86 & -- & -- & -- & \bfseries 0.87 & \bfseries 0.88 & 0.87 & \bfseries 0.87 & -- & -- & -- & 0.81 & \bfseries 0.85 & -- & -- & 0.87 & \bfseries 0.87 \\
\toprule
rank & quantisation & af & ar & as & bg & bn & cs & da & de & en & es & fi & fr & gu & he & hi & hr & id & it & ja & ko & ml & mr & nb & nl & pt & ru & sv & sw & ta & te & th & tr & ur & vi & zh & avg \\
\bottomrule
\end{tabular}
\caption{Detailed performance of various \textsc{alpaca} finetuned \llm\ models on Belebele \cite{bandarkar2023belebele}.}
\label{tab:results_belebele_llama_alpaca}

\end{table}
    \begin{table}
\tiny
\centering
\begin{tabular}{llrrrrrrrrrrrrrrrrrrrrrrrrrrrrrrrrrrrr}
\toprule
\multicolumn{2}{c}{Model}& bn & cs & da & de & en & es & fi & fr & gu & he & hi & hr & id & it & ja & ko & ml & mr & nb & nl & pt & ru & sv & sw & ta & te & th & tr & ur & vi & zh & avg \\
\midrule
\multicolumn{2}{c}{LLaMA-70B-Chat} & -- & 0.42 & -- & -- & -- & 0.65 & 0.66 & 0.69 & 0.79 & 0.68 & 0.63 & 0.72 & -- & 0.41 & -- & -- & -- & 0.68 & 0.57 & 0.56 & -- & -- & -- & 0.66 & 0.70 & 0.67 & 0.67 & -- & -- & -- & 0.39 & 0.47 & -- & -- & 0.62 & 0.61 \\
\multicolumn{2}{c}{Mistral-7B-Instruct} & -- & 0.09 & -- & -- & -- & 0.54 & 0.51 & 0.52 & 0.65 & 0.59 & 0.35 & 0.60 & -- & 0.02 & -- & -- & -- & 0.51 & 0.42 & 0.41 & -- & -- & -- & 0.55 & 0.54 & 0.62 & 0.55 & -- & -- & -- & 0.04 & 0.31 & -- & -- & 0.60 & 0.44 \\
\multicolumn{2}{c}{GPT-3.5-Turbo} & -- & 0.69 & -- & -- & -- & 0.77 & 0.81 & 0.83 & 0.88 & 0.79 & 0.78 & 0.83 & -- & 0.64 & -- & -- & -- & 0.80 & 0.71 & 0.67 & -- & -- & -- & 0.80 & 0.83 & 0.78 & 0.82 & -- & -- & -- & 0.56 & 0.70 & -- & -- & 0.78 & 0.76 \\
\multicolumn{2}{c}{GPT-4} & -- & \bfseries 0.92 & -- & -- & -- & 0.85 & 0.82 & 0.85 & 0.79 & 0.83 & \bfseries 0.87 & \bfseries 0.91 & -- & \bfseries 0.87 & -- & -- & -- & 0.81 & \bfseries 0.86 & \bfseries 0.89 & -- & -- & -- & 0.85 & 0.79 & \bfseries 0.91 & 0.80 & -- & -- & -- & \bfseries 0.84 & 0.78 & -- & -- & \bfseries 0.89 & 0.85 \\
\multicolumn{2}{c}{PALM2} & -- & 0.87 & -- & -- & -- & \bfseries 0.88 & \bfseries 0.88 & \bfseries 0.88 & \bfseries 0.92 & \bfseries 0.86 & 0.87 & 0.88 & -- & 0.85 & -- & -- & -- & \bfseries 0.86 & 0.84 & 0.86 & -- & -- & -- & \bfseries 0.87 & \bfseries 0.88 & 0.87 & \bfseries 0.87 & -- & -- & -- & 0.81 & \bfseries 0.85 & -- & -- & 0.87 & \bfseries 0.87 \\
\midrule
rank & quantisation & af & ar & as & bg & bn & cs & da & de & en & es & fi & fr & gu & he & hi & hr & id & it & ja & ko & ml & mr & nb & nl & pt & ru & sv & sw & ta & te & th & tr & ur & vi & zh & avg \\
\bottomrule
\end{tabular}
\caption{Detailed performance of various \textsc{alpaca} finetuned \mis\ models on Belebele \cite{bandarkar2023belebele}.}
\label{tab:results_belebele_mistral_alpaca}
\end{table}
\end{landscape}

\begin{landscape}
    \setlength{\tabcolsep}{3pt}
    \begin{table}
\centering
\tiny
\begin{tabular}{llrrrrrrrrrrrrrrrrrrrrrrrrrrrrrrrrrrrr}
\toprule
\multicolumn{2}{c}{Model} & af & ar & as & bg & bn & cs & da & de & en & es & fi & fr & gu & he & hi & hr & id & it & ja & ko & ml & mr & nb & nl & pt & ru & sv & sw & ta & te & th & tr & ur & vi & zh & avg \\
\midrule
\multicolumn{2}{c}{LLaMA-70B-Chat} & -- & 0.42 & -- & -- & -- & 0.65 & 0.66 & 0.69 & 0.79 & 0.68 & 0.63 & 0.72 & -- & 0.41 & -- & -- & -- & 0.68 & 0.57 & 0.56 & -- & -- & -- & 0.66 & 0.70 & 0.67 & 0.67 & -- & -- & -- & 0.39 & 0.47 & -- & -- & 0.62 & 0.61 \\
\multicolumn{2}{c}{Mistral-7B-Instruct} & -- & 0.09 & -- & -- & -- & 0.54 & 0.51 & 0.52 & 0.65 & 0.59 & 0.35 & 0.60 & -- & 0.02 & -- & -- & -- & 0.51 & 0.42 & 0.41 & -- & -- & -- & 0.55 & 0.54 & 0.62 & 0.55 & -- & -- & -- & 0.04 & 0.31 & -- & -- & 0.60 & 0.44 \\
\multicolumn{2}{c}{GPT-3.5-Turbo} & -- & 0.69 & -- & -- & -- & 0.77 & 0.81 & 0.83 & 0.88 & 0.79 & 0.78 & 0.83 & -- & 0.64 & -- & -- & -- & 0.80 & 0.71 & 0.67 & -- & -- & -- & 0.80 & 0.83 & 0.78 & 0.82 & -- & -- & -- & 0.56 & 0.70 & -- & -- & 0.78 & 0.76 \\
\multicolumn{2}{c}{GPT-4} & -- & \bfseries 0.92 & -- & -- & -- & 0.85 & 0.82 & 0.85 & 0.79 & 0.83 & \bfseries 0.87 & \bfseries 0.91 & -- & \bfseries 0.87 & -- & -- & -- & 0.81 & \bfseries 0.86 & \bfseries 0.89 & -- & -- & -- & 0.85 & 0.79 & \bfseries 0.91 & 0.80 & -- & -- & -- & \bfseries 0.84 & 0.78 & -- & -- & \bfseries 0.89 & 0.85 \\
\multicolumn{2}{c}{PALM2} & -- & 0.87 & -- & -- & -- & \bfseries 0.88 & \bfseries 0.88 & \bfseries 0.88 & \bfseries 0.92 & \bfseries 0.86 & 0.87 & 0.88 & -- & 0.85 & -- & -- & -- & \bfseries 0.86 & 0.84 & 0.86 & -- & -- & -- & \bfseries 0.87 & \bfseries 0.88 & 0.87 & \bfseries 0.87 & -- & -- & -- & 0.81 & \bfseries 0.85 & -- & -- & 0.87 & \bfseries 0.87 \\
\midrule
rank & quantisation & af & ar & as & bg & bn & cs & da & de & en & es & fi & fr & gu & he & hi & hr & id & it & ja & ko & ml & mr & nb & nl & pt & ru & sv & sw & ta & te & th & tr & ur & vi & zh & avg \\
\bottomrule
\end{tabular}
\caption{Detailed performance of various \textsc{multialpaca} finetuned \llm\ models on Belebele \cite{bandarkar2023belebele}.}
\label{tab:results_belebele_llama_multialpaca}
\end{table}
    \begin{table}
\tiny
\centering
\begin{tabular}{llrrrrrrrrrrrrrrrrrrrrrrrrrrrrrrrrrrrr}
\toprule
\multicolumn{2}{c}{Model} & af & ar & as & bg & bn & cs & da & de & en & es & fi & fr & gu & he & hi & hr & id & it & ja & ko & ml & mr & nb & nl & pt & ru & sv & sw & ta & te & th & tr & ur & vi & zh & avg \\
\midrule
\multicolumn{2}{c}{LLaMA-70B-Chat} & -- & 0.42 & -- & -- & -- & 0.65 & 0.66 & 0.69 & 0.79 & 0.68 & 0.63 & 0.72 & -- & 0.41 & -- & -- & -- & 0.68 & 0.57 & 0.56 & -- & -- & -- & 0.66 & 0.70 & 0.67 & 0.67 & -- & -- & -- & 0.39 & 0.47 & -- & -- & 0.62 & 0.61 \\
\multicolumn{2}{c}{Mistral-7B-Instruct} & -- & 0.09 & -- & -- & -- & 0.54 & 0.51 & 0.52 & 0.65 & 0.59 & 0.35 & 0.60 & -- & 0.02 & -- & -- & -- & 0.51 & 0.42 & 0.41 & -- & -- & -- & 0.55 & 0.54 & 0.62 & 0.55 & -- & -- & -- & 0.04 & 0.31 & -- & -- & 0.60 & 0.44 \\
\multicolumn{2}{c}{GPT-3.5-Turbo} & -- & 0.69 & -- & -- & -- & 0.77 & 0.81 & 0.83 & 0.88 & 0.79 & 0.78 & 0.83 & -- & 0.64 & -- & -- & -- & 0.80 & 0.71 & 0.67 & -- & -- & -- & 0.80 & 0.83 & 0.78 & 0.82 & -- & -- & -- & 0.56 & 0.70 & -- & -- & 0.78 & 0.76 \\
\multicolumn{2}{c}{GPT-4} & -- & \bfseries 0.92 & -- & -- & -- & 0.85 & 0.82 & 0.85 & 0.79 & 0.83 & \bfseries 0.87 & \bfseries 0.91 & -- & \bfseries 0.87 & -- & -- & -- & 0.81 & \bfseries 0.86 & \bfseries 0.89 & -- & -- & -- & 0.85 & 0.79 & \bfseries 0.91 & 0.80 & -- & -- & -- & \bfseries 0.84 & 0.78 & -- & -- & \bfseries 0.89 & 0.85 \\
\multicolumn{2}{c}{PALM2} & -- & 0.87 & -- & -- & -- & \bfseries 0.88 & \bfseries 0.88 & \bfseries 0.88 & \bfseries 0.92 & \bfseries 0.86 & 0.87 & 0.88 & -- & 0.85 & -- & -- & -- & \bfseries 0.86 & 0.84 & 0.86 & -- & -- & -- & \bfseries 0.87 & \bfseries 0.88 & 0.87 & \bfseries 0.87 & -- & -- & -- & 0.81 & \bfseries 0.85 & -- & -- & 0.87 & \bfseries 0.87 \\
\midrule
rank & quantisation & af & ar & as & bg & bn & cs & da & de & en & es & fi & fr & gu & he & hi & hr & id & it & ja & ko & ml & mr & nb & nl & pt & ru & sv & sw & ta & te & th & tr & ur & vi & zh & avg \\
\bottomrule
\end{tabular}

\caption{Detailed performance of various \textsc{multialpaca} finetuned \mis\ models on Belebele \cite{bandarkar2023belebele}.}
\label{tab:results_belebele_mistral_multialpaca}

\end{table}
\end{landscape}

\begin{landscape}
    \setlength{\tabcolsep}{3pt}
    \begin{table}
\tiny
\centering
\begin{tabular}{llrrrrrrrrrrrrrrrrrrrrrrrrrrrrrrrrrrrr}
\toprule
\multicolumn{2}{c}{Model} & af & ar & as & bg & bn & cs & da & de & en & es & fi & fr & gu & he & hi & hr & id & it & ja & ko & ml & mr & nb & nl & pt & ru & sv & sw & ta & te & th & tr & ur & vi & zh & avg \\
\midrule
\multicolumn{2}{c}{LLaMA-70B-Chat} & -- & 0.42 & -- & -- & -- & 0.65 & 0.66 & 0.69 & 0.79 & 0.68 & 0.63 & 0.72 & -- & 0.41 & -- & -- & -- & 0.68 & 0.57 & 0.56 & -- & -- & -- & 0.66 & 0.70 & 0.67 & 0.67 & -- & -- & -- & 0.39 & 0.47 & -- & -- & 0.62 & 0.61 \\
\multicolumn{2}{c}{Mistral-7B-Instruct} & -- & 0.09 & -- & -- & -- & 0.54 & 0.51 & 0.52 & 0.65 & 0.59 & 0.35 & 0.60 & -- & 0.02 & -- & -- & -- & 0.51 & 0.42 & 0.41 & -- & -- & -- & 0.55 & 0.54 & 0.62 & 0.55 & -- & -- & -- & 0.04 & 0.31 & -- & -- & 0.60 & 0.44 \\
\multicolumn{2}{c}{GPT-3.5-Turbo} & -- & 0.69 & -- & -- & -- & 0.77 & 0.81 & 0.83 & 0.88 & 0.79 & 0.78 & 0.83 & -- & 0.64 & -- & -- & -- & 0.80 & 0.71 & 0.67 & -- & -- & -- & 0.80 & 0.83 & 0.78 & 0.82 & -- & -- & -- & 0.56 & 0.70 & -- & -- & 0.78 & 0.76 \\
\multicolumn{2}{c}{GPT-4} & -- & \bfseries 0.92 & -- & -- & -- & 0.85 & 0.82 & 0.85 & 0.79 & 0.83 & \bfseries 0.87 & \bfseries 0.91 & -- & \bfseries 0.87 & -- & -- & -- & 0.81 & \bfseries 0.86 & \bfseries 0.89 & -- & -- & -- & 0.85 & 0.79 & \bfseries 0.91 & 0.80 & -- & -- & -- & \bfseries 0.84 & 0.78 & -- & -- & \bfseries 0.89 & 0.85 \\
\multicolumn{2}{c}{PALM2} & -- & 0.87 & -- & -- & -- & \bfseries 0.88 & \bfseries 0.88 & \bfseries 0.88 & \bfseries 0.92 & \bfseries 0.86 & 0.87 & 0.88 & -- & 0.85 & -- & -- & -- & \bfseries 0.86 & 0.84 & 0.86 & -- & -- & -- & \bfseries 0.87 & \bfseries 0.88 & 0.87 & \bfseries 0.87 & -- & -- & -- & 0.81 & \bfseries 0.85 & -- & -- & 0.87 & \bfseries 0.87 \\
\midrule

rank & quantisation & af & ar & as & bg & bn & cs & da & de & en & es & fi & fr & gu & he & hi & hr & id & it & ja & ko & ml & mr & nb & nl & pt & ru & sv & sw & ta & te & th & tr & ur & vi & zh & avg \\
\bottomrule
\end{tabular}

\caption{Detailed performance of various \textsc{Bactrian-X-11} finetuned \llm\  models on Belebele \cite{bandarkar2023belebele}.}
\label{tab:results_belebele_llama_bactrian_x_11}
\end{table}
    \begin{table}
\tiny
\centering
\begin{tabular}{llrrrrrrrrrrrrrrrrrrrrrrrrrrrrrrrrrrrr}
\toprule
\multicolumn{2}{c}{Model} & af & ar & as & bg & bn & cs & da & de & en & es & fi & fr & gu & he & hi & hr & id & it & ja & ko & ml & mr & nb & nl & pt & ru & sv & sw & ta & te & th & tr & ur & vi & zh & avg \\
\midrule
\multicolumn{2}{c}{LLaMA-70B-Chat} & -- & 0.42 & -- & -- & -- & 0.65 & 0.66 & 0.69 & 0.79 & 0.68 & 0.63 & 0.72 & -- & 0.41 & -- & -- & -- & 0.68 & 0.57 & 0.56 & -- & -- & -- & 0.66 & 0.70 & 0.67 & 0.67 & -- & -- & -- & 0.39 & 0.47 & -- & -- & 0.62 & 0.61 \\
\multicolumn{2}{c}{Mistral-7B-Instruct} & -- & 0.09 & -- & -- & -- & 0.54 & 0.51 & 0.52 & 0.65 & 0.59 & 0.35 & 0.60 & -- & 0.02 & -- & -- & -- & 0.51 & 0.42 & 0.41 & -- & -- & -- & 0.55 & 0.54 & 0.62 & 0.55 & -- & -- & -- & 0.04 & 0.31 & -- & -- & 0.60 & 0.44 \\
\multicolumn{2}{c}{GPT-3.5-Turbo} & -- & 0.69 & -- & -- & -- & 0.77 & 0.81 & 0.83 & 0.88 & 0.79 & 0.78 & 0.83 & -- & 0.64 & -- & -- & -- & 0.80 & 0.71 & 0.67 & -- & -- & -- & 0.80 & 0.83 & 0.78 & 0.82 & -- & -- & -- & 0.56 & 0.70 & -- & -- & 0.78 & 0.76 \\
\multicolumn{2}{c}{GPT-4} & -- & \bfseries 0.92 & -- & -- & -- & 0.85 & 0.82 & 0.85 & 0.79 & 0.83 & \bfseries 0.87 & \bfseries 0.91 & -- & \bfseries 0.87 & -- & -- & -- & 0.81 & \bfseries 0.86 & \bfseries 0.89 & -- & -- & -- & 0.85 & 0.79 & \bfseries 0.91 & 0.80 & -- & -- & -- & \bfseries 0.84 & 0.78 & -- & -- & \bfseries 0.89 & 0.85 \\
\multicolumn{2}{c}{PALM2} & -- & 0.87 & -- & -- & -- & \bfseries 0.88 & \bfseries 0.88 & \bfseries 0.88 & \bfseries 0.92 & \bfseries 0.86 & 0.87 & 0.88 & -- & 0.85 & -- & -- & -- & \bfseries 0.86 & 0.84 & 0.86 & -- & -- & -- & \bfseries 0.87 & \bfseries 0.88 & 0.87 & \bfseries 0.87 & -- & -- & -- & 0.81 & \bfseries 0.85 & -- & -- & 0.87 & \bfseries 0.87 \\
\midrule
rank & quantisation & af & ar & as & bg & bn & cs & da & de & en & es & fi & fr & gu & he & hi & hr & id & it & ja & ko & ml & mr & nb & nl & pt & ru & sv & sw & ta & te & th & tr & ur & vi & zh & avg \\
\bottomrule
\end{tabular}

\caption{Detailed performance of various \textsc{Bactrian-X-11} finetuned \mis\  models on Belebele \cite{bandarkar2023belebele}.}
\label{tab:results_belebele_mistral_bactrian_x_11}
\end{table}
\end{landscape}

\begin{landscape}
    \setlength{\tabcolsep}{3pt}
    \begin{table}
\tiny
\centering
\begin{tabular}{llrrrrrrrrrrrrrrrrrrrrrrrrrrrrrrrrrrrr}
\toprule
\multicolumn{2}{c}{Model} & af & ar & as & bg & bn & cs & da & de & en & es & fi & fr & gu & he & hi & hr & id & it & ja & ko & ml & mr & nb & nl & pt & ru & sv & sw & ta & te & th & tr & ur & vi & zh & avg \\
\midrule
\multicolumn{2}{c}{LLaMA-70B-Chat} & -- & 0.42 & -- & -- & -- & 0.65 & 0.66 & 0.69 & 0.79 & 0.68 & 0.63 & 0.72 & -- & 0.41 & -- & -- & -- & 0.68 & 0.57 & 0.56 & -- & -- & -- & 0.66 & 0.70 & 0.67 & 0.67 & -- & -- & -- & 0.39 & 0.47 & -- & -- & 0.62 & 0.61 \\
\multicolumn{2}{c}{Mistral-7B-Instruct} & -- & 0.09 & -- & -- & -- & 0.54 & 0.51 & 0.52 & 0.65 & 0.59 & 0.35 & 0.60 & -- & 0.02 & -- & -- & -- & 0.51 & 0.42 & 0.41 & -- & -- & -- & 0.55 & 0.54 & 0.62 & 0.55 & -- & -- & -- & 0.04 & 0.31 & -- & -- & 0.60 & 0.44 \\
\multicolumn{2}{c}{GPT-3.5-Turbo} & -- & 0.69 & -- & -- & -- & 0.77 & 0.81 & 0.83 & 0.88 & 0.79 & 0.78 & 0.83 & -- & 0.64 & -- & -- & -- & 0.80 & 0.71 & 0.67 & -- & -- & -- & 0.80 & 0.83 & 0.78 & 0.82 & -- & -- & -- & 0.56 & 0.70 & -- & -- & 0.78 & 0.76 \\
\multicolumn{2}{c}{GPT-4} & -- & \bfseries 0.92 & -- & -- & -- & 0.85 & 0.82 & 0.85 & 0.79 & 0.83 & \bfseries 0.87 & \bfseries 0.91 & -- & \bfseries 0.87 & -- & -- & -- & 0.81 & \bfseries 0.86 & \bfseries 0.89 & -- & -- & -- & 0.85 & 0.79 & \bfseries 0.91 & 0.80 & -- & -- & -- & \bfseries 0.84 & 0.78 & -- & -- & \bfseries 0.89 & 0.85 \\
\multicolumn{2}{c}{PALM2} & -- & 0.87 & -- & -- & -- & \bfseries 0.88 & \bfseries 0.88 & \bfseries 0.88 & \bfseries 0.92 & \bfseries 0.86 & 0.87 & 0.88 & -- & 0.85 & -- & -- & -- & \bfseries 0.86 & 0.84 & 0.86 & -- & -- & -- & \bfseries 0.87 & \bfseries 0.88 & 0.87 & \bfseries 0.87 & -- & -- & -- & 0.81 & \bfseries 0.85 & -- & -- & 0.87 & \bfseries 0.87 \\
\midrule
rank & quantisation & af & ar & as & bg & bn & cs & da & de & en & es & fi & fr & gu & he & hi & hr & id & it & ja & ko & ml & mr & nb & nl & pt & ru & sv & sw & ta & te & th & tr & ur & vi & zh & avg \\
\bottomrule
\end{tabular}
\caption{Detailed performance of various \textsc{Bactrian-X-22} finetuned \llm\ models on Belebele \cite{bandarkar2023belebele}.}
\label{tab:results_belebele_llama_bactrian}

\end{table}
    \begin{table}
\tiny
\centering
\begin{tabular}{llrrrrrrrrrrrrrrrrrrrrrrrrrrrrrrrrrrrr}
\toprule
\multicolumn{2}{c}{Model} & af & ar & as & bg & bn & cs & da & de & en & es & fi & fr & gu & he & hi & hr & id & it & ja & ko & ml & mr & nb & nl & pt & ru & sv & sw & ta & te & th & tr & ur & vi & zh & avg \\
\midrule
\multicolumn{2}{c}{LLaMA-70B-Chat} & -- & 0.42 & -- & -- & -- & 0.65 & 0.66 & 0.69 & 0.79 & 0.68 & 0.63 & 0.72 & -- & 0.41 & -- & -- & -- & 0.68 & 0.57 & 0.56 & -- & -- & -- & 0.66 & 0.70 & 0.67 & 0.67 & -- & -- & -- & 0.39 & 0.47 & -- & -- & 0.62 & 0.61 \\
\multicolumn{2}{c}{Mistral-7B-Instruct} & -- & 0.09 & -- & -- & -- & 0.54 & 0.51 & 0.52 & 0.65 & 0.59 & 0.35 & 0.60 & -- & 0.02 & -- & -- & -- & 0.51 & 0.42 & 0.41 & -- & -- & -- & 0.55 & 0.54 & 0.62 & 0.55 & -- & -- & -- & 0.04 & 0.31 & -- & -- & 0.60 & 0.44 \\
\multicolumn{2}{c}{GPT-3.5-Turbo} & -- & 0.69 & -- & -- & -- & 0.77 & 0.81 & 0.83 & 0.88 & 0.79 & 0.78 & 0.83 & -- & 0.64 & -- & -- & -- & 0.80 & 0.71 & 0.67 & -- & -- & -- & 0.80 & 0.83 & 0.78 & 0.82 & -- & -- & -- & 0.56 & 0.70 & -- & -- & 0.78 & 0.76 \\
\multicolumn{2}{c}{GPT-4} & -- & \bfseries 0.92 & -- & -- & -- & 0.85 & 0.82 & 0.85 & 0.79 & 0.83 & \bfseries 0.87 & \bfseries 0.91 & -- & \bfseries 0.87 & -- & -- & -- & 0.81 & \bfseries 0.86 & \bfseries 0.89 & -- & -- & -- & 0.85 & 0.79 & \bfseries 0.91 & 0.80 & -- & -- & -- & \bfseries 0.84 & 0.78 & -- & -- & \bfseries 0.89 & 0.85 \\
\multicolumn{2}{c}{PALM2} & -- & 0.87 & -- & -- & -- & \bfseries 0.88 & \bfseries 0.88 & \bfseries 0.88 & \bfseries 0.92 & \bfseries 0.86 & 0.87 & 0.88 & -- & 0.85 & -- & -- & -- & \bfseries 0.86 & 0.84 & 0.86 & -- & -- & -- & \bfseries 0.87 & \bfseries 0.88 & 0.87 & \bfseries 0.87 & -- & -- & -- & 0.81 & \bfseries 0.85 & -- & -- & 0.87 & \bfseries 0.87 \\
\midrule
rank & quantisation & af & ar & as & bg & bn & cs & da & de & en & es & fi & fr & gu & he & hi & hr & id & it & ja & ko & ml & mr & nb & nl & pt & ru & sv & sw & ta & te & th & tr & ur & vi & zh & avg \\
\bottomrule
\end{tabular}
\caption{Detailed performance of various \textsc{Bactrian-X-22} finetuned \mis\ models on Belebele \cite{bandarkar2023belebele}.}
\label{tab:results_belebele_mistral_bactrian}

\end{table}
\end{landscape}

\begin{table}[t]
\small
\centering
\begin{tabular}{rrr}
\toprule
\multicolumn{2}{c}{Model} & win\_rate \\
\midrule
\multicolumn{2}{c}{LLaMA-70B-Chat} & 22.36 \\
\multicolumn{2}{c}{Mistral-7B-Instruct} & 35.13 \\
\multicolumn{2}{c}{GPT-4} & \bfseries 93.78 \\
\multicolumn{2}{c}{PALM2} & 79.66 \\
\toprule
rank & quantisation & win\_rate \\
\midrule
64 & 8 & 13.05 \\
64 & 16 & 13.11 \\
128 & 8 & 13.28 \\
128 & 16 & 13.23 \\
\bottomrule
\end{tabular}
\caption{Detailed performance of various \textsc{alpaca} finetuned \llm\  models on AlpacaEval \cite{li2023alpacaeval}.}
\label{tab:results_alpaca_eval_llama_alpaca}
\end{table}
\begin{table}[t]
\small
\centering
\begin{tabular}{rrr}
\toprule
\multicolumn{2}{c}{Model} & win\_rate \\
\midrule
\multicolumn{2}{c}{LLaMA-70B-Chat} & 22.36 \\
\multicolumn{2}{c}{Mistral-7B-Instruct} & 35.13 \\
\multicolumn{2}{c}{GPT-4} & \bfseries 93.78 \\
\multicolumn{2}{c}{PALM2} & 79.66 \\
\midrule
rank & quantisation & win\_rate \\
\midrule
8 & 4 & 11.61 \\
8 & 8 & 13.56 \\
8 & 16 & 13.36 \\
16 & 4 & 11.99 \\
16 & 8 & 13.38 \\
16 & 16 & 13.73 \\
32 & 4 & 11.86 \\
32 & 8 & 12.47 \\
32 & 16 & 13.73 \\
64 & 4 & 11.61 \\
64 & 8 & 13.47 \\
64 & 16 & 13.86 \\
128 & 4 & 11.35 \\
128 & 8 & 13.56 \\
128 & 16 & 13.73 \\
\bottomrule
\end{tabular}
\caption{Detailed performance of various \textsc{multialpaca} finetuned \llm\ models on AlpacaEval \cite{li2023alpacaeval}.}
\label{tab:results_alpaca_eval_llama_multialpaca}
\end{table}
\begin{table}
\small
\centering
\begin{tabular}{rrr}
\toprule
\multicolumn{2}{c}{Model} & win\_rate \\
\midrule
\multicolumn{2}{c}{LLaMA-70B-Chat} & 22.36 \\
\multicolumn{2}{c}{Mistral-7B-Instruct} & 35.13 \\
\multicolumn{2}{c}{GPT-4} & \bfseries 93.78 \\
\multicolumn{2}{c}{PALM2} & 79.66 \\
\midrule
rank & quantisation & win\_rate \\
\midrule
8 & 4 & 12.20 \\
8 & 8 & 13.03 \\
8 & 16 & 13.39 \\
16 & 4 & 11.79 \\
16 & 8 & 12.74 \\
16 & 16 & 13.83 \\
32 & 4 & 11.49 \\
32 & 8 & 12.56 \\
32 & 16 & 13.80 \\
64 & 4 & 11.99 \\
64 & 8 & 13.21 \\
64 & 16 & 13.22 \\
128 & 4 & 10.71 \\
128 & 8 & 12.75 \\
128 & 16 & 0.00 \\
\bottomrule
\end{tabular}

\caption{Detailed performance of various \textsc{Bactrian-X--11} finetuned \llm\  models on Alpaca Eval \cite{li2023alpacaeval}.}
\label{tab:results_alpaca_eval_llama_bactrian_x_11}
\end{table}
\begin{table}[t]
\small
\centering
\begin{tabular}{rrr}
\toprule
\multicolumn{2}{c}{Model} & win\_rate \\
\midrule
\multicolumn{2}{c}{LLaMA-70B-Chat} & 22.36 \\
\multicolumn{2}{c}{Mistral-7B-Instruct} & 35.13 \\
\multicolumn{2}{c}{GPT-4} & \bfseries 93.78 \\
\multicolumn{2}{c}{PALM2} & 79.66 \\
\toprule
rank & quantisation & win\_rate \\
\midrule
8 & 4 & 11.86 \\
8 & 8 & 12.81 \\
8 & 16 & 13.23 \\
16 & 4 & 11.97 \\
16 & 8 & 13.06 \\
16 & 16 & 13.36 \\
32 & 4 & 11.72 \\
32 & 8 & 13.15 \\
32 & 16 & 13.36 \\
64 & 4 & 11.47 \\
64 & 8 & 13.67 \\
64 & 16 & 13.73 \\
128 & 4 & 11.10 \\
128 & 8 & 13.05 \\
128 & 16 & 0.00 \\
\bottomrule
\end{tabular}
\caption{Detailed performance of various \textsc{Bactrian-X-22} finetuned \llm\ models on AlpacaEval \cite{li2023alpacaeval}.}
\label{tab:results_alpaca_eval_llama_bactrian}
\end{table}

\begin{table}[t]
\small
\centering
\begin{tabular}{rrr}
\toprule
\multicolumn{2}{c}{Model} & win\_rate \\
\midrule
\multicolumn{2}{c}{LLaMA-70B-Chat} & 22.36 \\
\multicolumn{2}{c}{Mistral-7B-Instruct} & 35.13 \\
\multicolumn{2}{c}{GPT-4} & \bfseries 93.78 \\
\multicolumn{2}{c}{PALM2} & 79.66 \\
\midrule
rank & quantisation & win\_rate \\
\midrule
64 & 8 & 20.47 \\
64 & 16 & 18.89 \\
128 & 8 & 19.55 \\
128 & 16 & 19.33 \\
\bottomrule
\end{tabular}
\caption{Detailed performance of various \textsc{alpaca} finetuned \mis\ models on AlpacaEval \cite{li2023alpacaeval}.}
\label{tab:results_alpaca_eval_mistral_alpaca}

\end{table}
\begin{table}[t]
\small
\centering
\begin{tabular}{rrr}
\toprule
\multicolumn{2}{c}{Model} & win\_rate \\
\midrule
\multicolumn{2}{c}{LLaMA-70B-Chat} & 22.36 \\
\multicolumn{2}{c}{Mistral-7B-Instruct} & 35.13 \\
\multicolumn{2}{c}{GPT-4} & \bfseries 93.78 \\
\multicolumn{2}{c}{PALM2} & 79.66 \\
\midrule
rank & quantisation & win\_rate \\
\midrule
8 & 4 & 15.98 \\
8 & 8 & 21.42 \\
8 & 16 & 20.71 \\
16 & 4 & 15.90 \\
16 & 8 & 20.55 \\
16 & 16 & 21.27 \\
32 & 4 & 16.98 \\
32 & 8 & 22.45 \\
32 & 16 & 22.08 \\
64 & 4 & 16.83 \\
64 & 8 & 21.07 \\
64 & 16 & 18.70 \\
128 & 4 & 16.52 \\
128 & 8 & 20.92 \\
128 & 16 & 21.33 \\
\bottomrule
\end{tabular}

\caption{Detailed performance of various \textsc{multialpaca} finetuned \mis\ models on AlpacaEval \cite{li2023alpacaeval}.}
\label{tab:results_alpaca_eval_mistral_multialpaca}

\end{table}
\begin{table}
\small
\centering
\begin{tabular}{rrr}
\toprule
\multicolumn{2}{c}{Model} & win\_rate \\
\midrule
\multicolumn{2}{c}{LLaMA-70B-Chat} & 22.36 \\
\multicolumn{2}{c}{Mistral-7B-Instruct} & 35.13 \\
\multicolumn{2}{c}{GPT-4} & \bfseries 93.78 \\
\multicolumn{2}{c}{PALM2} & 79.66 \\
\midrule
rank & quantisation & win\_rate \\
8 & 4 & 14.43 \\
8 & 8 & 20.61 \\
8 & 16 & 21.93 \\
16 & 4 & 15.22 \\
16 & 8 & 22.23 \\
16 & 16 & 20.85 \\
32 & 4 & 16.43 \\
32 & 8 & 21.29 \\
32 & 16 & 21.06 \\
64 & 4 & 16.85 \\
64 & 8 & 22.01 \\
64 & 16 & 20.91 \\
128 & 4 & 17.18 \\
128 & 8 & 22.05 \\
128 & 16 & 22.57 \\
\bottomrule
\end{tabular}

\caption{Detailed performance of various \textsc{Bactrian-X-11} finetuned \mis\  models on Alpaca Eval \cite{li2023alpacaeval}.}
\label{tab:results_alpaca_eval_mistral_bactrian_x_11}
\end{table}
\begin{table}[t]
\small
\centering
\begin{tabular}{rrr}
\toprule
\multicolumn{2}{c}{Model} & win\_rate \\
\midrule
\multicolumn{2}{c}{LLaMA-70B-Chat} & 22.36 \\
\multicolumn{2}{c}{Mistral-7B-Instruct} & 35.13 \\
\multicolumn{2}{c}{GPT-4} & \bfseries 93.78 \\
\multicolumn{2}{c}{PALM2} & 79.66 \\
\midrule
rank & quantisation & win\_rate \\
\midrule
8 & 4 & 15.04 \\
8 & 8 & 20.67 \\
8 & 16 & 21.27 \\
16 & 4 & 15.77 \\
16 & 8 & 22.07 \\
16 & 16 & 21.08 \\
32 & 4 & 15.77 \\
32 & 8 & 21.07 \\
32 & 16 & 21.14 \\
64 & 4 & 16.90 \\
64 & 8 & 21.30 \\
64 & 16 & 21.27 \\
128 & 4 & 17.77 \\
128 & 8 & 22.04 \\
128 & 16 & 21.83 \\
\bottomrule
\end{tabular}
\caption{Detailed performance of various \textsc{Bactrian-X-22} finetuned \mis\  models on AlpacaEval \cite{li2023alpacaeval}.}
\label{tab:results_alpaca_eval_mistral_bactrian}

\end{table}

\begin{table*}
\scriptsize
\centering
\begin{tabular}{rrrrrrrrrr}
\toprule
\multicolumn{2}{c}{Models} & ar & de & en & es & hi & vi & zh & avg \\
\midrule
\multicolumn{2}{c}{LLaMA-7B-Chat} & 0.05 & 0.45 & 0.70 & 0.52 & 0.00 & 0.42 & 0.07 & 0.32 \\
\multicolumn{2}{c}{LLaMA--3B-Chat} & 0.55 & 0.73 & 0.62 & 0.62 & 0.00 & 0.09 & 0.09 & 0.39 \\
\multicolumn{2}{c}{LLaMA-70B-Chat} & 0.04 & 0.42 & 0.56 & 0.34 & 0.01 & 0.27 & 0.05 & 0.24 \\
\multicolumn{2}{c}{Mistral-7B-Instruct} & 0.17 & 0.30 & 0.32 & 0.33 & 0.20 & 0.36 & 0.02 & 0.24 \\
\multicolumn{2}{c}{DV003} & 0.38 & 0.58 & 0.75 & 0.63 & 0.25 & 0.48 & 0.32 & 0.48 \\
\multicolumn{2}{c}{GPT-3.5-Turbo} & 0.48 & 0.51 & 0.73 & 0.54 & 0.51 & 0.59 & 0.57 & 0.56 \\
\multicolumn{2}{c}{GPT-4} & 0.59 & 0.65 & 0.80 & 0.70 & 0.57 & 0.72 & 0.67 & 0.67 \\
\multicolumn{2}{c}{TULRv6} & \bfseries 0.76 & \bfseries 0.80 & \bfseries 0.87 & \bfseries 0.82 & \bfseries 0.82 & \bfseries 0.82 & \bfseries 0.78 & \bfseries 0.81 \\
\multicolumn{2}{c}{XLM-R} & 0.67 & 0.70 & 0.83 & 0.74 & 0.71 & 0.74 & 0.62 & 0.72 \\
\multicolumn{2}{c}{mBERT} & 0.52 & 0.59 & 0.80 & 0.67 & 0.50 & 0.61 & 0.60 & 0.61 \\
\multicolumn{2}{c}{mT5} & 0.57 & 0.62 & 0.82 & 0.67 & 0.55 & 0.66 & 0.62 & 0.64 \\
\multicolumn{2}{c}{PALM2} & 0.22 & 0.44 & 0.69 & 0.45 & 0.36 & 0.49 & 0.06 & 0.39 \\
\midrule
rank & quantisation & ar & de & en & es & hi & vi & zh & avg \\
\midrule
64 & 8 & 0.31 & 0.20 & 0.28 & 0.24 & 0.43 & 0.19 & 0.08 & 0.25 \\
64 & 16 & 0.27 & 0.36 & 0.50 & 0.43 & 0.43 & 0.33 & 0.15 & 0.35 \\
128 & 8 & 0.31 & 0.20 & 0.29 & 0.25 & 0.43 & 0.19 & 0.08 & 0.25 \\
128 & 16 & 0.27 & 0.36 & 0.50 & 0.43 & 0.43 & 0.33 & 0.15 & 0.35 \\
\bottomrule
\end{tabular}

\caption{Detailed performance of various \textsc{alpaca} finetuned \llm\ models on MLQA \cite{lewis2020mlqa}.}
\label{tab:results_mlqa_llama_alpaca}

\end{table*}
\begin{table*}
\scriptsize
\centering
\begin{tabular}{rrrrrrrrrr}
\toprule
\multicolumn{2}{c}{Models} & ar & de & en & es & hi & vi & zh & avg \\
\midrule
\multicolumn{2}{c}{LLaMA-7B-Chat} & 0.05 & 0.45 & 0.70 & 0.52 & 0.00 & 0.42 & 0.07 & 0.32 \\
\multicolumn{2}{c}{LLaMA--3B-Chat} & 0.55 & 0.73 & 0.62 & 0.62 & 0.00 & 0.09 & 0.09 & 0.39 \\
\multicolumn{2}{c}{LLaMA-70B-Chat} & 0.04 & 0.42 & 0.56 & 0.34 & 0.01 & 0.27 & 0.05 & 0.24 \\
\multicolumn{2}{c}{Mistral-7B-Instruct} & 0.17 & 0.30 & 0.32 & 0.33 & 0.20 & 0.36 & 0.02 & 0.24 \\
\multicolumn{2}{c}{DV003} & 0.38 & 0.58 & 0.75 & 0.63 & 0.25 & 0.48 & 0.32 & 0.48 \\
\multicolumn{2}{c}{GPT-3.5-Turbo} & 0.48 & 0.51 & 0.73 & 0.54 & 0.51 & 0.59 & 0.57 & 0.56 \\
\multicolumn{2}{c}{GPT-4} & 0.59 & 0.65 & 0.80 & 0.70 & 0.57 & 0.72 & 0.67 & 0.67 \\
\bottomrule
\end{tabular}

\caption{Detailed performance of various \textsc{multialpaca} finetuned \llm\ models on MLQA \cite{lewis2020mlqa}.}
\label{tab:results_mlqa_llama_multialpaca}

\end{table*}
\begin{table*}
\scriptsize
\centering
\begin{tabular}{rrrrrrrrrr}
\toprule
\multicolumn{2}{c}{Models} & ar & de & en & es & hi & vi & zh & avg \\
\midrule
\multicolumn{2}{c}{LLaMA-7B-Chat} & 0.05 & 0.45 & 0.70 & 0.52 & 0.00 & 0.42 & 0.07 & 0.32 \\
\multicolumn{2}{c}{LLaMA--3B-Chat} & 0.55 & 0.73 & 0.62 & 0.62 & 0.00 & 0.09 & 0.09 & 0.39 \\
\multicolumn{2}{c}{LLaMA-70B-Chat} & 0.04 & 0.42 & 0.56 & 0.34 & 0.01 & 0.27 & 0.05 & 0.24 \\
\multicolumn{2}{c}{Mistral-7B-Instruct} & 0.17 & 0.30 & 0.32 & 0.33 & 0.20 & 0.36 & 0.02 & 0.24 \\
\multicolumn{2}{c}{DV003} & 0.38 & 0.58 & 0.75 & 0.63 & 0.25 & 0.48 & 0.32 & 0.48 \\
\multicolumn{2}{c}{GPT-3.5-Turbo} & 0.48 & 0.51 & 0.73 & 0.54 & 0.51 & 0.59 & 0.57 & 0.56 \\
\multicolumn{2}{c}{GPT-4} & 0.59 & 0.65 & 0.80 & 0.70 & 0.57 & 0.72 & 0.67 & 0.67 \\
\bottomrule
\end{tabular}

\caption{Detailed performance of various \textsc{Bactrian-X-11} finetuned \llm\  models on MLQA \cite{lewis2020mlqa}.}
\label{tab:results_mlqa_llama_bactrian_x_11}
\end{table*}
\begin{table*}
\scriptsize
\centering
\begin{tabular}{rrrrrrrrrr}
\toprule
\multicolumn{2}{c}{Models} & ar & de & en & es & hi & vi & zh & avg \\
\midrule
\multicolumn{2}{c}{LLaMA-7B-Chat} & 0.05 & 0.45 & 0.70 & 0.52 & 0.00 & 0.42 & 0.07 & 0.32 \\
\multicolumn{2}{c}{LLaMA--3B-Chat} & 0.55 & 0.73 & 0.62 & 0.62 & 0.00 & 0.09 & 0.09 & 0.39 \\
\multicolumn{2}{c}{LLaMA-70B-Chat} & 0.04 & 0.42 & 0.56 & 0.34 & 0.01 & 0.27 & 0.05 & 0.24 \\
\multicolumn{2}{c}{Mistral-7B-Instruct} & 0.17 & 0.30 & 0.32 & 0.33 & 0.20 & 0.36 & 0.02 & 0.24 \\
\multicolumn{2}{c}{DV003} & 0.38 & 0.58 & 0.75 & 0.63 & 0.25 & 0.48 & 0.32 & 0.48 \\
\multicolumn{2}{c}{GPT-3.5-Turbo} & 0.48 & 0.51 & 0.73 & 0.54 & 0.51 & 0.59 & 0.57 & 0.56 \\
\multicolumn{2}{c}{GPT-4} & 0.59 & 0.65 & 0.80 & 0.70 & 0.57 & 0.72 & 0.67 & 0.67 \\
\bottomrule
\end{tabular}

\caption{Detailed performance of various \textsc{Bactrian-X-22} finetuned \llm\ models on MLQA \cite{lewis2020mlqa}.}
\label{tab:results_mlqa_llama_bactrian}

\end{table*}

\begin{table*}
\scriptsize
\centering
\begin{tabular}{rrrrrrrrrr}
\toprule
\multicolumn{2}{c}{Models} & ar & de & en & es & hi & vi & zh & avg \\
\midrule
\multicolumn{2}{c}{LLaMA-7B-Chat} & 0.05 & 0.45 & 0.70 & 0.52 & 0.00 & 0.42 & 0.07 & 0.32 \\
\multicolumn{2}{c}{LLaMA--3B-Chat} & 0.55 & 0.73 & 0.62 & 0.62 & 0.00 & 0.09 & 0.09 & 0.39 \\
\multicolumn{2}{c}{LLaMA-70B-Chat} & 0.04 & 0.42 & 0.56 & 0.34 & 0.01 & 0.27 & 0.05 & 0.24 \\
\multicolumn{2}{c}{Mistral-7B-Instruct} & 0.17 & 0.30 & 0.32 & 0.33 & 0.20 & 0.36 & 0.02 & 0.24 \\
\multicolumn{2}{c}{DV003} & 0.38 & 0.58 & 0.75 & 0.63 & 0.25 & 0.48 & 0.32 & 0.48 \\
\multicolumn{2}{c}{GPT-3.5-Turbo} & 0.48 & 0.51 & 0.73 & 0.54 & 0.51 & 0.59 & 0.57 & 0.56 \\
\multicolumn{2}{c}{GPT-4} & 0.59 & 0.65 & 0.80 & 0.70 & 0.57 & 0.72 & 0.67 & 0.67 \\
\multicolumn{2}{c}{TULRv6} & \bfseries 0.76 & \bfseries 0.80 & 0.87 & \bfseries 0.82 & \bfseries 0.82 & \bfseries 0.82 & \bfseries 0.78 & \bfseries 0.81 \\
\multicolumn{2}{c}{XLM-R} & 0.67 & 0.70 & 0.83 & 0.74 & 0.71 & 0.74 & 0.62 & 0.72 \\
\multicolumn{2}{c}{mBERT} & 0.52 & 0.59 & 0.80 & 0.67 & 0.50 & 0.61 & 0.60 & 0.61 \\
\multicolumn{2}{c}{mT5} & 0.57 & 0.62 & 0.82 & 0.67 & 0.55 & 0.66 & 0.62 & 0.64 \\
\multicolumn{2}{c}{PALM2} & 0.22 & 0.44 & 0.69 & 0.45 & 0.36 & 0.49 & 0.06 & 0.39 \\
\midrule
rank & quantisation & ar & de & en & es & hi & vi & zh & avg \\
\midrule
64 & 8 & 0.61 & 0.73 & 0.89 & 0.81 & 0.60 & 0.80 & 0.41 & 0.69 \\
64 & 16 & 0.61 & 0.74 & \bfseries 0.89 & 0.81 & 0.61 & 0.81 & 0.41 & 0.70 \\
128 & 8 & 0.61 & 0.74 & \bfseries 0.89 & 0.81 & 0.60 & 0.80 & 0.41 & 0.69 \\
128 & 16 & 0.61 & 0.74 & 0.88 & 0.81 & 0.60 & 0.81 & 0.41 & 0.70 \\
\bottomrule
\end{tabular}

\caption{Detailed performance of various \textsc{alpaca} finetuned \mis\ models on MLQA \cite{lewis2020mlqa}.}
\label{tab:results_mlqa_mistral_alpaca}

\end{table*}
\begin{table*}
\scriptsize
\centering
\begin{tabular}{rrrrrrrrrr}
\toprule
\multicolumn{2}{c}{Model} & ar & de & en & es & hi & vi & zh & avg \\
\midrule
\multicolumn{2}{c}{LLaMA-7B-Chat} & 0.05 & 0.45 & 0.70 & 0.52 & 0.00 & 0.42 & 0.07 & 0.32 \\
\multicolumn{2}{c}{LLaMA--3B-Chat} & 0.55 & 0.73 & 0.62 & 0.62 & 0.00 & 0.09 & 0.09 & 0.39 \\
\multicolumn{2}{c}{LLaMA-70B-Chat} & 0.04 & 0.42 & 0.56 & 0.34 & 0.01 & 0.27 & 0.05 & 0.24 \\
\multicolumn{2}{c}{Mistral-7B-Instruct} & 0.17 & 0.30 & 0.32 & 0.33 & 0.20 & 0.36 & 0.02 & 0.24 \\
\multicolumn{2}{c}{DV003} & 0.38 & 0.58 & 0.75 & 0.63 & 0.25 & 0.48 & 0.32 & 0.48 \\
\multicolumn{2}{c}{GPT-3.5-Turbo} & 0.48 & 0.51 & 0.73 & 0.54 & 0.51 & 0.59 & 0.57 & 0.56 \\
\multicolumn{2}{c}{GPT-4} & 0.59 & 0.65 & 0.80 & 0.70 & 0.57 & 0.72 & 0.67 & 0.67 \\
\bottomrule
\end{tabular}

\caption{Detailed performance of various \textsc{multialpaca} finetuned \mis\ models on MLQA \cite{lewis2020mlqa}.}
\label{tab:results_mlqa_mistral_multialpaca}

\end{table*}
\begin{table*}
\scriptsize
\centering
\begin{tabular}{rrrrrrrrrr}
\toprule
\multicolumn{2}{c}{Models} & ar & de & en & es & hi & vi & zh & avg \\
\midrule
\multicolumn{2}{c}{LLaMA-7B-Chat} & 0.05 & 0.45 & 0.70 & 0.52 & 0.00 & 0.42 & 0.07 & 0.32 \\
\multicolumn{2}{c}{LLaMA--3B-Chat} & 0.55 & 0.73 & 0.62 & 0.62 & 0.00 & 0.09 & 0.09 & 0.39 \\
\multicolumn{2}{c}{LLaMA-70B-Chat} & 0.04 & 0.42 & 0.56 & 0.34 & 0.01 & 0.27 & 0.05 & 0.24 \\
\multicolumn{2}{c}{Mistral-7B-Instruct} & 0.17 & 0.30 & 0.32 & 0.33 & 0.20 & 0.36 & 0.02 & 0.24 \\
\multicolumn{2}{c}{DV003} & 0.38 & 0.58 & 0.75 & 0.63 & 0.25 & 0.48 & 0.32 & 0.48 \\
\multicolumn{2}{c}{GPT-3.5-Turbo} & 0.48 & 0.51 & 0.73 & 0.54 & 0.51 & 0.59 & 0.57 & 0.56 \\
\multicolumn{2}{c}{GPT-4} & 0.59 & 0.65 & 0.80 & 0.70 & 0.57 & 0.72 & 0.67 & 0.67 \\
\bottomrule
\end{tabular}

\caption{Detailed performance of various \textsc{Bactrian-X-11} finetuned \mis\  models on MLQA \cite{lewis2020mlqa}.}
\label{tab:results_mlqa_mistral_bactrian_x_11}
\end{table*}
\begin{table*}
\scriptsize
\centering
\begin{tabular}{rrrrrrrrrr}
\toprule
\multicolumn{2}{c}{Model} & ar & de & en & es & hi & vi & zh & avg \\
\midrule
\multicolumn{2}{c}{LLaMA-7B-Chat} & 0.05 & 0.45 & 0.70 & 0.52 & 0.00 & 0.42 & 0.07 & 0.32 \\
\multicolumn{2}{c}{LLaMA--3B-Chat} & 0.55 & 0.73 & 0.62 & 0.62 & 0.00 & 0.09 & 0.09 & 0.39 \\
\multicolumn{2}{c}{LLaMA-70B-Chat} & 0.04 & 0.42 & 0.56 & 0.34 & 0.01 & 0.27 & 0.05 & 0.24 \\
\multicolumn{2}{c}{Mistral-7B-Instruct} & 0.17 & 0.30 & 0.32 & 0.33 & 0.20 & 0.36 & 0.02 & 0.24 \\
\multicolumn{2}{c}{DV003} & 0.38 & 0.58 & 0.75 & 0.63 & 0.25 & 0.48 & 0.32 & 0.48 \\
\multicolumn{2}{c}{GPT-3.5-Turbo} & 0.48 & 0.51 & 0.73 & 0.54 & 0.51 & 0.59 & 0.57 & 0.56 \\
\multicolumn{2}{c}{GPT-4} & 0.59 & 0.65 & 0.80 & 0.70 & 0.57 & 0.72 & 0.67 & 0.67 \\
\bottomrule
\end{tabular}

\caption{Detailed performance of various \textsc{Bactrian-X-22} finetuned \mis\ models on MLQA \cite{lewis2020mlqa}.}
\label{tab:results_mlqa_mistral_bactrian}

\end{table*}

\begin{table*}
\scriptsize
\centering
\begin{tabular}{rrrrrrrrrrrrrr}
\toprule
\multicolumn{2}{c}{Model} & et & ht & id & it & qu & sw & ta & th & tr & vi & zh & avg \\
\midrule
\multicolumn{2}{c}{LLaMA-7B-Chat} & 0.51 & 0.51 & 0.59 & 0.71 & 0.50 & 0.51 & 0.50 & 0.52 & 0.53 & 0.58 & 0.59 & 0.55 \\
\multicolumn{2}{c}{LLaMA--3B-Chat} & 0.51 & 0.49 & 0.72 & 0.80 & 0.50 & 0.50 & 0.00 & 0.00 & 0.54 & 0.02 & 0.70 & 0.44 \\
\multicolumn{2}{c}{LLaMA-70B-Chat} & 0.47 & 0.44 & 0.40 & 0.53 & 0.37 & 0.34 & 0.21 & 0.25 & 0.44 & 0.52 & 0.35 & 0.39 \\
\multicolumn{2}{c}{Mistral-7B-Instruct} & 0.52 & 0.50 & 0.61 & 0.73 & 0.51 & 0.49 & 0.41 & 0.51 & 0.54 & 0.50 & 0.50 & 0.53 \\
\multicolumn{2}{c}{DV003} & 0.88 & 0.75 & 0.91 & 0.96 & 0.55 & 0.64 & 0.54 & 0.67 & 0.88 & -- & -- & 0.75 \\
\multicolumn{2}{c}{GPT-3.5-Turbo} & 0.91 & 0.72 & 0.90 & 0.95 & 0.55 & 0.82 & 0.59 & 0.78 & 0.91 & -- & -- & 0.79 \\
\multicolumn{2}{c}{GPT-4} & \bfseries 0.99 & \bfseries 0.93 & 0.98 & \bfseries 1.00 & 0.59 & \bfseries 0.94 & 0.80 & 0.88 & \bfseries 0.97 & -- & -- & 0.90 \\
\bottomrule
\end{tabular}

\caption{Detailed performance of various \textsc{alpaca} finetuned \llm\ models on XCOPA \cite{ponti2020xcopa}.}
\label{tab:results_xcopa_llama_alpaca}

\end{table*}
\begin{table*}
\scriptsize
\centering
\begin{tabular}{rrrrrrrrrrrrrr}
\toprule
\multicolumn{2}{c}{Model} & et & ht & id & it & qu & sw & ta & th & tr & vi & zh & avg \\
\midrule
\multicolumn{2}{c}{LLaMA-7B-Chat} & 0.51 & 0.51 & 0.59 & 0.71 & 0.50 & 0.51 & 0.50 & 0.52 & 0.53 & 0.58 & 0.59 & 0.55 \\
\multicolumn{2}{c}{LLaMA--3B-Chat} & 0.51 & 0.49 & 0.72 & 0.80 & 0.50 & 0.50 & 0.00 & 0.00 & 0.54 & 0.02 & 0.70 & 0.44 \\
\multicolumn{2}{c}{LLaMA-70B-Chat} & 0.47 & 0.44 & 0.40 & 0.53 & 0.37 & 0.34 & 0.21 & 0.25 & 0.44 & 0.52 & 0.35 & 0.39 \\
\multicolumn{2}{c}{Mistral-7B-Instruct} & 0.52 & 0.50 & 0.61 & 0.73 & 0.51 & 0.49 & 0.41 & 0.51 & 0.54 & 0.50 & 0.50 & 0.53 \\
\multicolumn{2}{c}{DV003} & 0.88 & 0.75 & 0.91 & 0.96 & 0.55 & 0.64 & 0.54 & 0.67 & 0.88 & -- & -- & 0.75 \\
\multicolumn{2}{c}{GPT-3.5-Turbo} & 0.91 & 0.72 & 0.90 & 0.95 & 0.55 & 0.82 & 0.59 & 0.78 & 0.91 & -- & -- & 0.79 \\
\multicolumn{2}{c}{GPT-4} & \bfseries 0.99 & \bfseries 0.93 & 0.98 & \bfseries 1.00 & 0.59 & \bfseries 0.94 & 0.80 & 0.88 & \bfseries 0.97 & -- & -- & 0.90 \\
\bottomrule
\end{tabular}

\caption{Detailed performance of various \textsc{multialpaca} finetuned \llm\ models on XCOPA \cite{ponti2020xcopa}.}
\label{tab:results_xcopa_llama_multialpaca}

\end{table*}
\begin{table*}
\scriptsize
\centering
\begin{tabular}{rrrrrrrrrrrrrr}
\toprule
\multicolumn{2}{c}{Model} & et & ht & id & it & qu & sw & ta & th & tr & vi & zh & avg \\
\midrule
\multicolumn{2}{c}{LLaMA-7B-Chat} & 0.51 & 0.51 & 0.59 & 0.71 & 0.50 & 0.51 & 0.50 & 0.52 & 0.53 & 0.58 & 0.59 & 0.55 \\
\multicolumn{2}{c}{LLaMA--3B-Chat} & 0.51 & 0.49 & 0.72 & 0.80 & 0.50 & 0.50 & 0.00 & 0.00 & 0.54 & 0.02 & 0.70 & 0.44 \\
\multicolumn{2}{c}{LLaMA-70B-Chat} & 0.47 & 0.44 & 0.40 & 0.53 & 0.37 & 0.34 & 0.21 & 0.25 & 0.44 & 0.52 & 0.35 & 0.39 \\
\multicolumn{2}{c}{Mistral-7B-Instruct} & 0.52 & 0.50 & 0.61 & 0.73 & 0.51 & 0.49 & 0.41 & 0.51 & 0.54 & 0.50 & 0.50 & 0.53 \\
\multicolumn{2}{c}{DV003} & 0.88 & 0.75 & 0.91 & 0.96 & 0.55 & 0.64 & 0.54 & 0.67 & 0.88 & -- & -- & 0.75 \\
\multicolumn{2}{c}{GPT-3.5-Turbo} & 0.91 & 0.72 & 0.90 & 0.95 & 0.55 & 0.82 & 0.59 & 0.78 & 0.91 & -- & -- & 0.79 \\
\multicolumn{2}{c}{GPT-4} & \bfseries 0.99 & \bfseries 0.93 & 0.98 & \bfseries 1.00 & 0.59 & \bfseries 0.94 & 0.80 & 0.88 & \bfseries 0.97 & -- & -- & 0.90 \\
\bottomrule
\end{tabular}

\caption{Detailed performance of various \textsc{Bactrian-X-11} finetuned \llm\  models on XCOPA \cite{ponti2020xcopa}.}
\label{tab:results_xcopa_llama_bactrian_x_11}
\end{table*}
\begin{table*}
\scriptsize
\centering
\begin{tabular}{rrrrrrrrrrrrrr}
\toprule
\multicolumn{2}{c}{Model} & et & ht & id & it & qu & sw & ta & th & tr & vi & zh & avg \\
\midrule
\multicolumn{2}{c}{LLaMA-7B-Chat} & 0.51 & 0.51 & 0.59 & 0.71 & 0.50 & 0.51 & 0.50 & 0.52 & 0.53 & 0.58 & 0.59 & 0.55 \\
\multicolumn{2}{c}{LLaMA--3B-Chat} & 0.51 & 0.49 & 0.72 & 0.80 & 0.50 & 0.50 & 0.00 & 0.00 & 0.54 & 0.02 & 0.70 & 0.44 \\
\multicolumn{2}{c}{LLaMA-70B-Chat} & 0.47 & 0.44 & 0.40 & 0.53 & 0.37 & 0.34 & 0.21 & 0.25 & 0.44 & 0.52 & 0.35 & 0.39 \\
\multicolumn{2}{c}{Mistral-7B-Instruct} & 0.52 & 0.50 & 0.61 & 0.73 & 0.51 & 0.49 & 0.41 & 0.51 & 0.54 & 0.50 & 0.50 & 0.53 \\
\multicolumn{2}{c}{DV003} & 0.88 & 0.75 & 0.91 & 0.96 & 0.55 & 0.64 & 0.54 & 0.67 & 0.88 & -- & -- & 0.75 \\
\multicolumn{2}{c}{GPT-3.5-Turbo} & 0.91 & 0.72 & 0.90 & 0.95 & 0.55 & 0.82 & 0.59 & 0.78 & 0.91 & -- & -- & 0.79 \\
\multicolumn{2}{c}{GPT-4} & \bfseries 0.99 & \bfseries 0.93 & 0.98 & \bfseries 1.00 & 0.59 & \bfseries 0.94 & 0.80 & 0.88 & \bfseries 0.97 & -- & -- & 0.90 \\
\bottomrule
\end{tabular}

\caption{Detailed performance of various \textsc{Bactrian-X-22} finetuned \llm\ models on XCOPA \cite{ponti2020xcopa}.}
\label{tab:results_xcopa_llama_bactrian}

\end{table*}

\begin{table*}
\scriptsize
\centering
\begin{tabular}{rrrrrrrrrrrrrr}
\toprule
\multicolumn{2}{c}{Model} & et & ht & id & it & qu & sw & ta & th & tr & vi & zh & avg \\
\midrule
\multicolumn{2}{c}{LLaMA-7B-Chat} & 0.51 & 0.51 & 0.59 & 0.71 & 0.50 & 0.51 & 0.50 & 0.52 & 0.53 & 0.58 & 0.59 & 0.55 \\
\multicolumn{2}{c}{LLaMA--3B-Chat} & 0.51 & 0.49 & 0.72 & 0.80 & 0.50 & 0.50 & 0.00 & 0.00 & 0.54 & 0.02 & 0.70 & 0.44 \\
\multicolumn{2}{c}{LLaMA-70B-Chat} & 0.47 & 0.44 & 0.40 & 0.53 & 0.37 & 0.34 & 0.21 & 0.25 & 0.44 & 0.52 & 0.35 & 0.39 \\
\multicolumn{2}{c}{Mistral-7B-Instruct} & 0.52 & 0.50 & 0.61 & 0.73 & 0.51 & 0.49 & 0.41 & 0.51 & 0.54 & 0.50 & 0.50 & 0.53 \\
\multicolumn{2}{c}{DV003} & 0.88 & 0.75 & 0.91 & 0.96 & 0.55 & 0.64 & 0.54 & 0.67 & 0.88 & -- & -- & 0.75 \\
\multicolumn{2}{c}{GPT-3.5-Turbo} & 0.91 & 0.72 & 0.90 & 0.95 & 0.55 & 0.82 & 0.59 & 0.78 & 0.91 & -- & -- & 0.79 \\
\multicolumn{2}{c}{GPT-4} & \bfseries 0.99 & \bfseries 0.93 & 0.98 & \bfseries 1.00 & 0.59 & \bfseries 0.94 & 0.80 & 0.88 & \bfseries 0.97 & -- & -- & 0.90 \\
\bottomrule
\end{tabular}

\caption{Detailed performance of various \textsc{alpaca} finetuned \mis\ models on XCOPA \cite{ponti2020xcopa}.}
\label{tab:results_xcopa_mistral_alpaca}

\end{table*}
\begin{table*}
\scriptsize
\centering
\begin{tabular}{rrrrrrrrrrrrrr}
\toprule
\multicolumn{2}{c}{Model} & et & ht & id & it & qu & sw & ta & th & tr & vi & zh & avg \\
\midrule
\multicolumn{2}{c}{LLaMA-7B-Chat} & 0.51 & 0.51 & 0.59 & 0.71 & 0.50 & 0.51 & 0.50 & 0.52 & 0.53 & 0.58 & 0.59 & 0.55 \\
\multicolumn{2}{c}{LLaMA--3B-Chat} & 0.51 & 0.49 & 0.72 & 0.80 & 0.50 & 0.50 & 0.00 & 0.00 & 0.54 & 0.02 & 0.70 & 0.44 \\
\multicolumn{2}{c}{LLaMA-70B-Chat} & 0.47 & 0.44 & 0.40 & 0.53 & 0.37 & 0.34 & 0.21 & 0.25 & 0.44 & 0.52 & 0.35 & 0.39 \\
\multicolumn{2}{c}{Mistral-7B-Instruct} & 0.52 & 0.50 & 0.61 & 0.73 & 0.51 & 0.49 & 0.41 & 0.51 & 0.54 & 0.50 & 0.50 & 0.53 \\
\multicolumn{2}{c}{DV003} & 0.88 & 0.75 & 0.91 & 0.96 & 0.55 & 0.64 & 0.54 & 0.67 & 0.88 & -- & -- & 0.75 \\
\multicolumn{2}{c}{GPT-3.5-Turbo} & 0.91 & 0.72 & 0.90 & 0.95 & 0.55 & 0.82 & 0.59 & 0.78 & 0.91 & -- & -- & 0.79 \\
\multicolumn{2}{c}{GPT-4} & \bfseries 0.99 & \bfseries 0.93 & 0.98 & \bfseries 1.00 & 0.59 & \bfseries 0.94 & 0.80 & 0.88 & \bfseries 0.97 & -- & -- & 0.90 \\
\bottomrule
\end{tabular}

\caption{Detailed performance of various \textsc{multialpaca} finetuned \mis\ models on XCOPA \cite{ponti2020xcopa}.}
\label{tab:results_xcopa_mistral_multialpaca}

\end{table*}
\begin{table*}
\scriptsize
\centering
\begin{tabular}{rrrrrrrrrrrrrr}
\toprule
\multicolumn{2}{c}{Model} & et & ht & id & it & qu & sw & ta & th & tr & vi & zh & avg \\
\midrule
\multicolumn{2}{c}{LLaMA-7B-Chat} & 0.51 & 0.51 & 0.59 & 0.71 & 0.50 & 0.51 & 0.50 & 0.52 & 0.53 & 0.58 & 0.59 & 0.55 \\
\multicolumn{2}{c}{LLaMA--3B-Chat} & 0.51 & 0.49 & 0.72 & 0.80 & 0.50 & 0.50 & 0.00 & 0.00 & 0.54 & 0.02 & 0.70 & 0.44 \\
\multicolumn{2}{c}{LLaMA-70B-Chat} & 0.47 & 0.44 & 0.40 & 0.53 & 0.37 & 0.34 & 0.21 & 0.25 & 0.44 & 0.52 & 0.35 & 0.39 \\
\multicolumn{2}{c}{Mistral-7B-Instruct} & 0.52 & 0.50 & 0.61 & 0.73 & 0.51 & 0.49 & 0.41 & 0.51 & 0.54 & 0.50 & 0.50 & 0.53 \\
\multicolumn{2}{c}{DV003} & 0.88 & 0.75 & 0.91 & 0.96 & 0.55 & 0.64 & 0.54 & 0.67 & 0.88 & -- & -- & 0.75 \\
\multicolumn{2}{c}{GPT-3.5-Turbo} & 0.91 & 0.72 & 0.90 & 0.95 & 0.55 & 0.82 & 0.59 & 0.78 & 0.91 & -- & -- & 0.79 \\
\multicolumn{2}{c}{GPT-4} & \bfseries 0.99 & \bfseries 0.93 & 0.98 & \bfseries 1.00 & 0.59 & \bfseries 0.94 & 0.80 & 0.88 & \bfseries 0.97 & -- & -- & 0.90 \\
\bottomrule
\end{tabular}

\caption{Detailed performance of various \textsc{Bactrian-X-11} finetuned \mis\  models on XCOPA \cite{ponti2020xcopa}.}
\label{tab:results_xcopa_mistral_bactrian_x_11}
\end{table*}
\begin{table*}
\scriptsize
\centering
\begin{tabular}{rrrrrrrrrrrrrr}
\toprule
\multicolumn{2}{c}{Model} & et & ht & id & it & qu & sw & ta & th & tr & vi & zh & avg \\
\midrule
\multicolumn{2}{c}{LLaMA-7B-Chat} & 0.51 & 0.51 & 0.59 & 0.71 & 0.50 & 0.51 & 0.50 & 0.52 & 0.53 & 0.58 & 0.59 & 0.55 \\
\multicolumn{2}{c}{LLaMA--3B-Chat} & 0.51 & 0.49 & 0.72 & 0.80 & 0.50 & 0.50 & 0.00 & 0.00 & 0.54 & 0.02 & 0.70 & 0.44 \\
\multicolumn{2}{c}{LLaMA-70B-Chat} & 0.47 & 0.44 & 0.40 & 0.53 & 0.37 & 0.34 & 0.21 & 0.25 & 0.44 & 0.52 & 0.35 & 0.39 \\
\multicolumn{2}{c}{Mistral-7B-Instruct} & 0.52 & 0.50 & 0.61 & 0.73 & 0.51 & 0.49 & 0.41 & 0.51 & 0.54 & 0.50 & 0.50 & 0.53 \\
\multicolumn{2}{c}{DV003} & 0.88 & 0.75 & 0.91 & 0.96 & 0.55 & 0.64 & 0.54 & 0.67 & 0.88 & -- & -- & 0.75 \\
\multicolumn{2}{c}{GPT-3.5-Turbo} & 0.91 & 0.72 & 0.90 & 0.95 & 0.55 & 0.82 & 0.59 & 0.78 & 0.91 & -- & -- & 0.79 \\
\multicolumn{2}{c}{GPT-4} & \bfseries 0.99 & \bfseries 0.93 & 0.98 & \bfseries 1.00 & 0.59 & \bfseries 0.94 & 0.80 & 0.88 & \bfseries 0.97 & -- & -- & 0.90 \\
\bottomrule
\end{tabular}

\caption{Detailed performance of various \textsc{Bactrian-X-22} finetuned \mis\ models on XCOPA \cite{ponti2020xcopa}.}
\label{tab:results_xcopa_mistral_bactrian}

\end{table*}

\begin{table*}
\tiny
\centering
\begin{tabular}{rrrrrrrrrrrrrrrrrr}
\toprule
\multicolumn{2}{c}{Model} & ar & bg & de & el & en & es & fr & hi & ru & sw & th & tr & ur & vi & zh & avg \\
\midrule
\multicolumn{2}{c}{LLaMA-7B-Chat} & 0.39 & 0.45 & 0.45 & 0.39 & 0.56 & 0.50 & 0.50 & 0.37 & 0.48 & 0.33 & 0.35 & 0.40 & 0.36 & 0.41 & 0.45 & 0.43 \\
\multicolumn{2}{c}{LLaMA--3B-Chat} & 0.37 & 0.51 & 0.50 & 0.00 & 0.55 & 0.51 & 0.52 & 0.00 & 0.50 & 0.36 & 0.00 & 0.45 & 0.29 & 0.48 & 0.48 & 0.37 \\
\multicolumn{2}{c}{LLaMA-70B-Chat} & 0.35 & 0.55 & 0.58 & 0.41 & 0.63 & 0.55 & 0.55 & 0.31 & 0.55 & 0.39 & 0.35 & 0.49 & 0.42 & 0.49 & 0.51 & 0.48 \\
\multicolumn{2}{c}{Mistral-7B-Instruct} & 0.35 & 0.40 & 0.38 & 0.35 & 0.43 & 0.41 & 0.41 & 0.35 & 0.43 & 0.34 & 0.33 & 0.40 & 0.34 & 0.33 & 0.40 & 0.38 \\
\multicolumn{2}{c}{DV003} & 0.52 & 0.62 & 0.66 & 0.60 & 0.80 & 0.71 & 0.66 & 0.48 & 0.62 & 0.50 & 0.51 & 0.58 & 0.50 & 0.56 & 0.58 & 0.59 \\
\multicolumn{2}{c}{GPT-3.5-Turbo} & 0.59 & 0.64 & 0.67 & 0.65 & 0.76 & 0.70 & 0.68 & 0.55 & 0.62 & 0.56 & 0.54 & 0.63 & 0.49 & 0.61 & 0.62 & 0.62 \\
\multicolumn{2}{c}{GPT-4} & 0.73 & 0.77 & 0.79 & 0.79 & 0.85 & 0.79 & 0.79 & 0.72 & 0.74 & 0.71 & 0.69 & 0.76 & 0.68 & 0.74 & 0.75 & 0.75 \\
\bottomrule
\end{tabular}
\caption{Detailed performance of various \textsc{alpaca} finetuned \llm\ models on XNLI \cite{Conneau2018xnli}.}
\label{tab:results_xnli_llama_alpaca}
\end{table*}
\begin{table*}
\tiny
\centering
\begin{tabular}{rrrrrrrrrrrrrrrrrr}
\toprule
\multicolumn{2}{c}{Model} & ar & bg & de & el & en & es & fr & hi & ru & sw & th & tr & ur & vi & zh & avg \\
\midrule
\multicolumn{2}{c}{LLaMA-7B-Chat} & 0.39 & 0.45 & 0.45 & 0.39 & 0.56 & 0.50 & 0.50 & 0.37 & 0.48 & 0.33 & 0.35 & 0.40 & 0.36 & 0.41 & 0.45 & 0.43 \\
\multicolumn{2}{c}{LLaMA--3B-Chat} & 0.37 & 0.51 & 0.50 & 0.00 & 0.55 & 0.51 & 0.52 & 0.00 & 0.50 & 0.36 & 0.00 & 0.45 & 0.29 & 0.48 & 0.48 & 0.37 \\
\multicolumn{2}{c}{LLaMA-70B-Chat} & 0.35 & 0.55 & 0.58 & 0.41 & 0.63 & 0.55 & 0.55 & 0.31 & 0.55 & 0.39 & 0.35 & 0.49 & 0.42 & 0.49 & 0.51 & 0.48 \\
\multicolumn{2}{c}{Mistral-7B-Instruct} & 0.35 & 0.40 & 0.38 & 0.35 & 0.43 & 0.41 & 0.41 & 0.35 & 0.43 & 0.34 & 0.33 & 0.40 & 0.34 & 0.33 & 0.40 & 0.38 \\
\multicolumn{2}{c}{DV003} & 0.52 & 0.62 & 0.66 & 0.60 & 0.80 & 0.71 & 0.66 & 0.48 & 0.62 & 0.50 & 0.51 & 0.58 & 0.50 & 0.56 & 0.58 & 0.59 \\
\multicolumn{2}{c}{GPT-3.5-Turbo} & 0.59 & 0.64 & 0.67 & 0.65 & 0.76 & 0.70 & 0.68 & 0.55 & 0.62 & 0.56 & 0.54 & 0.63 & 0.49 & 0.61 & 0.62 & 0.62 \\
\multicolumn{2}{c}{GPT-4} & 0.73 & 0.77 & 0.79 & 0.79 & 0.85 & 0.79 & 0.79 & 0.72 & 0.74 & 0.71 & 0.69 & 0.76 & 0.68 & 0.74 & 0.75 & 0.75 \\
\bottomrule
\end{tabular}

\caption{Detailed performance of various \textsc{multialpaca} finetuned \llm\ models on XNLI \cite{Conneau2018xnli}.}
\label{tab:results_xnli_llama_multialpaca}

\end{table*}
\begin{table*}
\tiny
\centering
\begin{tabular}{rrrrrrrrrrrrrrrrrr}
\toprule
\multicolumn{2}{c}{Model} & ar & bg & de & el & en & es & fr & hi & ru & sw & th & tr & ur & vi & zh & avg \\
\midrule
\multicolumn{2}{c}{LLaMA-7B-Chat} & 0.39 & 0.45 & 0.45 & 0.39 & 0.56 & 0.50 & 0.50 & 0.37 & 0.48 & 0.33 & 0.35 & 0.40 & 0.36 & 0.41 & 0.45 & 0.43 \\
\multicolumn{2}{c}{LLaMA--3B-Chat} & 0.37 & 0.51 & 0.50 & 0.00 & 0.55 & 0.51 & 0.52 & 0.00 & 0.50 & 0.36 & 0.00 & 0.45 & 0.29 & 0.48 & 0.48 & 0.37 \\
\multicolumn{2}{c}{LLaMA-70B-Chat} & 0.35 & 0.55 & 0.58 & 0.41 & 0.63 & 0.55 & 0.55 & 0.31 & 0.55 & 0.39 & 0.35 & 0.49 & 0.42 & 0.49 & 0.51 & 0.48 \\
\multicolumn{2}{c}{Mistral-7B-Instruct} & 0.35 & 0.40 & 0.38 & 0.35 & 0.43 & 0.41 & 0.41 & 0.35 & 0.43 & 0.34 & 0.33 & 0.40 & 0.34 & 0.33 & 0.40 & 0.38 \\
\multicolumn{2}{c}{DV003} & 0.52 & 0.62 & 0.66 & 0.60 & 0.80 & 0.71 & 0.66 & 0.48 & 0.62 & 0.50 & 0.51 & 0.58 & 0.50 & 0.56 & 0.58 & 0.59 \\
\multicolumn{2}{c}{GPT-3.5-Turbo} & 0.59 & 0.64 & 0.67 & 0.65 & 0.76 & 0.70 & 0.68 & 0.55 & 0.62 & 0.56 & 0.54 & 0.63 & 0.49 & 0.61 & 0.62 & 0.62 \\
\multicolumn{2}{c}{GPT-4} & 0.73 & 0.77 & 0.79 & 0.79 & 0.85 & 0.79 & 0.79 & 0.72 & 0.74 & 0.71 & 0.69 & 0.76 & 0.68 & 0.74 & 0.75 & 0.75 \\
\bottomrule
\end{tabular}

\caption{Detailed performance of various \textsc{Bactrian-X-11} finetuned \llm\  models on XNLI \cite{Conneau2018xnli}.}
\label{tab:results_xnli_llama_bactrian_x_11}
\end{table*}
\begin{table*}
\tiny
\centering
\begin{tabular}{rrrrrrrrrrrrrrrrrr}
\toprule
\multicolumn{2}{c}{Model} & ar & bg & de & el & en & es & fr & hi & ru & sw & th & tr & ur & vi & zh & avg \\
\midrule
\multicolumn{2}{c}{LLaMA-7B-Chat} & 0.39 & 0.45 & 0.45 & 0.39 & 0.56 & 0.50 & 0.50 & 0.37 & 0.48 & 0.33 & 0.35 & 0.40 & 0.36 & 0.41 & 0.45 & 0.43 \\
\multicolumn{2}{c}{LLaMA--3B-Chat} & 0.37 & 0.51 & 0.50 & 0.00 & 0.55 & 0.51 & 0.52 & 0.00 & 0.50 & 0.36 & 0.00 & 0.45 & 0.29 & 0.48 & 0.48 & 0.37 \\
\multicolumn{2}{c}{LLaMA-70B-Chat} & 0.35 & 0.55 & 0.58 & 0.41 & 0.63 & 0.55 & 0.55 & 0.31 & 0.55 & 0.39 & 0.35 & 0.49 & 0.42 & 0.49 & 0.51 & 0.48 \\
\multicolumn{2}{c}{Mistral-7B-Instruct} & 0.35 & 0.40 & 0.38 & 0.35 & 0.43 & 0.41 & 0.41 & 0.35 & 0.43 & 0.34 & 0.33 & 0.40 & 0.34 & 0.33 & 0.40 & 0.38 \\
\multicolumn{2}{c}{DV003} & 0.52 & 0.62 & 0.66 & 0.60 & 0.80 & 0.71 & 0.66 & 0.48 & 0.62 & 0.50 & 0.51 & 0.58 & 0.50 & 0.56 & 0.58 & 0.59 \\
\multicolumn{2}{c}{GPT-3.5-Turbo} & 0.59 & 0.64 & 0.67 & 0.65 & 0.76 & 0.70 & 0.68 & 0.55 & 0.62 & 0.56 & 0.54 & 0.63 & 0.49 & 0.61 & 0.62 & 0.62 \\
\multicolumn{2}{c}{GPT-4} & 0.73 & 0.77 & 0.79 & 0.79 & 0.85 & 0.79 & 0.79 & 0.72 & 0.74 & 0.71 & 0.69 & 0.76 & 0.68 & 0.74 & 0.75 & 0.75 \\
\bottomrule
\end{tabular}

\caption{Detailed performance of various \textsc{Bactrian-X-22} finetuned \llm\ models on XNLI \cite{Conneau2018xnli}.}
\label{tab:results_xnli_llama_bactrian}

\end{table*}

\input{tables/table_xnli_mistral_alpaca}
\input{tables/table_xnli_mistral_multialpaca}
\input{tables/table_xnli_mistral_Bactrian-X-11}
\input{tables/table_xnli_mistral_Bactrian-X-22}

\input{tables/table_xquad_llama_alpaca}
\input{tables/table_xquad_llama_multialpaca}
\input{tables/table_xquad_llama_Bactrian-X-11}
\input{tables/table_xquad_llama_Bactrian-X-22}

\input{tables/table_xquad_mistral_alpaca}
\input{tables/table_xquad_mistral_multialpaca}
\input{tables/table_xquad_mistral_Bactrian-X-11}
\input{tables/table_xquad_mistral_Bactrian-X-22}

\input{tables/table_xlsum_llama_alpaca}
\input{tables/table_xlsum_llama_multialpaca}
\input{tables/table_xlsum_llama_Bactrian-X-11}
\input{tables/table_xlsum_llama_Bactrian-X-22}

\input{tables/table_xlsum_mistral_alpaca}
\input{tables/table_xlsum_mistral_multialpaca}
\input{tables/table_xlsum_mistral_Bactrian-X-11}
\input{tables/table_xlsum_mistral_Bactrian-X-22}

%\clearpage

% \begin{landscape}
%     \setlength{\tabcolsep}{3pt}
%     % \input{tables/table_belebele_llama_multialpaca}
%     \input{tables/table_belebele_llama_Bactrian-X-22}
% \end{landscape}

% \begin{landscape}
%     \setlength{\tabcolsep}{3pt}
%     \input{tables/table_belebele_mistral_multialpaca}
%     \input{tables/table_belebele_mistral_Bactrian-X-22}
% \end{landscape}

% \begin{landscape}
%     \setlength{\tabcolsep}{3pt}
%     \input{tables/table_belebele_mistral_multialpaca}
%     % \input{tables/table_belebele_mistral_Bactrian-X-22}
% \end{landscape}

\end{document}